\documentclass[lettersize,journal]{IEEEtran}
\usepackage{amsmath,amsfonts}
\usepackage{algorithm}
\usepackage{algorithmicx}
\usepackage{algpseudocode}
\usepackage{array}
\usepackage{textcomp}
\usepackage{stfloats}
\usepackage{url}
\usepackage{verbatim}
\usepackage{graphicx}
\usepackage{cite}
\usepackage{csquotes}
\usepackage{rotating}
\usepackage{multirow}
\usepackage{times}
\usepackage{epsfig}
\usepackage{color}
\usepackage[normalem]{ulem}
\usepackage{booktabs}
\usepackage{subfigure}
\usepackage{colortbl}

\usepackage[colorlinks=true,linkcolor=black,citecolor=blue,urlcolor=blue,]{hyperref}

\newcommand\eg{\emph{e.g.}} 
\newcommand\ie{\emph{i.e.}} 
 
\newcommand\etc{\emph{etc.}}
 
\newcommand\etal{\emph{et al.}}

\def\Rev#1{\textcolor{black}{#1}}

\definecolor{Ocean}{RGB}{230,235,250}
\begin{document}

\title{Towards Deeper Understanding of \\
Camouflaged Object Detection}


\author{Yunqiu Lv$^\sharp$,~
        Jing Zhang$^\sharp$,~
        Yuchao Dai*,~
        Aixuan Li,~
        Nick Barnes~
        and Deng-Ping Fan~\\
\IEEEcompsocitemizethanks{
\IEEEcompsocthanksitem Yunqiu Lv, Yuchao Dai and Aixuan Li are with School of Electronics and Information, Northwestern Polytechnical University, Xi'an, China and State Key Laboratory of Integrated Services Networks (Xidian University).
\IEEEcompsocthanksitem Jing Zhang and Nick Barnes are with School of Computing, the Australian National University, Canberra, Australia.
\IEEEcompsocthanksitem Deng-Ping Fan is with the CS, Nankai University, Tianjin, China.
\IEEEcompsocthanksitem A preliminary version of this work has appeared in CVPR2021~\cite{yunqiu_cod21}.
\IEEEcompsocthanksitem The first two authors contributed equally. Yuchao Dai is the corresponding author (daiyuchao@nwpu.edu.cn). This work was supported in part by the National Natural Science Foundation of China (Nos. 62271410, 61871325).
}
}

\markboth{Journal of \LaTeX\ Class Files,~Vol.~14, No.~8, August~2021}%
{Shell \MakeLowercase{\textit{et al.}}: A Sample Article Using IEEEtran.cls for IEEE Journals}


\maketitle

\begin{abstract}
Preys in the wild evolve to be camouflaged to avoid being recognized by predators. In this way, camouflage acts as a key defence mechanism across species that is critical to survival. To detect and segment the whole scope of a camouflaged object, camouflaged object detection (COD) is introduced as a binary segmentation task, with the binary ground truth camouflage map indicating the exact regions of the camouflaged objects. In this paper, we revisit this task and argue that the binary segmentation setting fails to fully understand the concept of camouflage. We find that explicitly modeling the conspicuousness of camouflaged objects against their particular backgrounds can not only lead to a better understanding about camouflage, but also provide guidance to designing more sophisticated camouflage techniques. Furthermore, we observe that it is some specific parts of camouflaged objects that make them detectable by predators. With the above understanding about camouflaged objects, we present the first triple-task learning framework to simultaneously \emph{localize, segment, and rank} camouflaged objects, indicating the conspicuousness level of camouflage. As no corresponding datasets exist for either the localization model or the ranking model, we generate localization maps with an eye tracker, which are then processed according to the instance level labels to generate our ranking-based training and testing dataset. We also contribute the largest COD testing set to comprehensively analyse performance of the COD models. Experimental results show that our triple-task learning framework achieves new state-of-the-art, leading to a more explainable COD network. Our code, data, and results are available at: \url{https://github.com/JingZhang617/COD-Rank-Localize-and-Segment}.

\end{abstract}


\section{Introduction}
\IEEEPARstart{C}{amouflage} is an important tool for animals to reduce the probability of being detected or recognized \cite{skelhorn2016cognition,merilaita2017camouflage, background_matching,background_matching_color_disruptive}. In order to hide in the environment, animals are evolved to share similar patterns with their habitats
to reduce the intensity of their distinctive signals or enhance the confusing signals. Specifically, background matching and disruptive coloration \cite{background_matching,background_matching_color_disruptive} are two main strategies that are widely adopted as
anti-predator defences to achieve camouflage. The former changes the appearance of prey by capturing and imitating the color and pattern from the background, thereby blending in the environment \cite{background_matching_color_disruptive}. The latter offers the high-contrast markings near the body edges and breaks the true edges that
act as the salient cue \cite{skelhorn2016cognition}.

No matter which camouflage strategy is used,
premise is that the prey aim to mimic
their habitats to generate sophisticated camouflage patterns to avoid being detected.
Generally, for highly evolved prey, 
 no clear salient features exist
 within the object. It is usually 
subtle contrast features, \eg~patterns around the
eyes of the prey, that make 
camouflaged objects noticeable.
We argue that with 
an understanding of
camouflage, the hierarchical levels of human visual perception and the evolution of animals in the wild could be finely investigated \cite{merilaita2017camouflage}. Specifically,
visual recognition relies on 
difference of appearance to distinguish the objects of interest from the background.
By delving into the mechanism that 
viewers use to
search for the camouflaged object, we can capture more subtle differences between the object and the background for more fine-grained recognition. 

Camouflaged object detection (COD) is the task to segment 
camouflaged objects, which is usually defined as a binary segmentation task, and has great potential in many computer vision related fields, such as
agriculture (\eg~pest control \cite{cheng2017pest, wu2019ip102}), city planning (\eg~road identification \cite{zhou2018d}), and medicine (\eg~lung infection detection \cite{fan2020inf}). It also nicely contrasts with salient object detection \cite{han2014background,sun2021ampnet, wei2020f3net,scrn_sal}, another widely studied binary segmentation task, in that contrast is hidden rather than clear.
Early works \cite{Quantifyingcamouflage ,pike2018quantifying, tankus2001convexity} in COD used
manual features, such as luminance, color, texture, and edge, to model the conspicuousness of 
camouflaged objects. However, 
manual features are
vulnerable to 
attack of sophisticated camouflage.
In recent years, many studies take advantage of the powerful feature representation ability of deep learning to construct 
structure contrast between the object and the background \cite{scrn_sal,wei2020f3net,wang2020progressive}. Although similar methods via contrast modeling for COD have achieved some favourable performance \cite{le2019anabranch,mei2021Ming,zhai2021Mutual,fan2020camouflaged, li2021uncertainty, ren2021deep}, it is still challenging because some detailed information that is necessary for fine-grained recognition and better understanding about camouflage may be ignored by the deep neural network, leading to less effective models.

\begin{figure}
    \centering
   \begin{center}
   \begin{tabular}{{c@{ } c@{ } c@{ } c@{ }}}
  {\includegraphics[width=0.23\linewidth, height=0.16\linewidth]{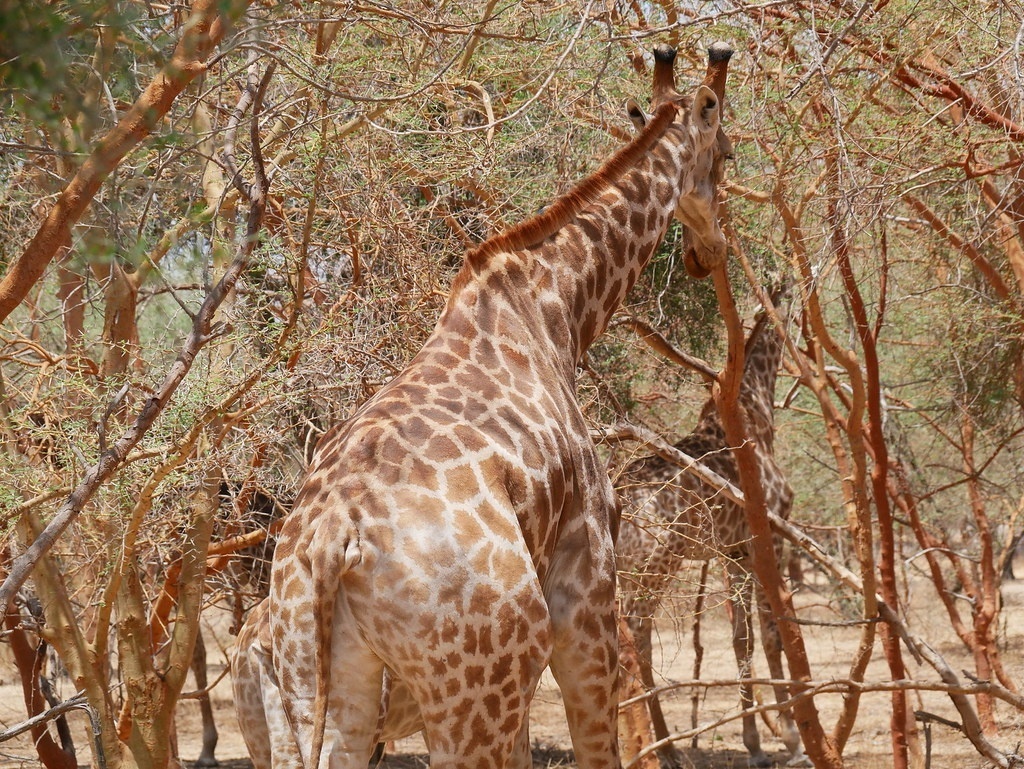}}&
    {\includegraphics[width=0.23\linewidth, height=0.16\linewidth]{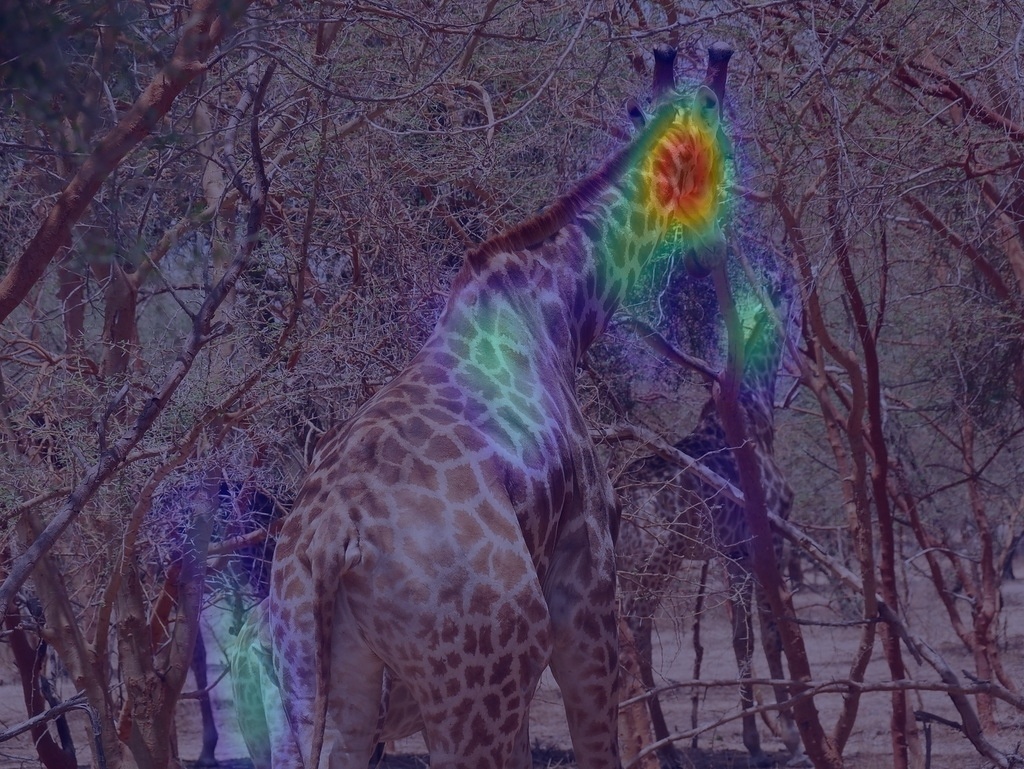}}&
    {\includegraphics[width=0.23\linewidth, height=0.16\linewidth]{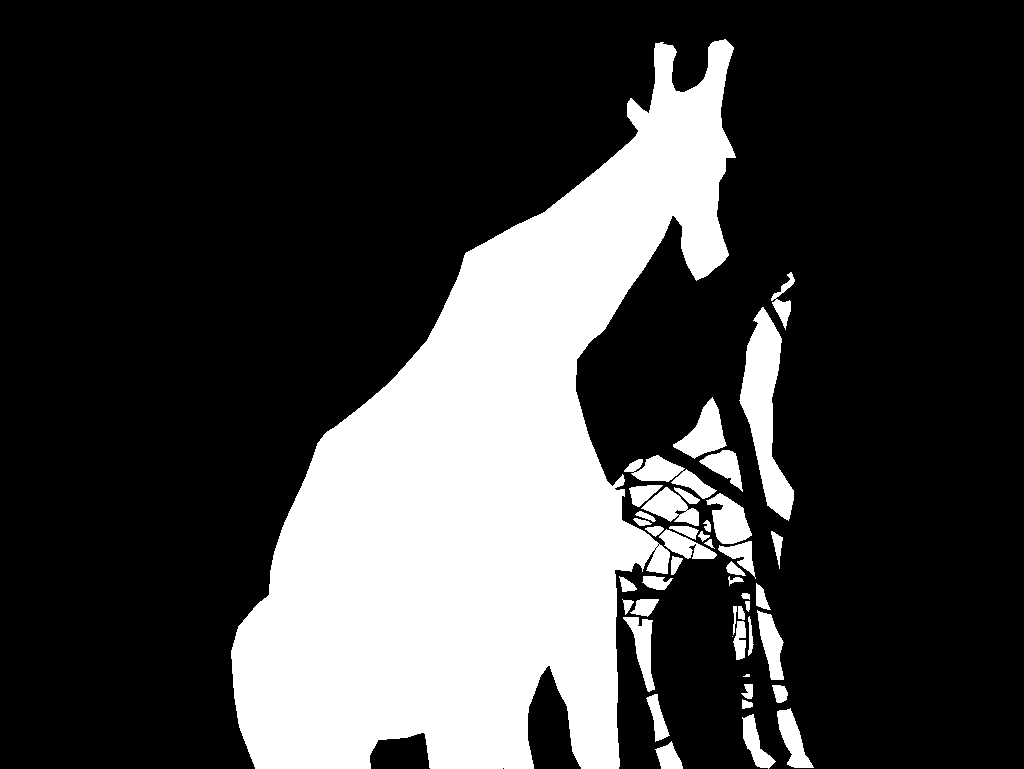}}&
    {\includegraphics[width=0.23\linewidth, height=0.16\linewidth]{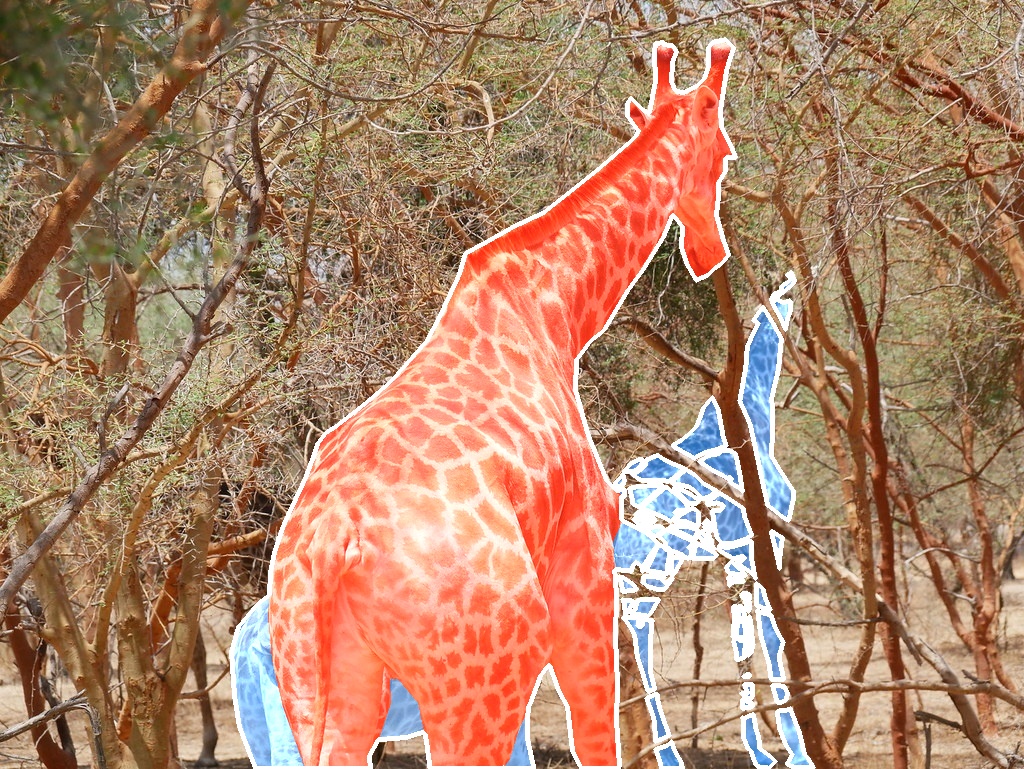}} \\
   {\includegraphics[width=0.23\linewidth, height=0.16\linewidth]{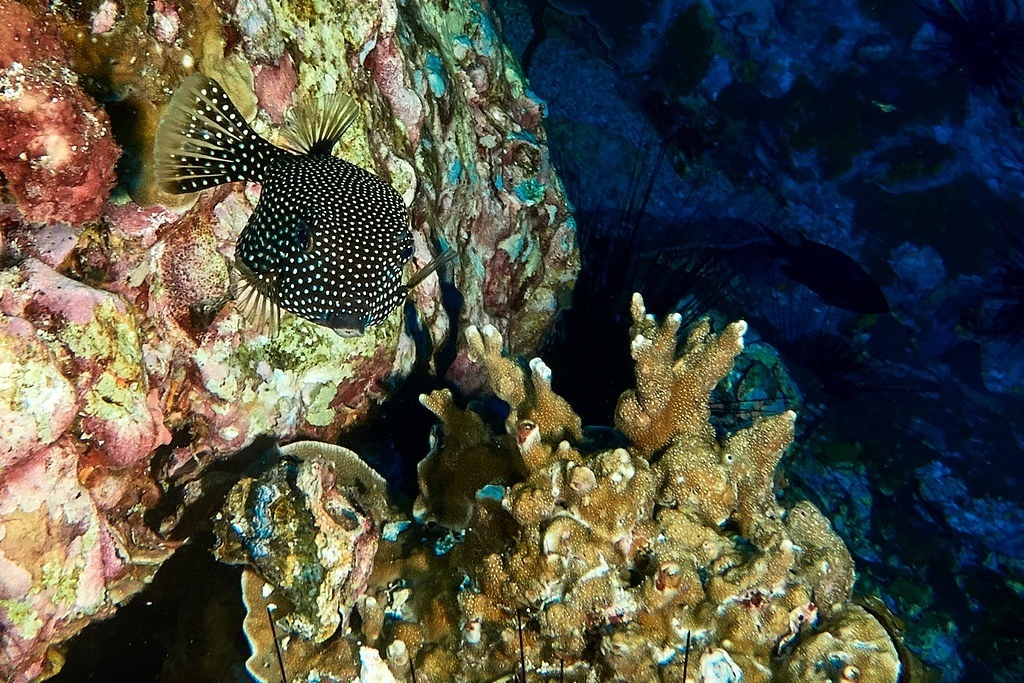}}&    
   {\includegraphics[width=0.23\linewidth, height=0.16\linewidth]{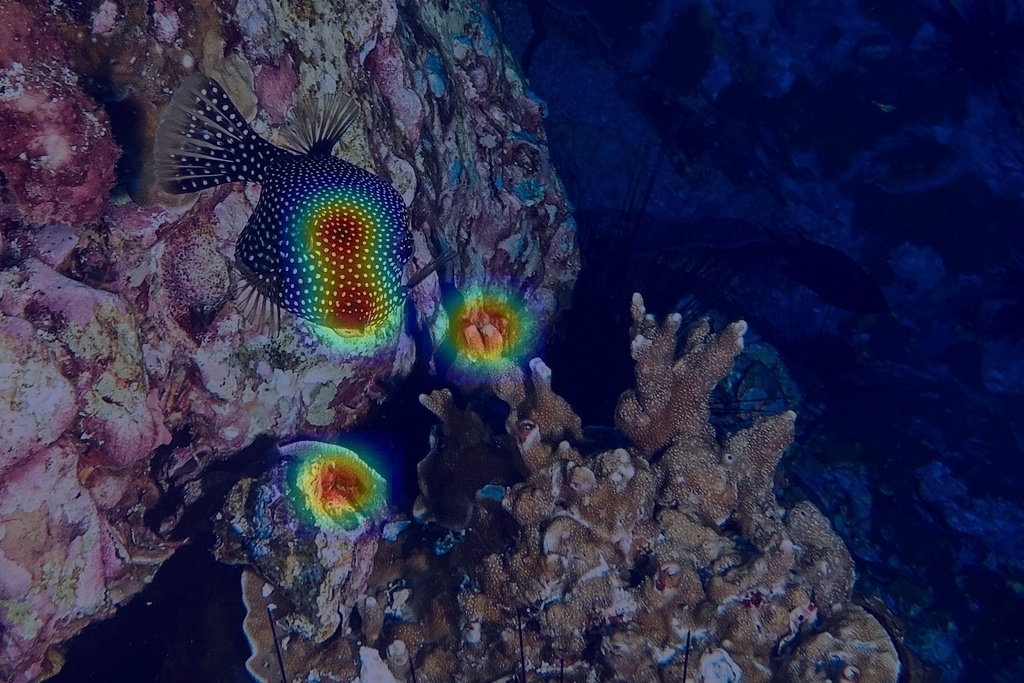}}&
    {\includegraphics[width=0.23\linewidth, height=0.16\linewidth]{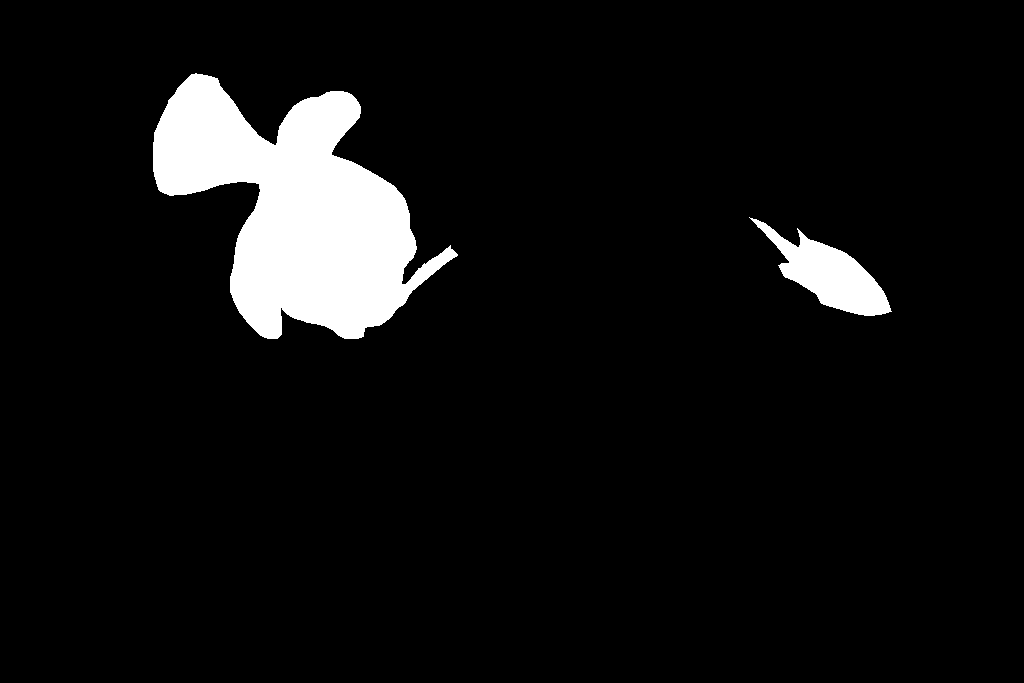}}&
    {\includegraphics[width=0.23\linewidth, height=0.16\linewidth]{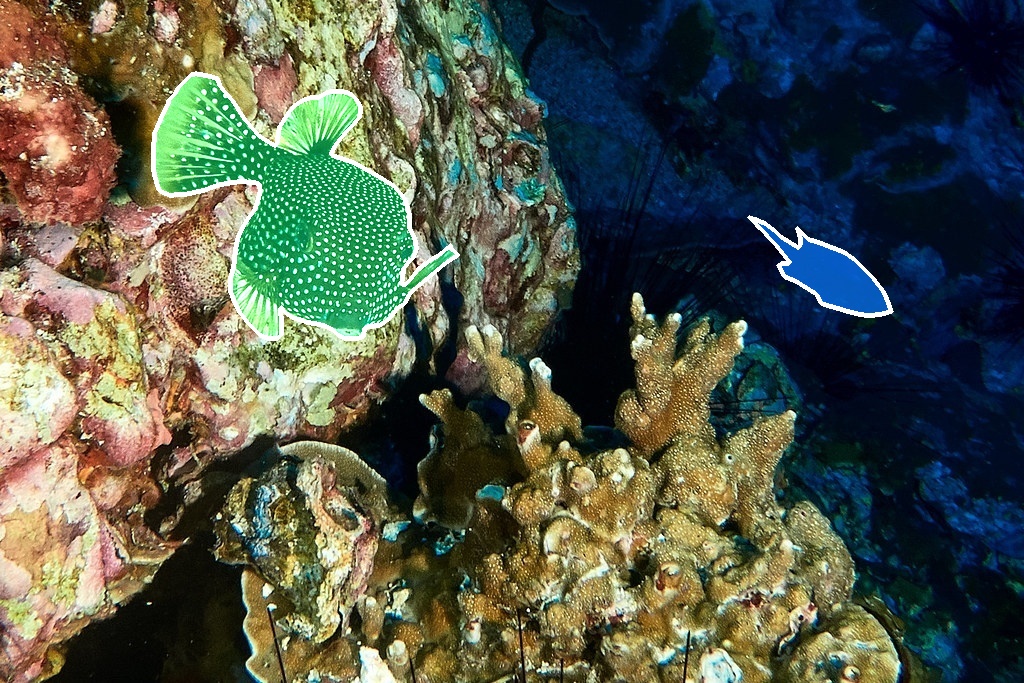}} \\
    \footnotesize{Image} &
    \footnotesize{Fixation} & \footnotesize{Binary} & \footnotesize{Ranking} \\
   \end{tabular}
   \end{center}
    \caption{
    We present the extra
    fixation (\enquote{Fixation}) and ranking (\enquote{Ranking}) annotations (blue, green, and red colors indicate hard, median, and easy levels of camouflage). \Rev{In ``Fixation'', the regions with higher level of red mean they attract more human attention while the regions with higher level of blue indicate less attention. }
    } 
    \label{fig:sample_show_front_page}
\end{figure}

The existing camouflaged object detection models \cite{mei2021Ming,zhai2021Mutual,li2021uncertainty,fan2020camouflaged,ren2021deep,chen2022camouflaged,sun2021c2fnet,bi2021rethinking} are designed based on 
binary camouflage ground truth maps as shown in Fig.~\ref{fig:sample_show_front_page}
(\enquote{Binary}), which only capture the scope of the camouflaged objects without fully exploring the basic mechanism of camouflage. We argue that this task setting fails to provide insights into some basic understanding about camouflage, \eg~\textit{1) how people find the camouflaged objects? Are they noticed as a whole or through some specific parts? 2) how to measure the degree of camouflage? Can we estimate the difficulty of finding each camouflaged instance?} With those two important questions, we find that
comparing with 
existing binary segmentation setting, localizing the discriminative regions of 
prey and estimating their camouflage degree
is more meaningful in understanding the evolution of animals.

We define the \enquote{discriminative regions} of camouflaged objects as the parts whose patterns are difficult to conceal, such as patterns with different colors to the surroundings in background matching and low contrast body outlines in disruptive coloration. The localization of the discriminative region can not only provide salient signals of the camouflaged objects,
but also offer an indicator for quantifying the conspicuousness of the camouflage.
To better explore human visual perception,
we propose to reveal the most detectable region of 
camouflaged objects, which is termed as \enquote{Camouflaged Object
Localization} (COL). Further, animals evolve to become more camouflaged to survive~\cite{cuthill2005disruptive}.
Therefore, knowing the level of camouflage is essential in
evolution biology and perception psychology. It is also beneficial to design 
sophisticated camouflaged patterns by evaluating the detectability of 
camouflaged objects. In this paper,
we define the level of camouflage as camouflage ranking, and claim that it is helpful to reveal the detectability of the camouflaged object to better understand camouflage strategies.
Accordingly, we intend to generate the instance-level ranking-based camouflaged object prediction, representing camouflage levels of 
prey, and term it 
\enquote{Camouflaged Object Ranking} (COR). 

As there exist no corresponding datasets for the two proposed new tasks, we
use an eye tracker to re-label an
existing COD dataset \cite{le2019anabranch,fan2020camouflaged} and generate a fixation dataset for the COR task.
As shown in the second column of Fig.~\ref{fig:sample_show_front_page}, the fixation map is employed to indicate the discriminative regions of the camouflaged object, which provides a better understanding about how the camouflaged object is noticed.
Then we assume that a longer observation time suggests the higher 
degree of camouflage. Following this assumption, we compute the detection delay\footnote{We define the median time for multiple observers to notice each camouflaged instance as the detection delay for this instance.} of each camouflaged instance by using the recorded time of each fixation point, and obtain the ranking dataset for our new COR task.
The ranking annotation is shown as the last column of Fig.~\ref{fig:sample_show_front_page}, which explains the level (or rank) of the camouflaged object.

\begin{figure}[!htp]
   \begin{center}
   \begin{tabular}{c@{ }}
   {\includegraphics[width=0.95\linewidth]{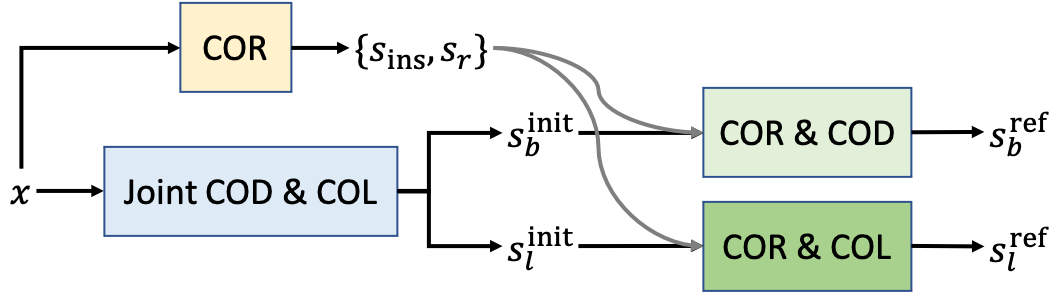}}\\
   \end{tabular}
   \end{center}
   \caption{Overview of the proposed network. We have two main tasks within our framework, namely 
   camouflaged object ranking (\enquote{COR}) which is supervised by the ranking ground truth and each rank-based binary segmentation map, and a joint learning framework for camouflaged object segmentation and discriminative region localization (\enquote{Joint COD \& COL}). With the input image, the joint learning framework produces initial predictions for the related two tasks (\enquote{$s_b^{\text{init}}$} and \enquote{$s_l^{\text{init}}$} for COD and COL respectively), which are then refined with the COR prediction (\enquote{$\{s_{\text{ins}},s_r\}$}) to generate our final predictions $s_b^{\text{ref}}$ and $s_l^{\text{ref}}$ for COD and COL.
   }
\label{fig:network_overview}
\end{figure}

This paper is an extension of our conference paper \cite{yunqiu_cod21}. Comparing with the original version, we first enlarge the fixation and ranking dataset.
In \cite{yunqiu_cod21}, we labeled 2,280 benchmark training images with three types of annotations, including 2,000 images for training and 280 images for testing. With our current version, we label the whole camouflage training dataset with 
multiple types of annotations, leading to 4,040 images in total. Further, we provide fixation and ranking annotations for the benchmark COD10K testing dataset \cite{fan2020camouflaged}, leading to another 2,026 images with the three types of annotations.
Secondly, we extensively analyse the relationship between the three tasks, and provide both a new baseline for each individual task, and a new benchmark with a triple-task learning framework.
Thirdly, we comprehensively discuss the contribution of each module and each task within our framework for better understanding of the superiority of our network. 
Fourthly, instead of treating the camouflaged object ranking task as an independent auxiliary task as in \cite{yunqiu_cod21}, we explore its contribution to the other two tasks (see Fig.~\ref{fig:network_overview}), leading to more effective triple-task relationship modeling.

\section{Related Work}

\noindent\textbf{Discriminative region localization technique:} Discriminative regions \cite{zhou2016learning} are those
leading to accurate classification, \eg~the head of animals, the lights of cars, \etc.
Zhou \etal \cite{zhou2016learning} introduced the class activation map (CAM) to estimate the discriminative region of each class,
which is the basis of many weakly supervised methods~\cite{ahn2018learning,wei2018revisiting,huang2018weakly,wei2017object,lee2019ficklenet,souly2017semi,wang2017learning}. Selvaraju \etal \cite{selvaraju2017grad} extended CAMs by utilizing the gradient of the class score w.r.t. activation of the last convolutional layer of CNN to investigate the importance of each neuron.
Chattopadhay \etal \cite{chattopadhay2018grad} used a linear combination of positive gradients w.r.t. activation maps of the last convolutional layer to capture the importance of each class activation map for the final classification. Zhang \etal \cite{zhang2018adversarial} erased the high activation area iteratively to force a CNN to learn all relevant features and therefore expanded the discriminative region.  

Similar to existing discriminative region localization techniques, we introduce the first camouflaged object discriminative region localization method to reveal the most salient region of camouflaged objects.

\noindent\textbf{Camouflaged object detection:} Camouflage is a useful technique for animals to conceal themselves from visual detection by others \cite{merilaita2017camouflage, troscianko2009camouflage}. In early research, some methods use hand-crafted features and discover the camouflaged object by means of the contrast between the foreground and the background. Tankus \etal \cite{tankus2001convexity} used an operator to represent the 3D structure and detected curved objects on a relatively smooth background. Bhajantri \etal \cite{bhajantri2006camouflage} employed the co-occurrence matrix based texture features to detect the camouflaged defect. Xue \etal \cite{xue2016camouflage} and Pike \etal \cite{pike2018quantifying} modeled visual saliency features such as color, orientation, luminance and corner point, to measure the conspicuousness of an object.
Recent research resorts to deep learning to recognize more complex properties of the camouflaged object. Among those, Le \etal \cite{le2019anabranch} introduced the joint image classification and camouflaged object segmentation framework. Yan \etal \cite{yan2020mirrornet} presented an adversarial segmentation stream using a flipped image as input to enhance the discriminative ability of the main segmentation stream for COD. Fan \etal \cite{fan2020camouflaged} proposed SINet with multiple receptive fields to find candidate regions and obtain the precise location by an identification module. Ren \etal \cite{ren2021deep} designed a texture-aware refinement module that emphasizes the texture difference between the foreground and the background and used a boundary-consistency loss to enhance the edge information. Dong \etal \cite{dong2021towards} claimed that the context information and effective multi-scale feature fusion are essential in COD. Therefore, they designed two corresponding modules to enlarge the receptive field and aggregate the multi-level features.


\noindent\textbf{Camouflage ranking and ranking-based models:}
To a certain extent, the effectiveness of the camouflage determines the difficulty of detection. Additionally, it is also a useful cue for optimizing the design of camouflage patterns. Therefore, many algorithms are formulated to quantify the camouflage quality. A prevalent technique is to extract the image features and evaluate the consistency between the object and the background. Bian \etal \cite{bian2010fuzzy} leveraged edge information to provide the degree of concealment of the object. 
Huang \etal \cite{huang2011new} used the gray spatial distribution of the edges of the foreground and the background to evaluate the camouflage performance. Song \etal \cite{song2010new} proposed to select some structural features such as illumination, texture direction and edges and used the weight structural texture similarity to evaluate the effects of camouflage texture. Xue \etal \cite{xue2016camouflage} and Pike \etal \cite{pike2018quantifying} also selected some features manually, but they quantified 
camouflage quality by computing the visual saliency maps that try to highlight the conspicuous visual signals. Troscianko \etal \cite{Quantifyingcamouflage} introduced a metric based on angle-sensitive Gabor filters to measure the proportion of the false edges to the coherent edges around the outline.
However, 
methods based on the manual-selected and manual-designed features are not sufficient to reveal the actual mechanism of viewers perceiving the camouflaged object. Instead, our camouflage ranking task aims to automatically learn the features determining camouflage rank of the object from the ranking dataset collected by human observations.

As far as we know, there exists no deep camouflage ranking model. We find another type of ranking models exists, namely saliency ranking, aiming to evaluate the
conspicuousness of the salient object.
Islam \etal \cite{amirul2018revisiting} argued that saliency is a relative concept when multiple observers are queried. 
To investigate this, they collected a saliency ranking dataset based on the PASCAL-S dataset \cite{pascal-s} with 850 images labeled by 12 observers. Based on this dataset, they designed
an encoder-decoder to predict saliency masks of different levels to achieve the final ranking prediction. Similarly,
Yildirim \etal \cite{yildirim2020evaluating} evaluated salient ranking based on the assumption that objects in natural images are perceived to have varying levels of importance. They also use some images from MSCOCO \cite{lin2014microsoft} to construct a new saliency ranking dataset, namely COCO-SalRank \cite{kalash2019relative}, according to the density of fixations provided by SALICON \cite{jiang2015salicon}. Siris \etal \cite{siris2020inferring} defined ranking by inferring the order of attention shift.
Their dataset is also based on the fixation data from
SALICON \cite{jiang2015salicon}.

We argue that the formulation of saliency ranking is significantly different from that of the camouflage ranking. Although both tasks give 
relative ranking, saliency ranking is relative within a single image while 
camouflage ranking is relative throughout the entire dataset. In other words, a higher degree
of saliency means that the corresponding object is more attractive to the observer in an image whereas the object with a higher rank of camouflage indicates a lower probability of being noticed.

\section{Our Method}
We introduce two biologically inspired camouflage related tasks, namely camouflaged object localization (COL) and camouflaged object ranking (COR). We then present a triple-task learning pipeline to simultaneously localize, segment, and rank 
the camouflaged objects as shown in Fig.~\ref{fig:network_overview}. Given the training dataset $D=\{x^i,y_b^i,y_l^i,y_r^i\}_{i=1}^N$, where $x^i$, $y_b^i$, $y_l^i$ and $y_r^i$ are the RGB image, the corresponding binary camouflage segmentation map, the localization map, and the ranking map, and $i$ indexes the images, while $N$ is the size of the training dataset. The goal of the proposed triple task learning framework is to estimate the camouflaged segmentation map, the localization map, and the ranking map simultaneously for the input RGB image.

\subsection{Introduction to the New Tasks}
\noindent\textbf{Camouflaged Object Localization \Rev{(COL)}:}
We define the \enquote{discriminative regions} as regions that make the camouflaged object apparent.
These regions are essential for the observer to identify the camouflaged object and in turn many camouflage strategies are committed to distracting the attention of the observers from these regions. Consequently, the localization of discriminative regions could not only help 
camouflaged object detection, but also function as 
guidance for
camouflage design to avoid 
detection.
Compared with other regions of the camouflaged object, the discriminative region should have a higher \enquote{contrast} with its surroundings. To discover those regions,
we use an eye tracker to record human attention towards the RGB image with the camouflaged instance.
In this way, the region with intense human attention represents the discriminative or high contrast region that makes the camouflaged object noticeable.

\begin{figure}[t!]
   \begin{center}
   \begin{tabular}{c@{ } c@{ } c@{ } c@{ } c@{ }}
      {\includegraphics[width=0.18\linewidth, height=0.14\linewidth]{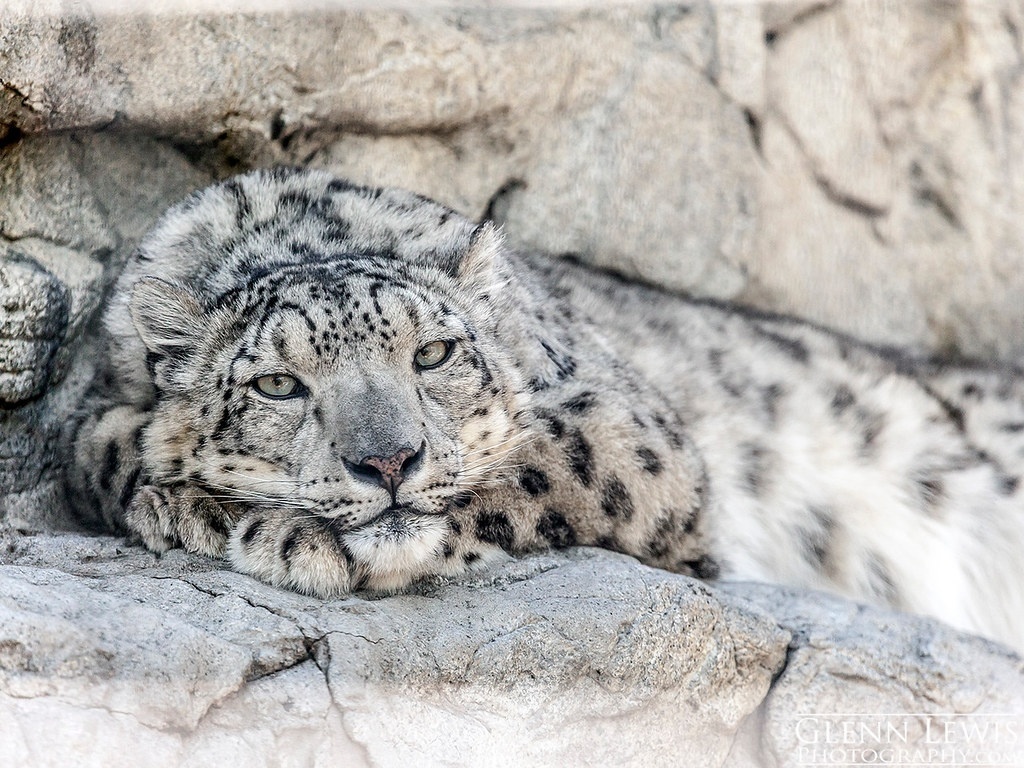}}&
      {\includegraphics[width=0.18\linewidth, height=0.14\linewidth]{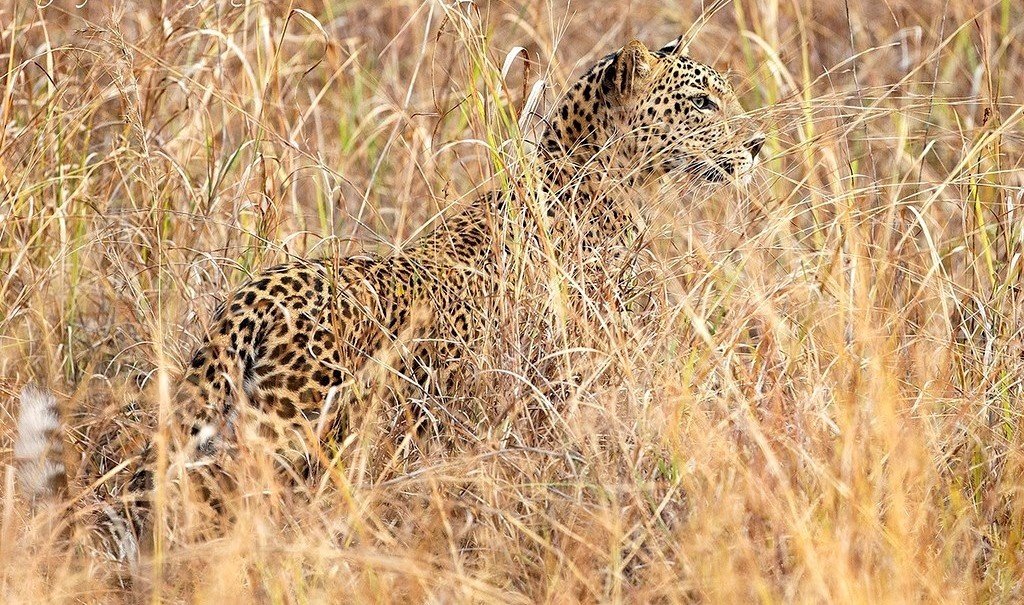}}&
      {\includegraphics[width=0.18\linewidth, height=0.14\linewidth]{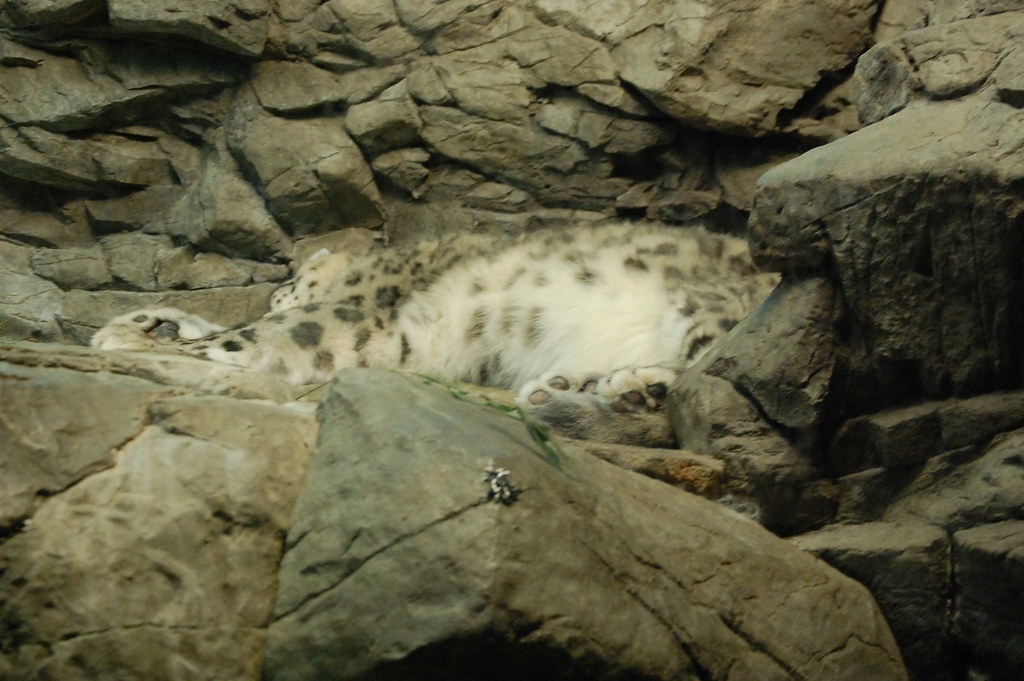}}&
      {\includegraphics[width=0.18\linewidth, height=0.14\linewidth]{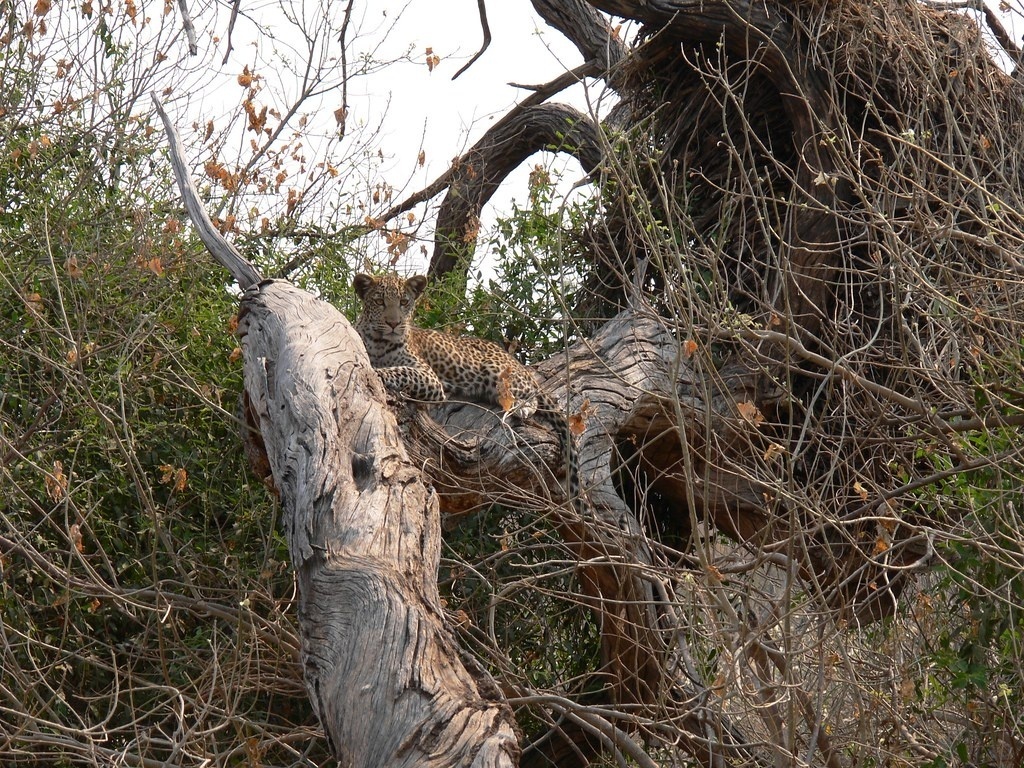}}&
      {\includegraphics[width=0.18\linewidth, height=0.14\linewidth]{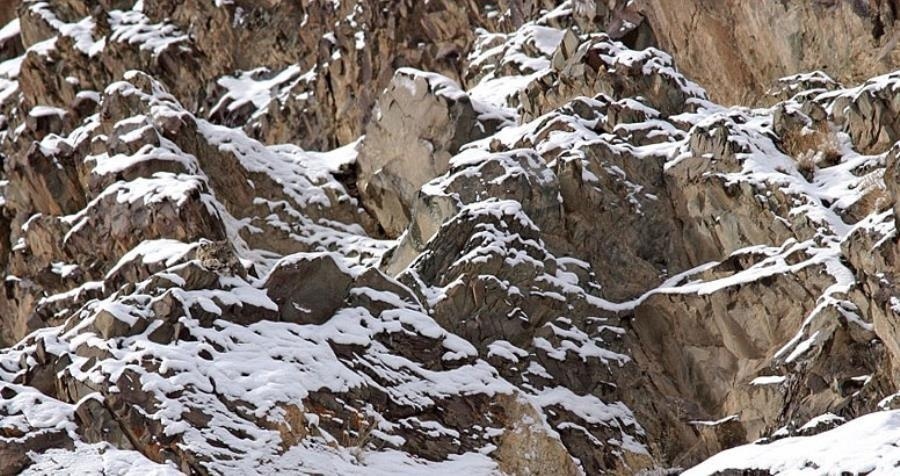}}\\
      {\includegraphics[width=0.18\linewidth, height=0.14\linewidth]{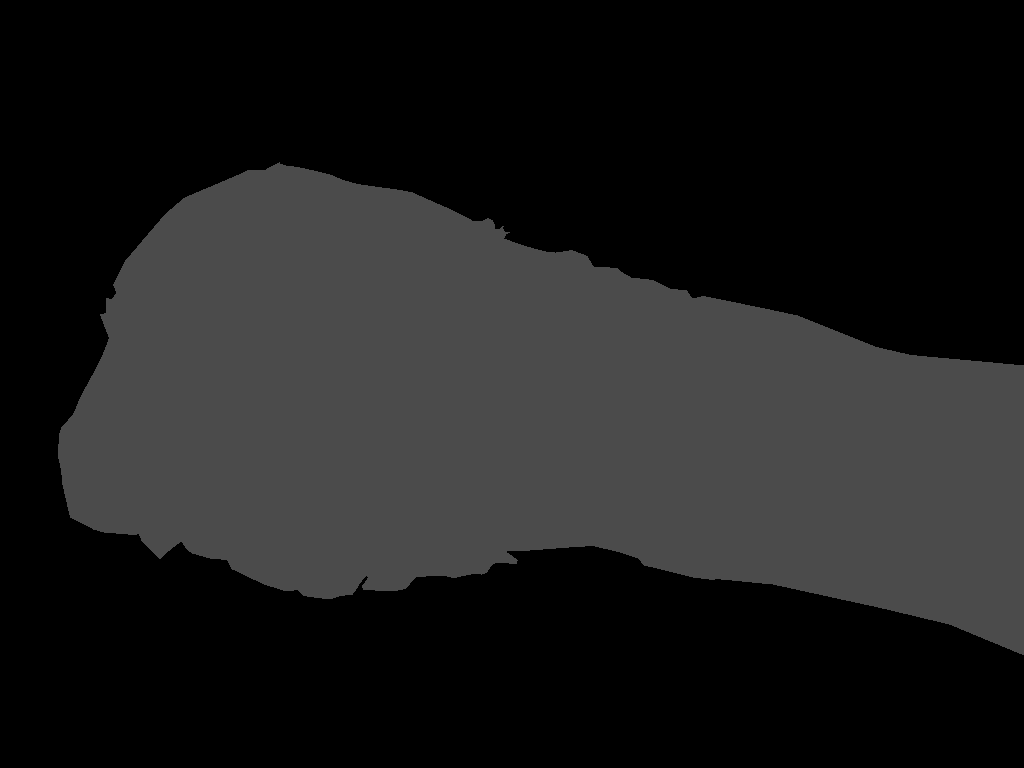}}&
      {\includegraphics[width=0.18\linewidth, height=0.14\linewidth]{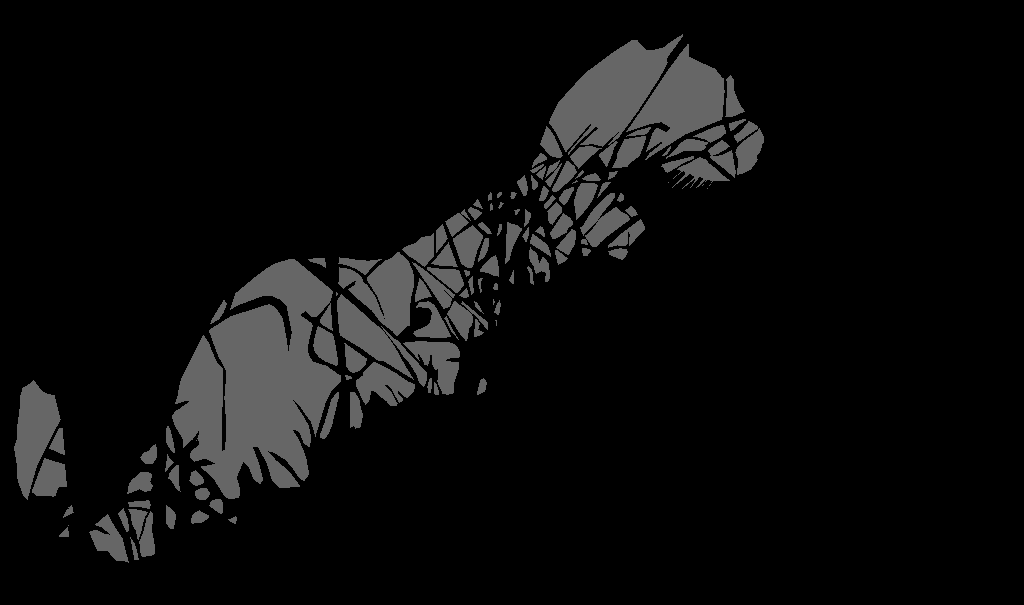}}&
      {\includegraphics[width=0.18\linewidth, height=0.14\linewidth]{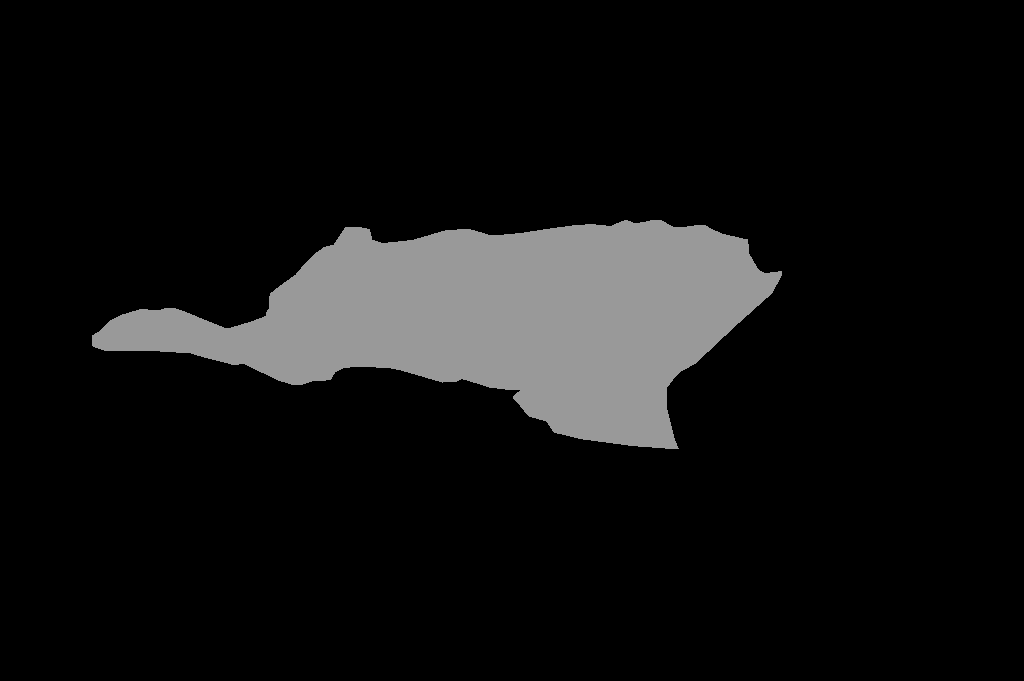}}&
      {\includegraphics[width=0.18\linewidth, height=0.14\linewidth]{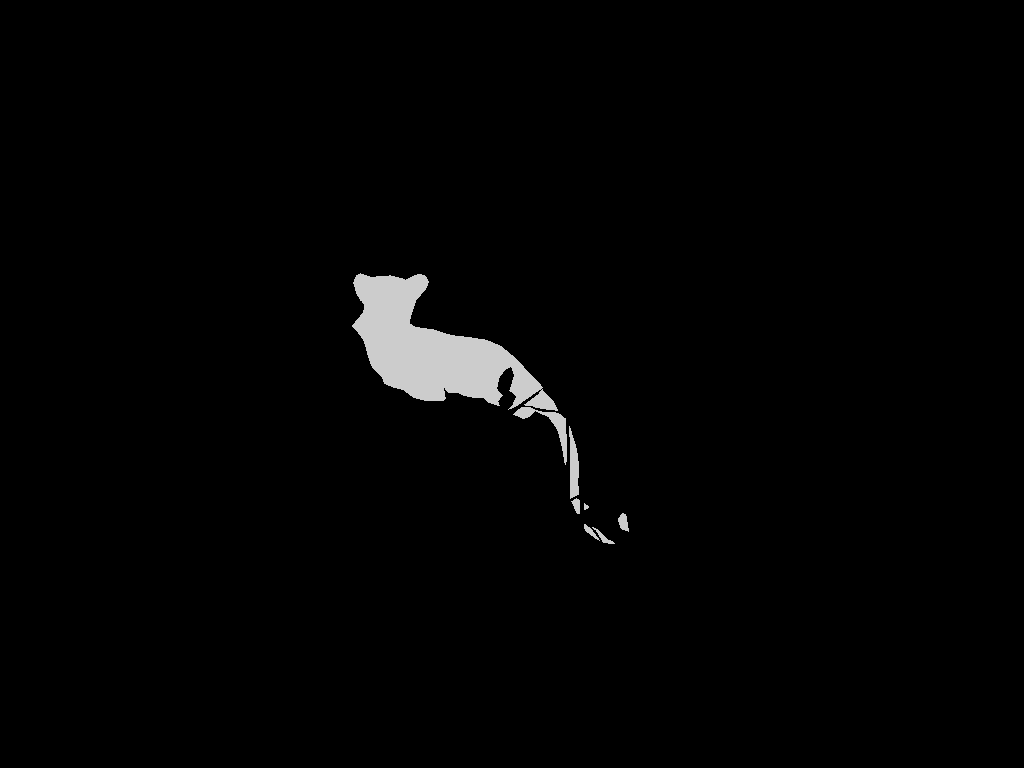}}&
      {\includegraphics[width=0.18\linewidth, height=0.14\linewidth]{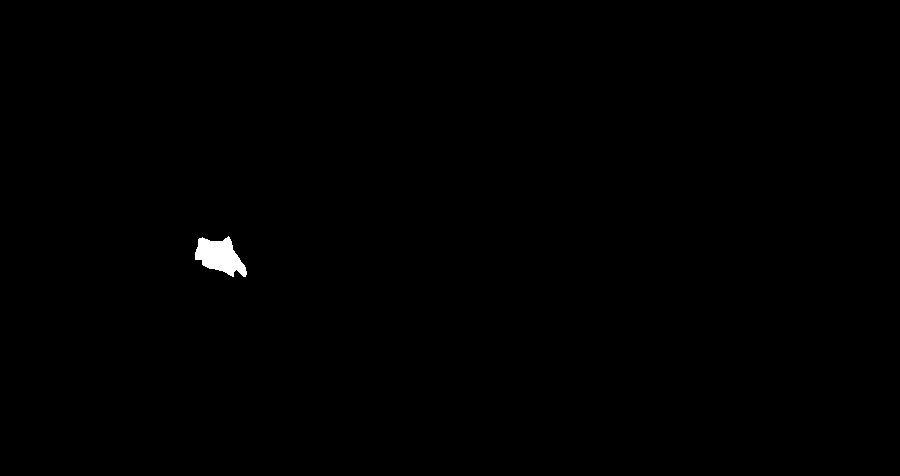}}\\
      \footnotesize{Easy} & \footnotesize{Median1} & \footnotesize{Median2} &
      \footnotesize{Median3} & \footnotesize{Hard} \\
   \end{tabular}
   \end{center}
   \caption{The ranking dataset demonstrates the varying degrees of difficulty in detecting 
   camouflaged instances. To better illustrate the definition of camouflage ranking, we select the images in the category \textit{Leopard} (first row), with different levels of camouflage. The second row indicates the camouflaged regions and different colors are used to indicate the camouflage levels.}
\label{fig:ranking_gt}
\end{figure}

\noindent\textbf{Camouflaged Object Ranking \Rev{(COR)}:}
We introduce \enquote{Camouflaged Object Ranking (COR)} to evaluate the ability of camouflaged objects to conceal themselves in the environment. The difficulty of camouflaged objects being detected varies due to the different camouflage strategies. For example, some imitations of the color and pattern are clumsy since the discriminative regions are still salient for
observers, while some animals generate more delicate disruptive patterns to make the object hard to identify.
In order to quantify the effectiveness of camouflage, it is straightforward to assume that the object using a more effective strategy tends to be less salient to the observer. We assume that the longer it takes for the observers to find the camouflaged instance, the harder the camouflaged instance. With an eye tracker for the camouflaged object localization map acquisition, we have access to the observation time for each camouflaged instance, which is then used to determine the
rank or level of the camouflaged object. In this paper, we define five levels of difficulty for a camouflaged instance, namely easy (ES), medium1 (M1), medium2 (M2), medium3 (M3), and hard (HD) as shown in Fig.~\ref{fig:ranking_gt}. We argue that the ranking annotation
provides additional understanding about
the degree of difficulty
of each camouflaged instance \cite{merilaita2017camouflage}.

\subsection{Task Relationship Modeling}
\label{sec:task_relationship_modeling}
We aim to train a triple-task learning framework to simultaneously localize, segment, and rank
camouflaged objects. We argue that the effectiveness of the multi-task learning pipeline lies in the inner correlation of the multiple tasks.

\noindent\textbf{COL \& COD:} As discussed before, the discriminative regions serve as an important indicator for the observer to find the camouflaged object. Based on \cite{lin2014evaluating}, we can conclude that the camouflage design confuses 
participants in two ways. Firstly, it decreases the contrast between the object and background to reduce the detectability. Secondly, it prevents 
discrimination by equipping the object with complicated features to distract 
focus. The binary camouflage map indicates the full scope of the camouflaged object without differentiating the difference of discriminativeness. The camouflage localization map then provides the most salient regions of the camouflaged object or the parts that aid recognition, 
which highlight the discriminative region of the camouflaged instance. Therefore, it could be employed directly to check the existence of 
camouflaged objects \cite{pike2018quantifying,skurowski2018evaluation,toet2020review}. Further, the discriminative regions can
help the detection of the whole camouflaged object because the other regions within a
camouflaged object are related with discriminative regions semantically. In this way, the discriminative regions could be employed as effective seeds for camouflaged object detection. 



\noindent\textbf{COL \& COR:} Eye movement data has always been a favorable tool to measure the effectiveness of camouflage. Existing literatures \cite{chang2012visual,lin2014developing,lin2014evaluating} show that the detection hit rate (the percentage of fixations for which
the target was corrected detected), the number of fixations on display, and the number of fixation on target are strongly correlated with the effectiveness of camouflage design. For a challenging camouflaged object, a global search is needed to identify it, which leads to widely distributed human attention, or fixation map. On the other hand, a camouflage localization map with clear attention in a compact area indicates an easier camouflaged instance, which explains the relationship between COL and COR.

\begin{figure}[t!]
   \begin{center}
   \begin{tabular}{c@{ }}
   {\includegraphics[width=0.95\linewidth]{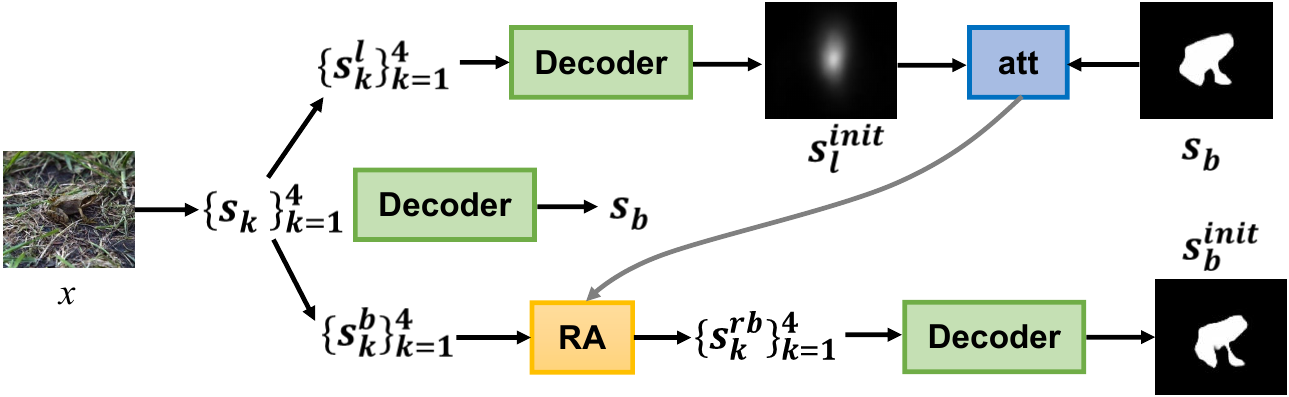}}\\
   \end{tabular}
   \end{center}
   \caption{\Rev{Overview of the residual learning based joint segmentation and localization
   network, where \enquote{RA} represents the \enquote{reverse attention} operation, and \enquote{Decoder} is adopted from \cite{Ranftl2020}. Note that the three \enquote{Decoder}s share the same network structure but not weights.}
   }
\label{fig:joint_cod_fixation}
\end{figure}


\noindent\textbf{COR \& COD:} Animals are evolved to be camouflaged \cite{troscianko2018camouflage} in order to prevent being noticed by their predators. With knowledge about the difficulty level of each camouflaged instance, we can design more sophisticated camouflaged object detection strategies to find the intrinsic characteristics of 
camouflaged objects for effective camouflaged object detection. In this way, hard-negative mining \cite{focal_loss} related strategies can be adopted to effectively explore 
prior knowledge about the ranking of each camouflaged instance for accurate camouflaged object detection.

\subsection{Triple-task Learning Framework}
We intend to achieve camouflaged object localization, segmentation, and ranking with one unified framework. Due to the close relationship between COL and COD, we first design a residual learning based joint localization and segmentation network, then we introduce the ranking task to the dual-task learning framework
by further exploring the contribution of camouflaged object ranking to the other two tasks.

\subsubsection{Joint Localization and Segmentation}
\label{sec:joint_cod_col}
Different from COR, which involves instance level segmentation, both COD and COL can be achieved with fully convolutional neural
networks,
while the latter is usually treated as a regression task instead of a classification task.
With ground truth for both COD and COL, a straightforward solution to achieve dual-task learning is to train the two tasks with a multi-head framework similar to \cite{fixation_driven}. With the above discussion that the COL map highlights the discriminative regions of the camouflaged instance, which can be defined as easier parts to detect compared with the less discriminative regions of the camouflaged instance. Following the hard-negative learning \cite{focal_loss} pipeline, we define the other parts of the camouflaged instance as hard samples, and introduce a residual learning based joint segmentation and localization network (\enquote{Joint COD \& COL} in Fig.~\ref{fig:network_overview}) as shown in Fig.~\ref{fig:joint_cod_fixation}.



\noindent\textbf{Network Overview:}
We built our joint learning framework with a ResNet50 \cite{he2016deep} backbone as shown in Fig.~\ref{fig:joint_cod_fixation}. Given an input image $x$ (we omit the index $i$ when it is clear), we first feed it to the backbone network to obtain feature representation $\{s_k\}_{k=1}^4$, representing feature maps from different stages of the backbone network. Similar to existing ResNet50-based networks, we define a group of convolutional layers that produce the same spatial size as belonging to
the same stage of the network. Then we design the \enquote{Localization Decoder} and the \enquote{Segmentation Decoder} modules to produce the initial localization map $s_l^{\text{init}}$ and the camouflage segmentation map $s_b^{\text{init}}$ respectively, where the localization map serves as attention to generate an accurate segmentation map.


\noindent\textbf{Localization Decoder:} With $\{s_k\}_{k=1}^4$, the \enquote{Localization Decoder} regresses the localization map, which is a heatmap representing the discriminative regions of the camouflaged instance. As there exists no clear structure within the localization map, the main focus of the \enquote{Localization Decoder} is to obtain a large receptive field with effective context modeling. To achieve this, we first map the backbone feature $\{s_k\}_{k=1}^4$ to the 
\Rev{new features} $\{s^l_k\}_{k=1}^4$ of channel size $C=64$ with a multi-scale dilated convolution \cite{denseaspp} to achieve a larger receptive field for each stage of the network, where $l$ within $s^l_k$ represents the new  feature for the localization decoder. Then, we feed $\{s^l_k\}_{k=1}^4$ to a decoder \cite{Ranftl2020}, which gradually aggregates higher-level and lower-level features with residual attention to
produce a one-channel camouflage localization map $s_l^{\text{init}}$.

\noindent\textbf{Segmentation Decoder:} Different from the \enquote{Localization Decoder} which aims to produce a heatmap (the localization map as shown in Fig.~\ref{fig:sample_show_front_page}) with no structure information, the \enquote{Segmentation Decoder} is designed to generate a structure-accurate camouflage map, showing the entire scope of the
camouflaged instance(s). To achieve this,
similar to the \enquote{Localization Decoder}, we first adopt a multi-scale dilated convolution \cite{denseaspp} after the backbone feature $\{s_k\}_{k=1}^4$ to achieve a larger receptive field for each stage of the network and achieve $\{s_k^b\}_{k=1}^4$ of channel size $C=64$, representing the segmentation related feature representation. Then, we adopt the same decoder \cite{Ranftl2020} as in the \enquote{Localization Decoder}, which takes $\{s_k^b\}_{k=1}^4$ as input and produces the camouflage segmentation map $s_b$.
Note that the produced camouflage map $s_b$ is the
camouflage prediction without considering the prediction from the \enquote{Localization Decoder} branch.

\noindent\textbf{Residual Learning based Joint COD \& COL:} 
We claim that the localization map highlights the discriminative regions of the camouflaged instance, which makes the camouflaged instance noticeable. We then define the other parts of the camouflaged instance as hard samples, and introduce a residual learning based joint COD \& COL
to achieve camouflage prediction with knowledge from the localization branch.
Specifically, given the discriminative region prediction $s_l^{\text{init}}$, we obtain the reverse attention as $\text{att}=\exp{|s_b-s_l^{\text{init}}|}$,
aiming to explore the hard camouflaged area, where $s_b$ is the camouflage segmentation map from the \enquote{Segmentation Decoder}.
In this way, $\text{att}$ serves as an indicator showing the pixel-level difficulty degree of the input image $x$ for camouflaged object detection. 
Then we treat $\text{att}$ as attention for COD,
and multiply it with the COD feature representation $\{s_k^b\}_{k=1}^4$
to generate the reverse localization attention guided COD feature $\{s_k^{rb}\}_{k=1}^4$.
Finally, the
same decoder \cite{Ranftl2020} as in the \enquote{Segmentation Decoder} module is adopted, which takes $\{s_k^{rb}\}_{k=1}^4$ as input
to generate the camouflage map $s_b^{\text{init}}$ with the localization map as guidance.


\noindent\textbf{Objective Function:}
We have two types of loss function in the joint learning
framework: the discriminative region localization loss and the camouflaged object detection loss. For the former, we use the
L2 loss $\mathcal{L}_2$, and for the latter, we adopt the
pixel position aware loss $\mathcal{L}_c$ as in \cite{wei2020f3net} to produce predictions with higher structure accuracy. Then we define our final loss function $\mathcal{L}_{dual}$ for the proposed \enquote{residual learning based joint COD \& COL network}
as:
\begin{equation}
\label{dual_loss}
\mathcal{L}_{dual} = \mathcal{L}_2(s_l^{\text{init}},y_l) + \lambda(\mathcal{L}_c(s_b,y_b) + \mathcal{L}_c(s_b^{\text{init}},y_b)),
\end{equation}
where $\lambda$ is a weight to measure the importance of each task (empirically we set $\lambda=1$ in this paper).
Let's define the parameters of the \enquote{Joint COD \& COL} module as $\theta^{(1)}$, given the input image $x$, we obtain its output as: $\{s_l^{\text{init}},s_b,s_b^{\text{init}}\}=f_{\theta^{(1)}}(x)$.






\begin{figure}[t!]
   \begin{center}
   \begin{tabular}{c@{ }}
   {\includegraphics[width=0.85\linewidth]{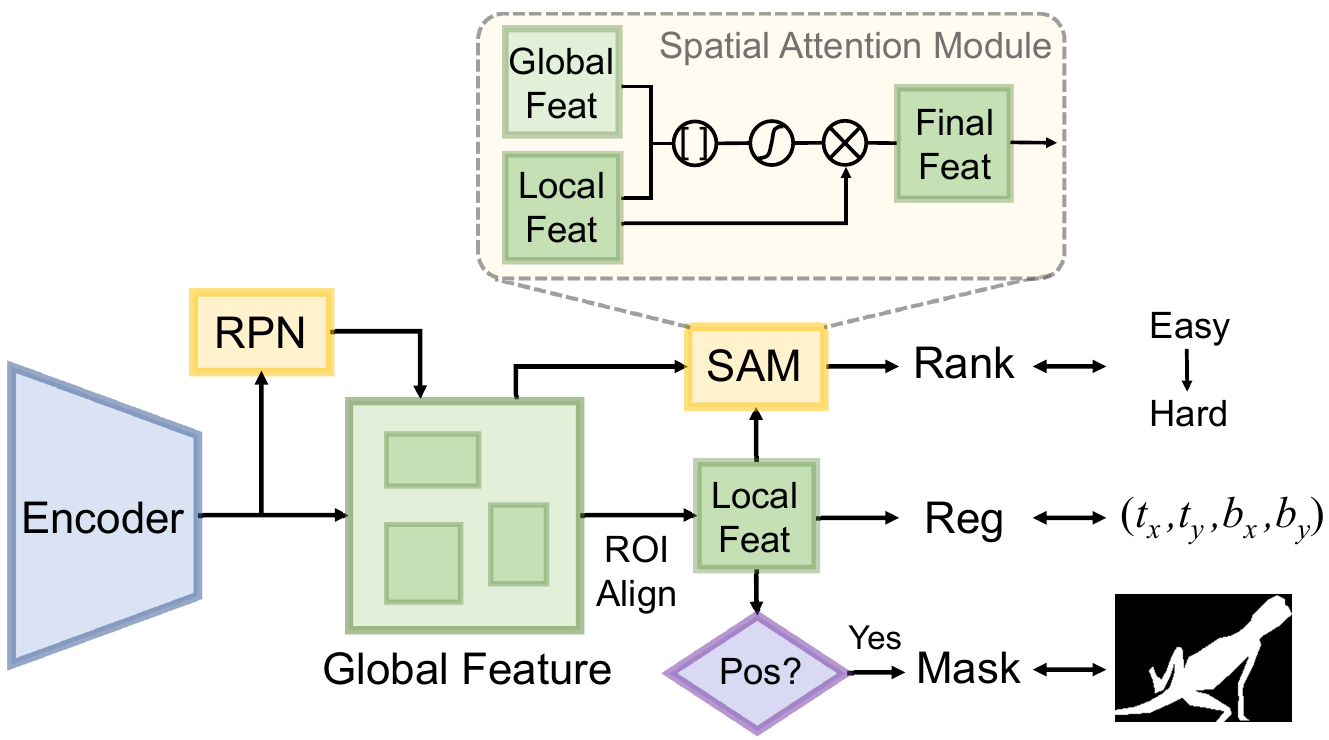}}\\
   \end{tabular}
   \end{center}
   \caption{Overview of camouflage ranking model, 
   where \enquote{RPN} represents the \enquote{region proposal network}, and \enquote{SAM} is the spatial attention module.
   }
\label{fig:cor_model}
\end{figure}

\subsubsection{Inferring the Rank of Camouflaged Instance}
\label{sec:ranking_model}
With the joint learning pipeline in Fig.~\ref{fig:joint_cod_fixation}, we achieve the camouflage localization map $s_l^{\text{init}}$ and camouflage segmentation map $s_b^{init}$.
We further introduce the camouflage ranking model (see Fig.~\ref{fig:cor_model}) as the third part of our triple-task learning framework to infer the rank or difficulty level of each camouflaged instance.

\noindent\textbf{The ranking model:}
We construct our camouflage ranking model on the basis of Mask R-CNN \cite{he2017mask} and modify its classification branch to infer the camouflage level of the detected camouflaged instance. Similar to the joint COD \& COL module, we first extract the global feature $\{s_k\}_{k=1}^4$ of the image $x$ from the pre-trained ResNet50 backbone. Then, the region proposal network (RPN) \cite{faster_rcnn} filters out the background regions to focus on
the region with a high probability of containing the \enquote{object}, which is defined as \enquote{Region of Interest} (ROI). Then the local features of the same scale $\{r_l\}_{l=1}^L$ are extracted from these ROIs by ROIAlign \cite{he2017mask}, where $L$ is the number of ROIs. 

In the original Mask R-CNN \cite{he2017mask}, the local feature is fed
into a bounding box head with a classification branch and a regression branch to infer the category and the position of each instance. If the instance belongs to one of the foreground categories,
the local feature is further fed into a mask head to generate the segmentation mask of the instance. 
However, different from the classification task, it is difficult to determine the degree of camouflage only from the appearance of the camouflaged object in the ranking task, as more context information about the camouflaged object is needed to determine its degree. We argue that the global contrast between the instance and its surrounding
significantly influences its camouflage level.
To this end, in the camouflage ranking model, we embed a spatial attention module (SAM) \cite{fu2019dual} into the classification branch in the bounding box head to model the relationship between the instance and its surroundings. The rank branch accepts both the global feature $\{s_k\}_{k=1}^4$ \Rev{corresponding to the whole image} from the backbone network and the local feature $\{r_l\}_{l=1}^L$ \Rev{corresponding to proposal bounding boxes}.
The SAM \cite{fu2019dual} first generates pooled features $s_{max}$ and $r_{avg}$ by operating global average pooling on the resized global feature and
local feature, respectively. Then, we concatenate the global and local features and embed the output feature with a $3\times 3$ convolutional layer. The final feature $\hat r$ can be obtained by multiplying the local feature with Sigmoid activation of the embedded feature via:
\begin{equation}
\hat r_l = r_l*\sigma(\text{Conv}3\times 3(\text{Concat}(s_{max}, r_{avg}))),
\end{equation}
where $\text{Conv}3\times 3$ indicates a $3\times3$
convolutional layer,
$\text{Concat}$ is the concatenate operation and $\sigma(\cdot)$ represents the Sigmoid function. The ranking model can then be trained with loss function from the original Mask-RCNN \cite{he2017mask} as:
\begin{equation}
\label{mask_rcnn_loss}
    \mathcal{L} = \mathcal{L}_{rpn}+\mathcal{L}_{bbox}+\Rev{\mathcal{L}_{rank}}+\mathcal{L}_{mask},
\end{equation}
where $\mathcal{L}_{rpn}$ is the loss function for \Rev{RPN\cite{faster_rcnn}}, \Rev{$\mathcal{L}_{bbox}$\cite{he2017mask}} is designed for the bounding box regression head, which includes the L1 loss \Rev{$\mathcal{L}_{reg}$\cite{he2017mask}} for regression and Cross Entropy loss $\mathcal{L}_{rank}$ for ranking, \Rev{$\mathcal{L}_{mask}$\cite{he2017mask}} is the pixel-wise cross-entropy loss for the segmentation task.

\begin{figure}[t!]
   \begin{center}
   \begin{tabular}{c@{ }}
   {\includegraphics[width=0.85\linewidth]{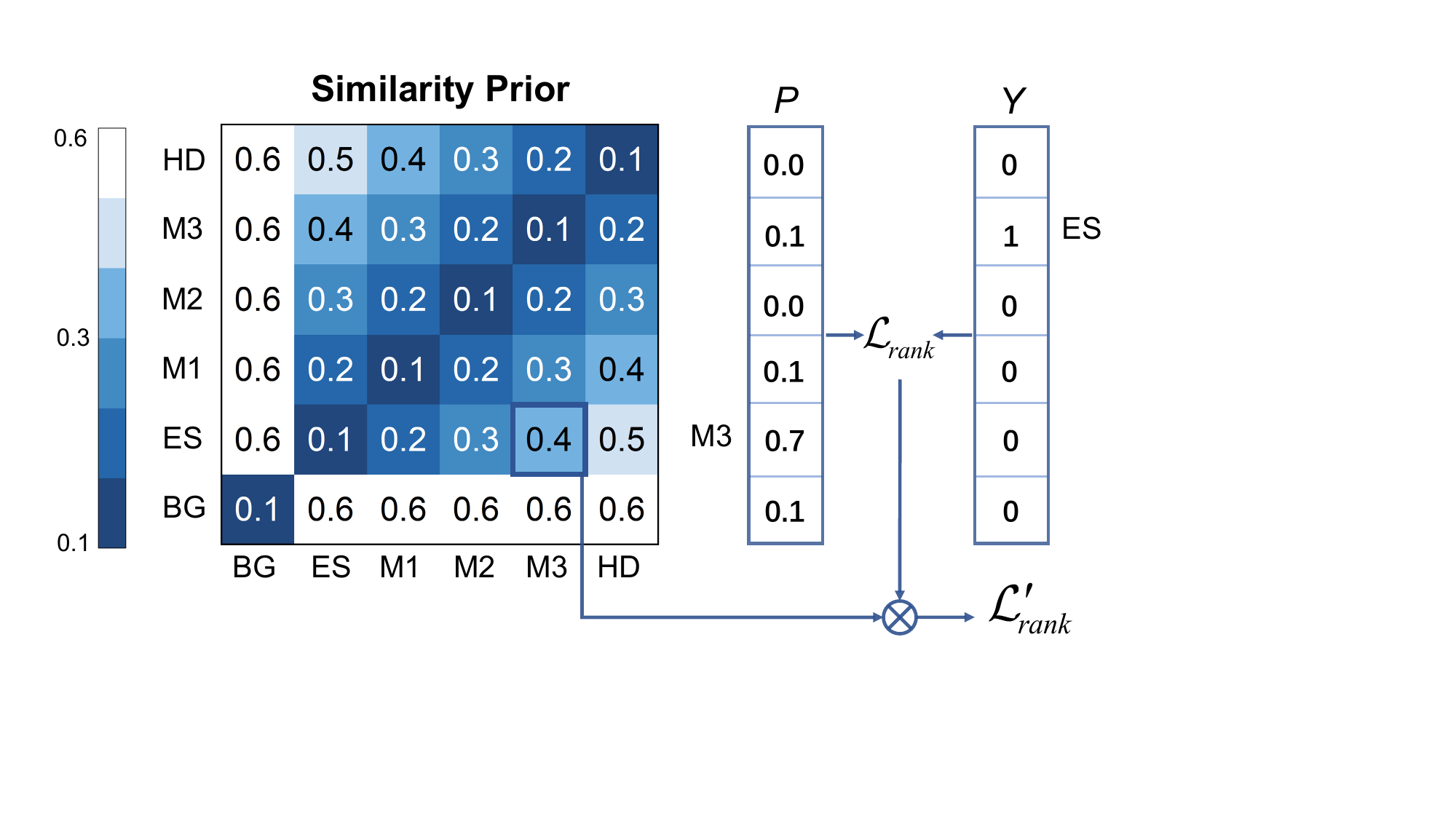}}\\
   \end{tabular}
   \end{center}
   \caption{Label similarity as a prior to consider the rank label dependency of our ranking dataset. $\mathrm{P}$ and $\mathrm{Y}$ denote the prediction and the one-hot ground truth, respectively.
   }
\label{fig:similarity_prior}
\end{figure}




\noindent\textbf{Label similarity as prior:}
Directly inferring ranks of camouflaged instances with Mask-RCNN \cite{he2017mask} may produce unsatisfactory results due to the independence of labels in the instance segmentation dataset.
In our ranking scenario, the ranks are progressive. More specifically,
because the ranks of camouflage degree are Easy (ES), Medium1 (M1), Medium2 (M2), Medium3 (M3), Hard (HD) in order of low to high, 
the instance of a sample of \enquote{ES} rank should be penalized more if it is misclassified as \enquote{HD} rank instead of a sample of \enquote{M1} rank. In the instance segmentation task, however, the penalty is the same regardless of which category the sample is misclassified into.
We intend to employ a constraint on $\mathcal{L}_{rank}$ in Eq.~\ref{mask_rcnn_loss} for more effective camouflage ranking model. Hence, a camouflaged instance similarity prior $w_p$ is defined, which is a $6\times6$ matrix as shown in Fig.~\ref{fig:similarity_prior}. Each $w_p(m,n)$ represents the penalty for
predicting the $n$-th rank as the $m$-th rank. Given the prediction of the instance classification network in Fig.~\ref{fig:network_overview}, and the ground truth instance rank, we first compute the original rank loss $\mathcal{L}_{rank}$ (before we compute the mean of $\mathcal{L}_{rank}$ in the mini-batch). Then, we assign weight to it with the specific similarity prior $w_p(m,n)$. As is illustrated in Fig.~\ref{fig:similarity_prior}, the predicted rank is \enquote{M3}, and the ground truth rank is \enquote{ES}, then we get penalty $w_p(5,2)=0.4$, and multiply it with the original rank loss $\mathcal{L}_{rank}$ to obtain the weighted loss $\mathcal{L}'_{rank}$.

Let's define the parameters of the camouflage ranking module (\enquote{COR} in Fig.~\ref{fig:network_overview}) as $\theta^{(2)}$, given the input image $x$, we obtain its output as: $\{s_{\text{ins}},s_r\}=f_{\theta^{(2)}}(x)$, where $s_{\text{ins}}$ is the instance-level segmentation map, and $s_r$ is also a segmentation map, with each pixel of $s_r$ representing its camouflage rank.

\subsubsection{The Triple-Task Learning Framework}
\label{subsubsec:triple_task}
With both the joint learning framework (see Section \ref{sec:joint_cod_col}) and the camouflaged instance ranking model (see Section \ref{sec:ranking_model}), we obtain camouflage localization map $s_l^{\text{init}}$, camouflage segmentation map $s_b^{\text{init}}$, and the camouflage instance/ranking maps $\{s_{\text{ins}},s_r\}$ (see Fig.~\ref{fig:network_overview}). As discussed in Section \ref{sec:task_relationship_modeling}, the three tasks are closely related, and the modeling of one task should be beneficial for the other two tasks. To further explore the contribution of COR for the joint learning framework, we introduce the \enquote{triple-task learning framework} (see Fig.~\ref{fig:network_overview}), where the prediction from COR is used to further refine $s_l^{\text{init}}$ and $s_b^{\text{init}}$.

Given the ranking map $s_r$ (for the foreground region, the smaller number of $s_r$ indicates a higher degree (more difficult) of camouflage), we claim it can serve as an instance-level difficulty indicator representing the difficulty-degree of each camouflaged instance. We then define ranking based attention as: $\text{att}^r=1+1/\text{exp}([s_r>0])$, where $[s_r>0]$ indicates the foreground region, and the ranking attention $\text{att}^r$ assigns higher attention to higher levels of camouflaged instances, achieving hard-negative mining. 
With the ranking prediction based attention map $\text{att}^r$, we define the new COL and COD backbone feature as $\{s^{rl}_k\}_{k=1}^4=\{s^{l}_k*\text{att}^r\}_{k=1}^4$ and $\{s_k^{r^2b}\}_{k=1}^4=\{s_k^{rb}*\text{att}^r\}_{k=1}^4$. The same decoder structure from the \enquote{Localization Decoder} and \enquote{Segmentation Decoder} is used to generate the ranking model refined prediction, namely $s_l^{\text{ref}}$ and $s_b^{\text{ref}}$ respectively as shown in Fig.~\ref{fig:network_overview}. We define the ranking refined COL and COD module as \enquote{COR \& COL} and \enquote{COR \& COD} respectively, and their corresponding parameters as $\theta^{(3)}$ and $\theta^{(4)}$.

\begin{figure}[!htp]
  \begin{center}
  \begin{tabular}{c@{ } c@{ } c@{ } c@{ } c@{ } c@{ }}
    {\includegraphics[height=0.153\linewidth]{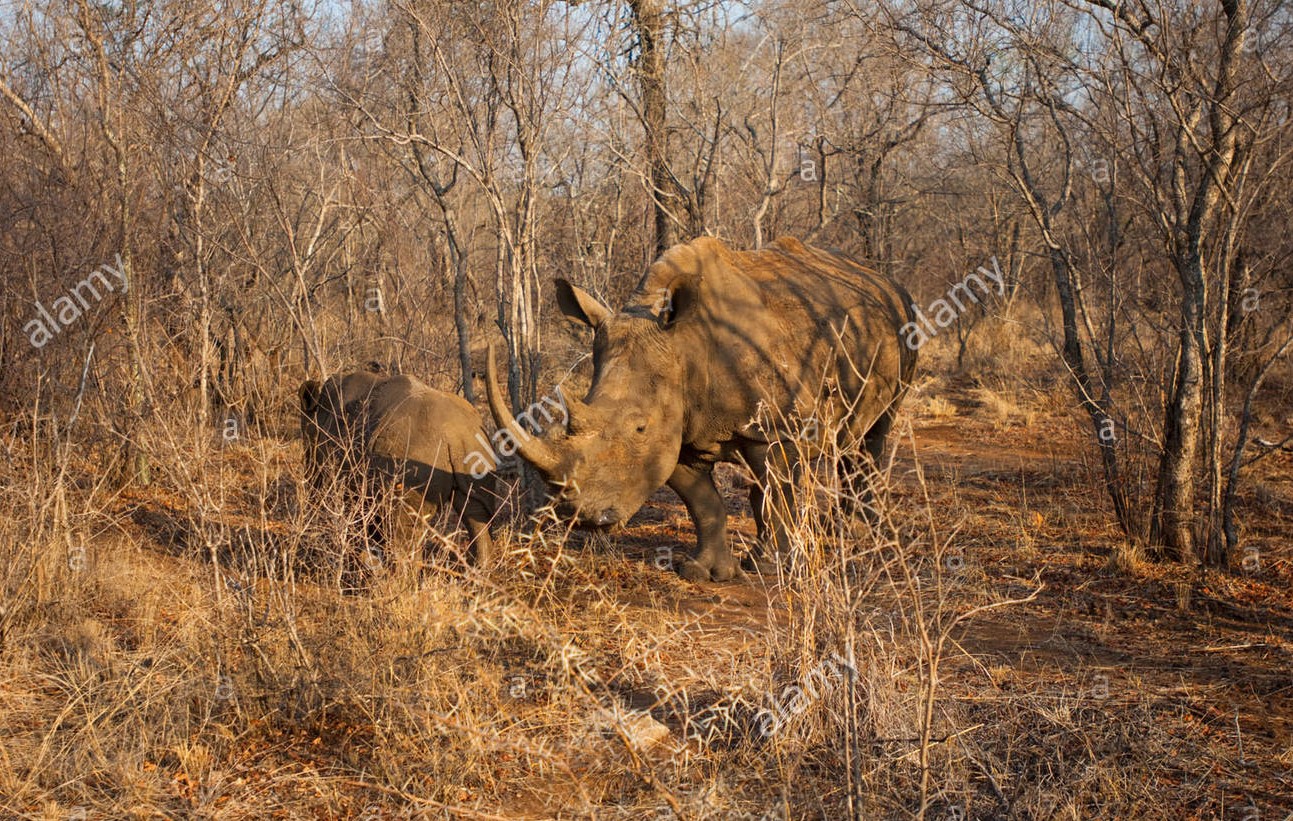}}&
    {\includegraphics[height=0.153\linewidth]{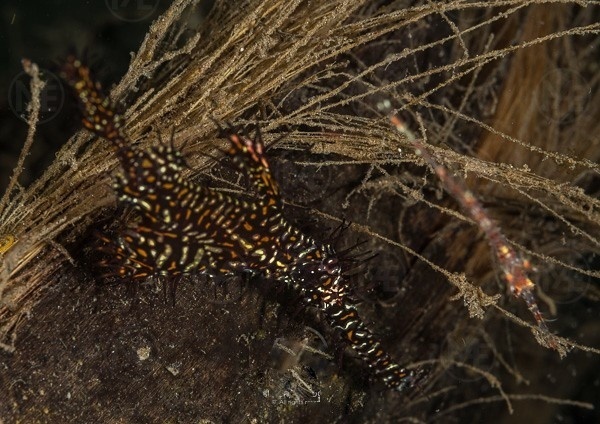}}&
    {\includegraphics[height=0.153\linewidth]{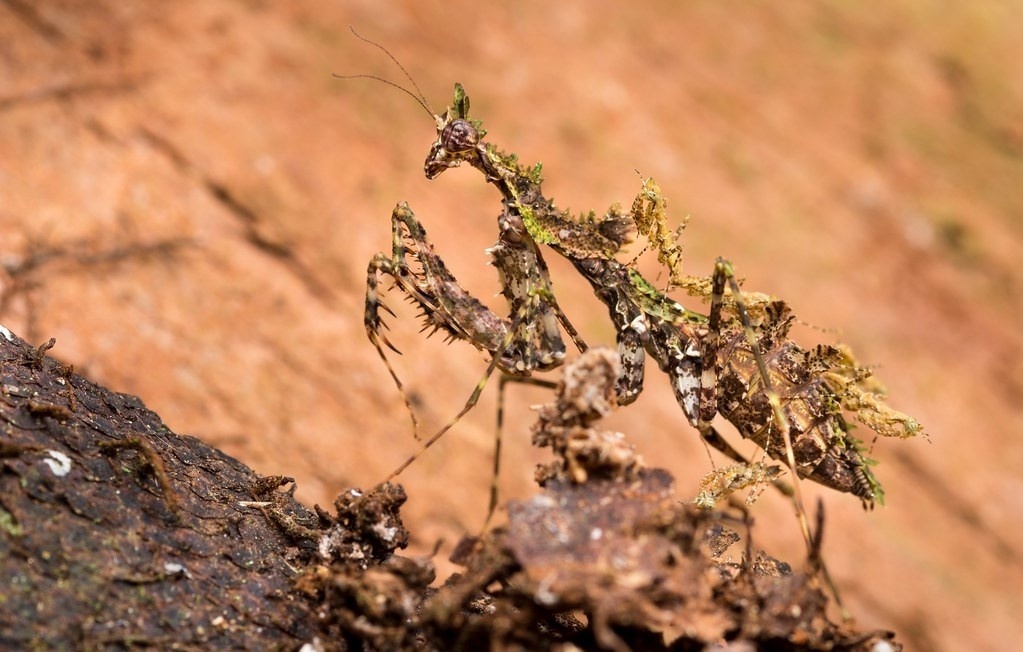}}&
    {\includegraphics[height=0.153\linewidth]{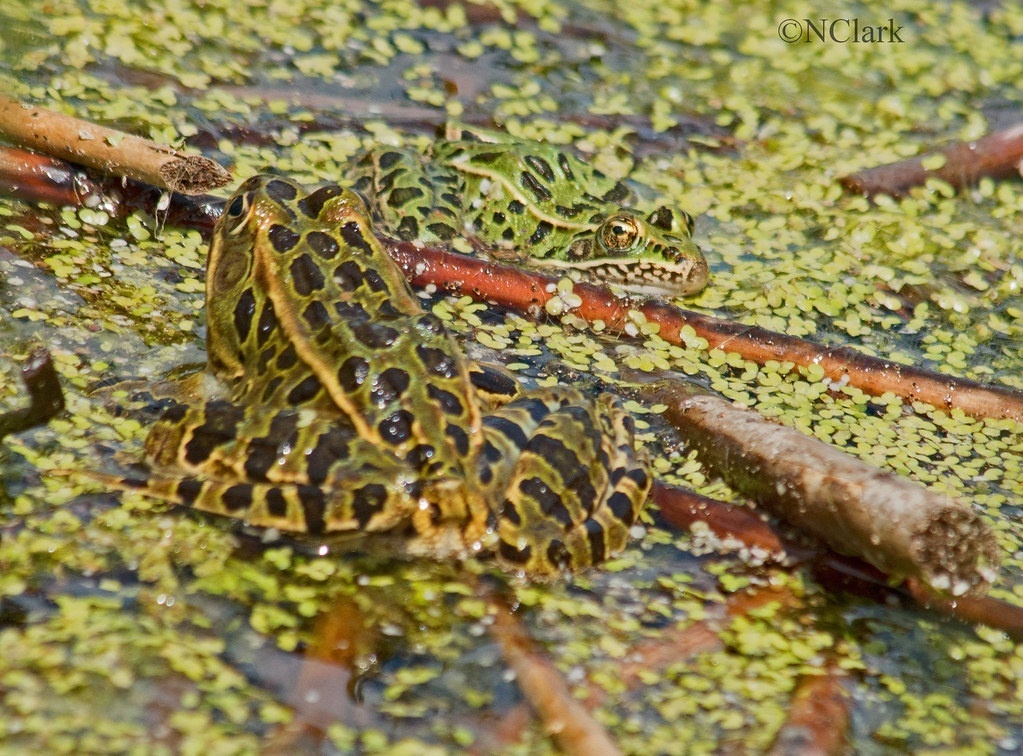}}
    \\
    {\includegraphics[height=0.153\linewidth]{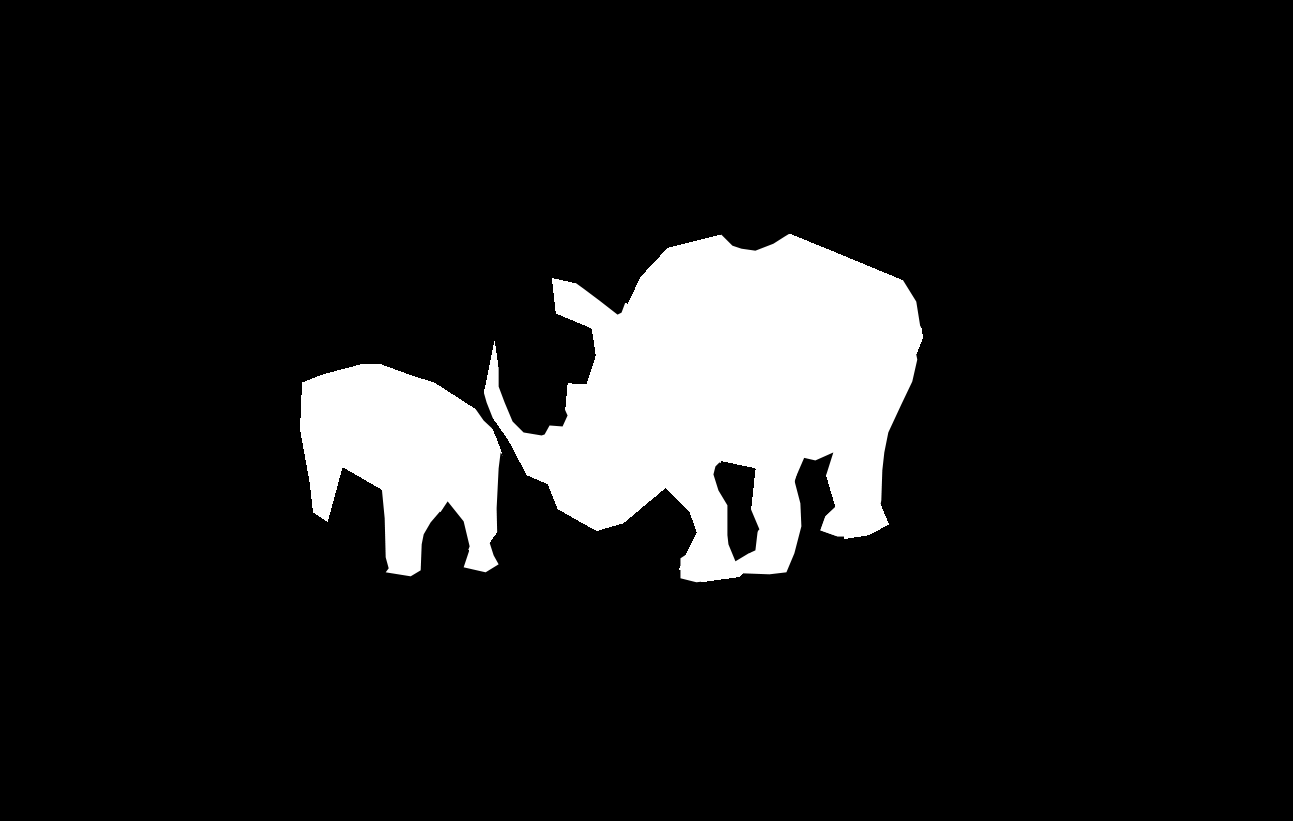}}&
    {\includegraphics[height=0.153\linewidth]{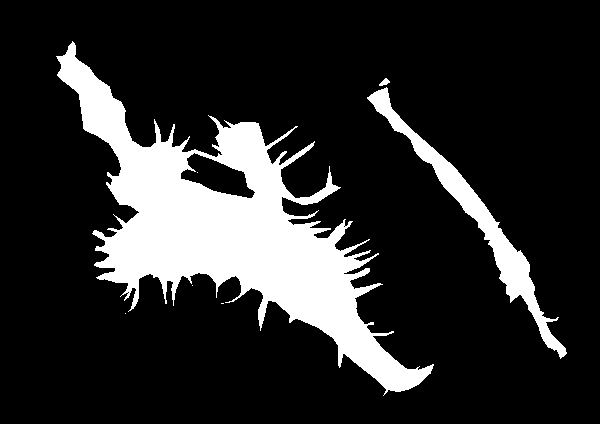}}&
    {\includegraphics[height=0.153\linewidth]{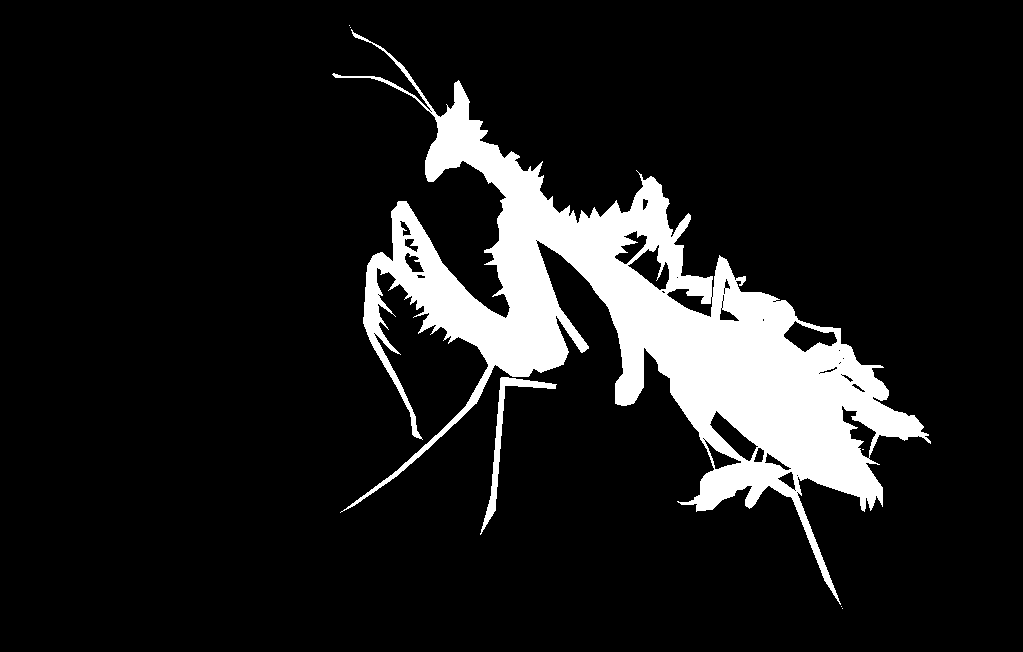}}&
    {\includegraphics[height=0.153\linewidth]{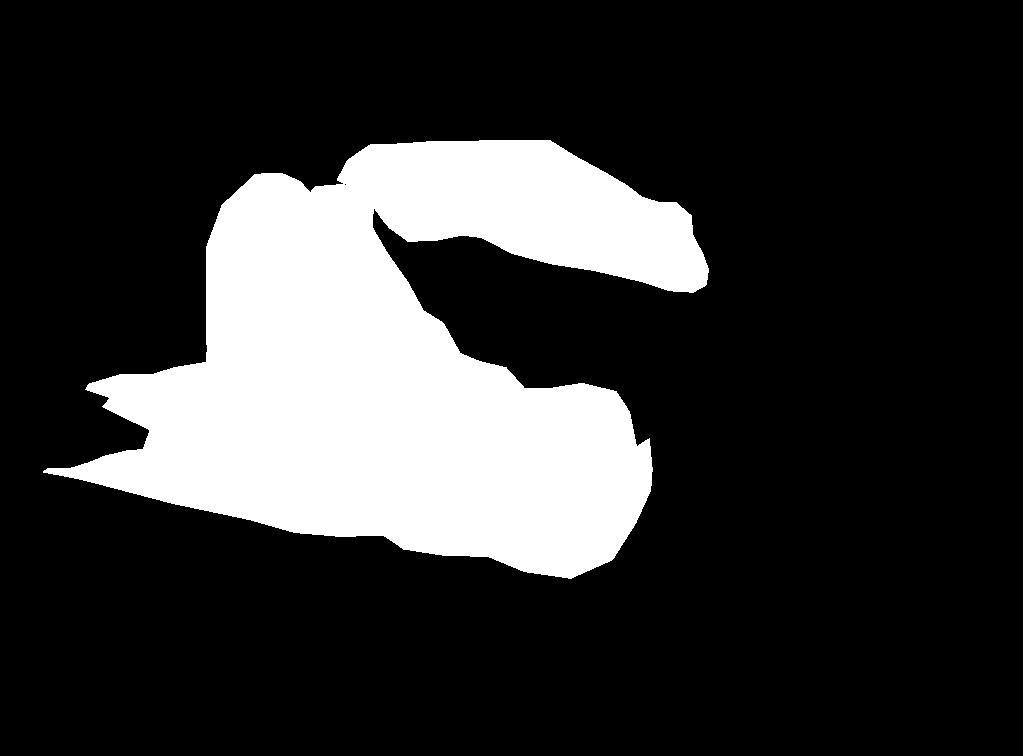}}
    \\
    {\includegraphics[height=0.153\linewidth]{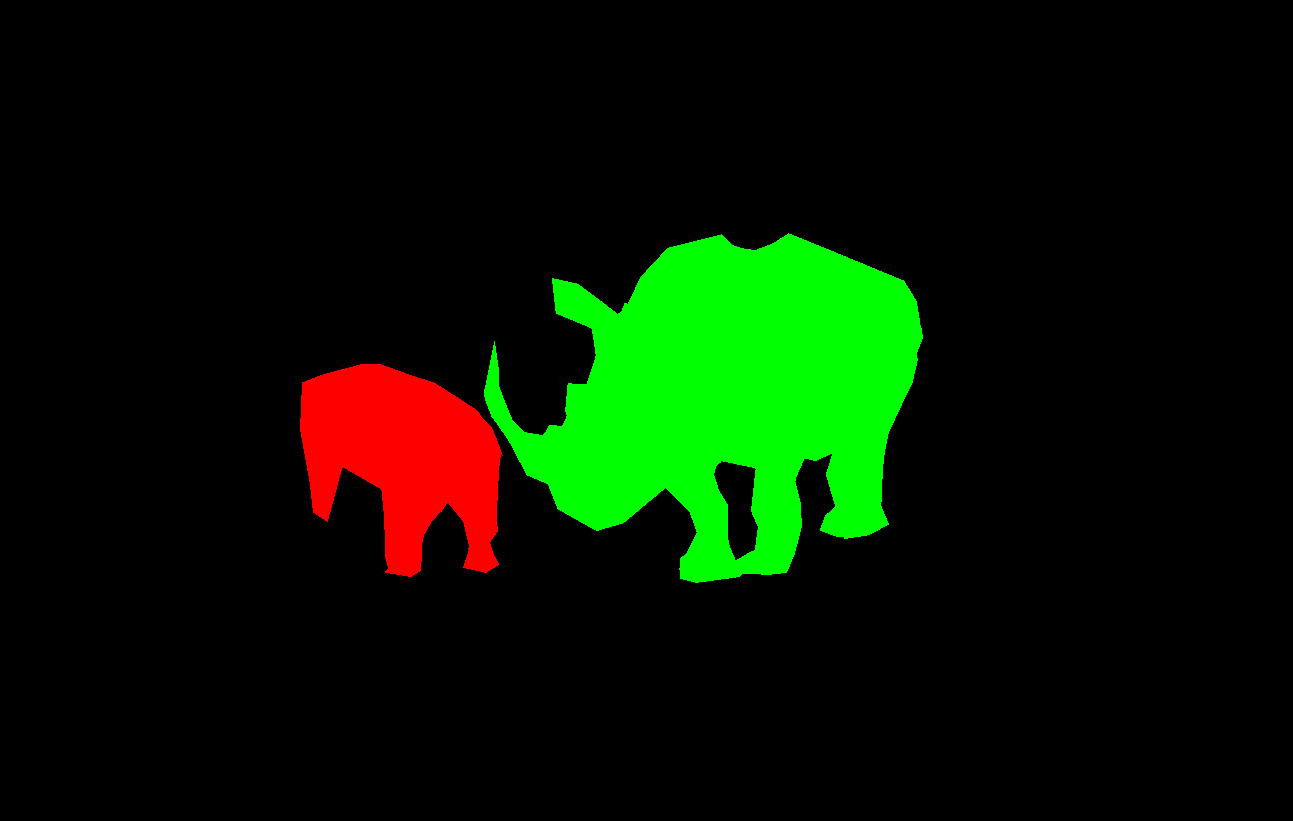}}&
    {\includegraphics[height=0.153\linewidth]{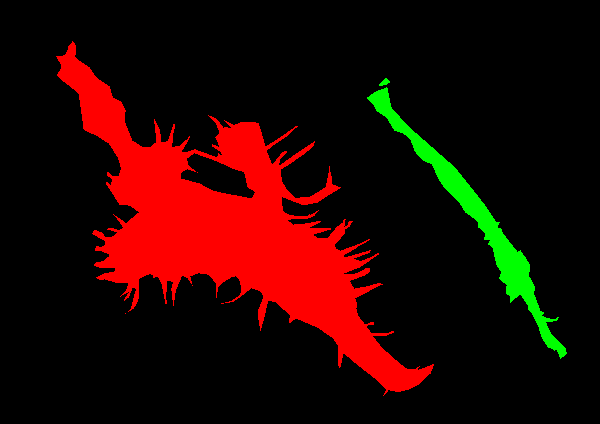}}&
    {\includegraphics[height=0.153\linewidth]{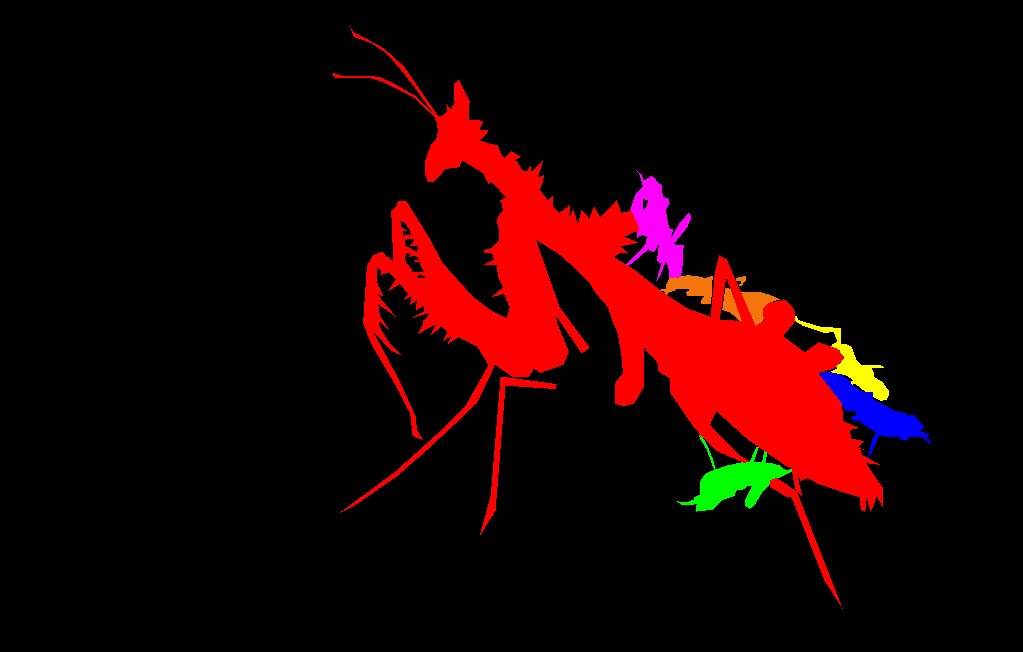}}&
    {\includegraphics[height=0.153\linewidth]{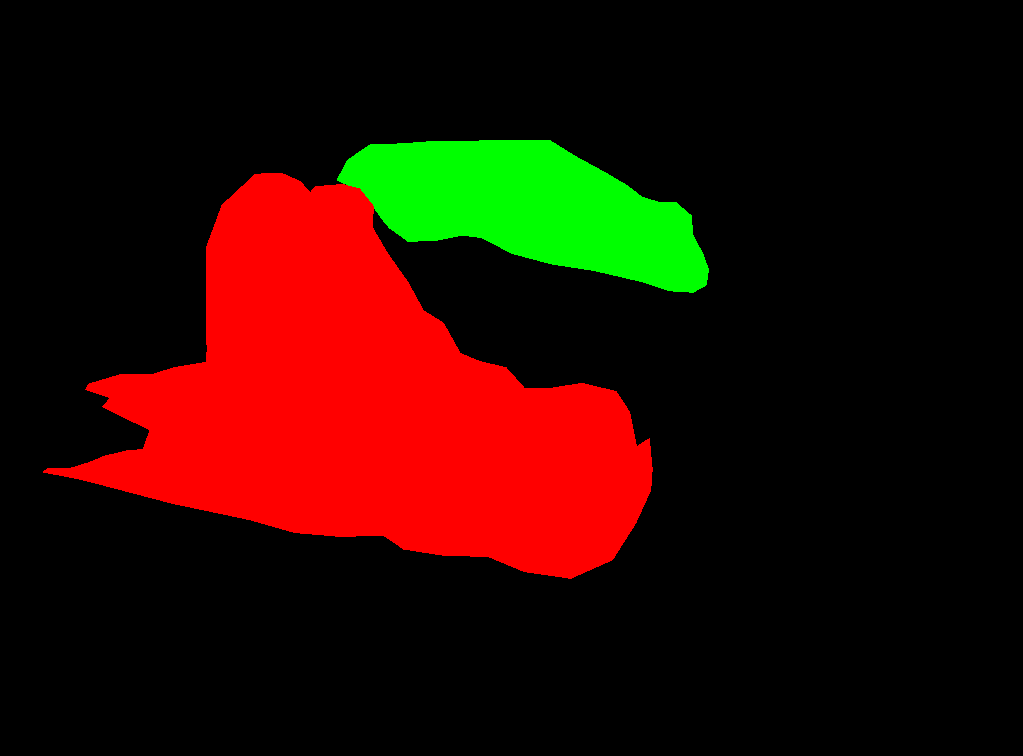}}
    \\  
    {\includegraphics[height=0.153\linewidth]{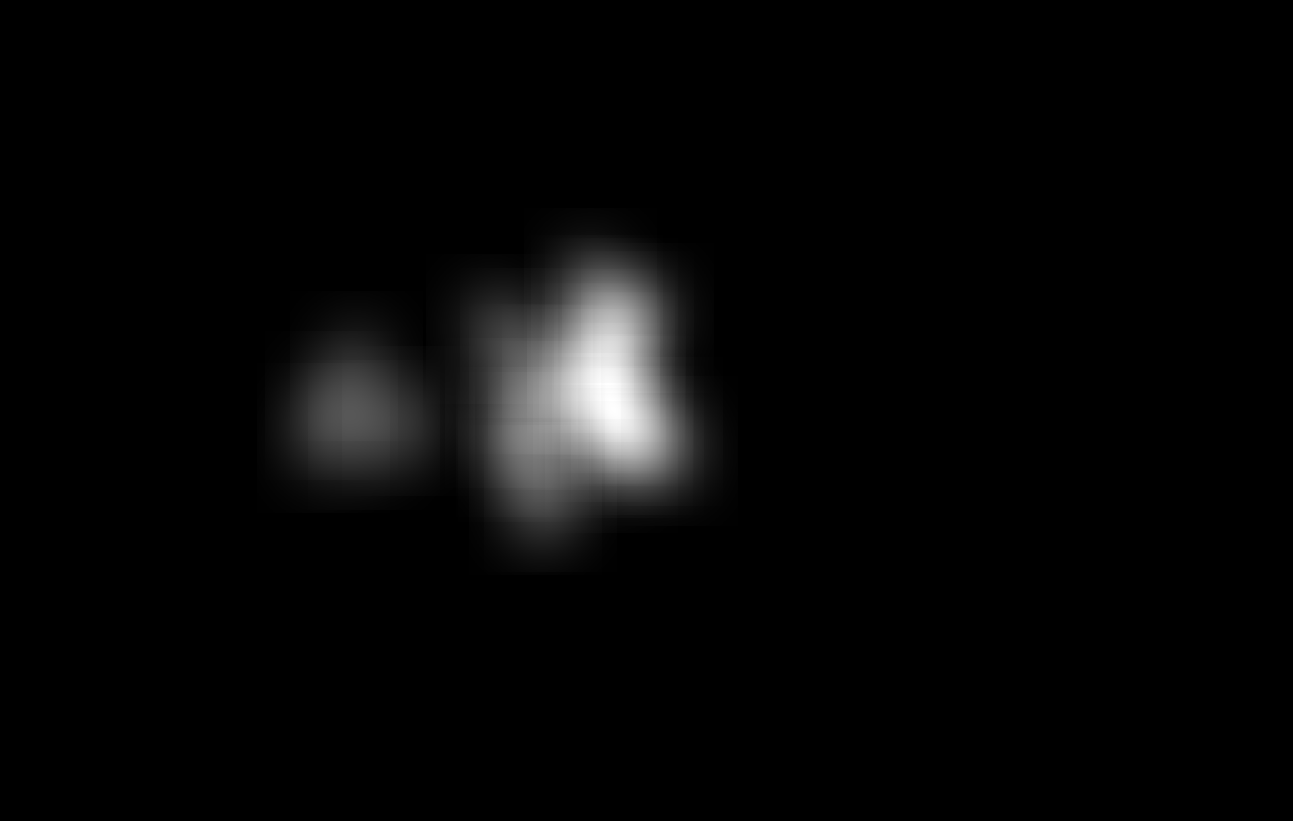}}&
    {\includegraphics[height=0.153\linewidth]{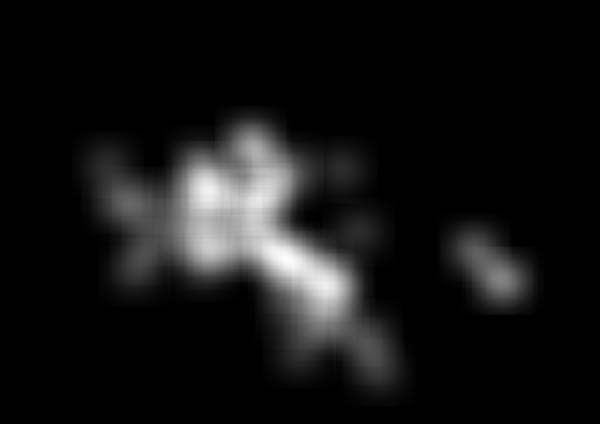}}&
    {\includegraphics[height=0.153\linewidth]{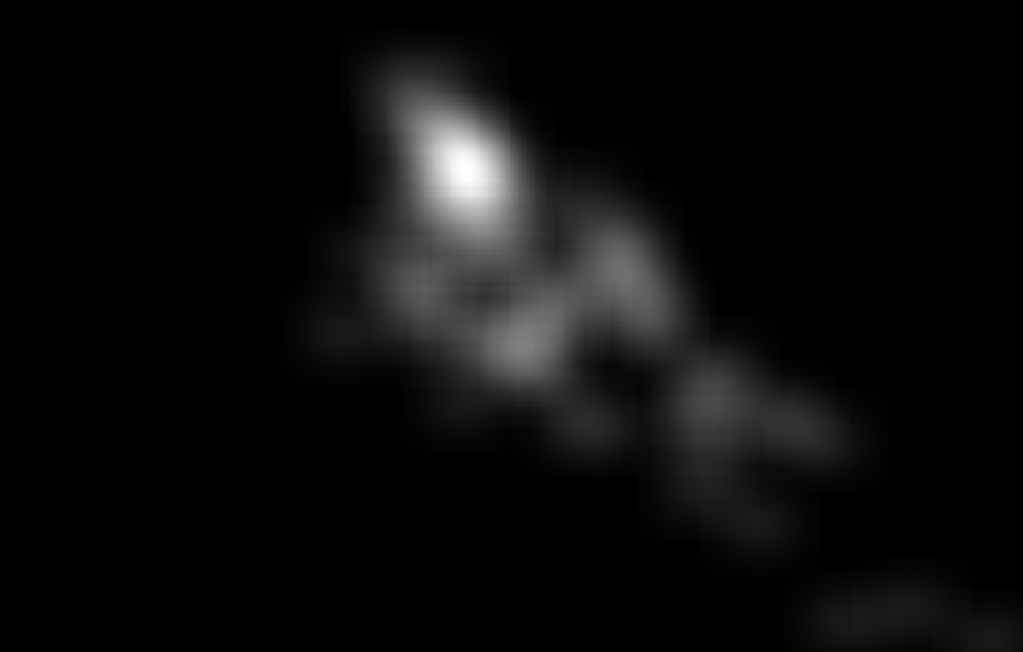}}&
    {\includegraphics[height=0.153\linewidth]{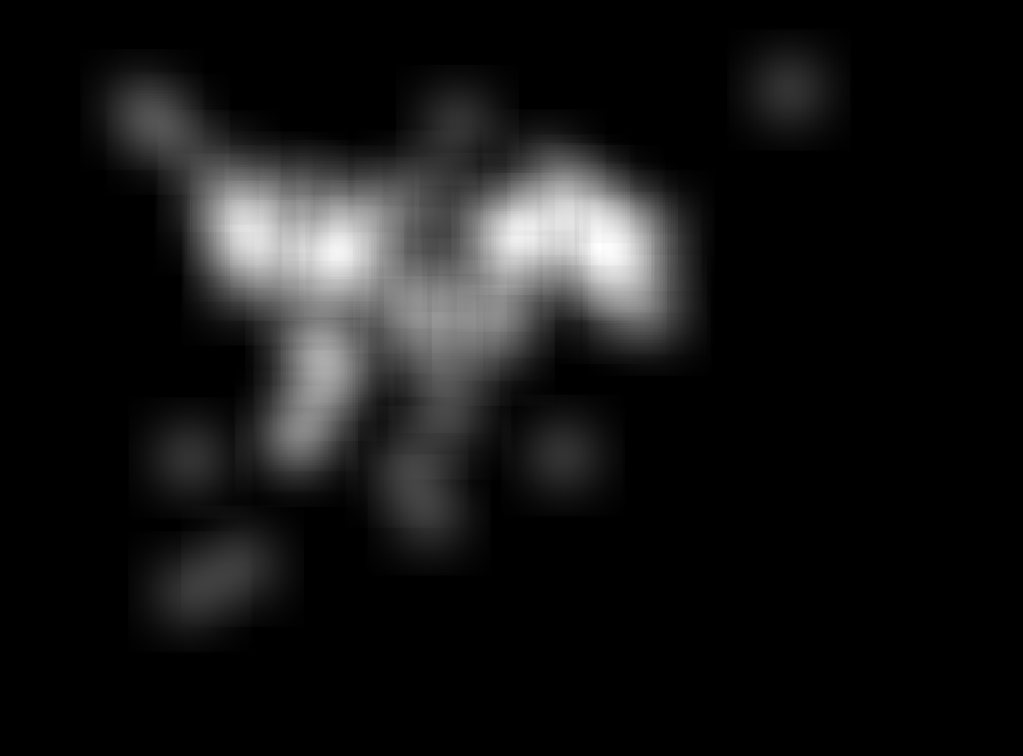}}
    \\
    {\includegraphics[height=0.153\linewidth]{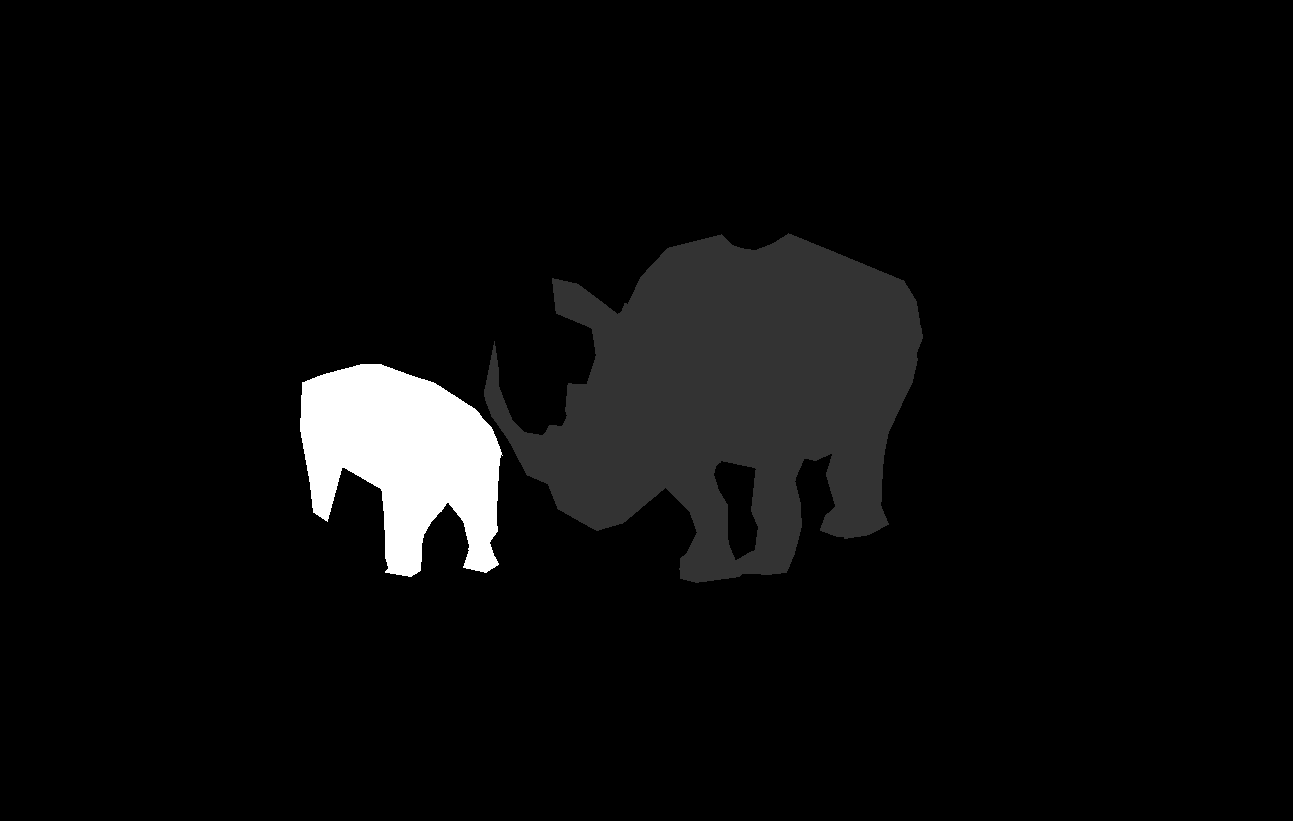}}&
    {\includegraphics[height=0.153\linewidth]{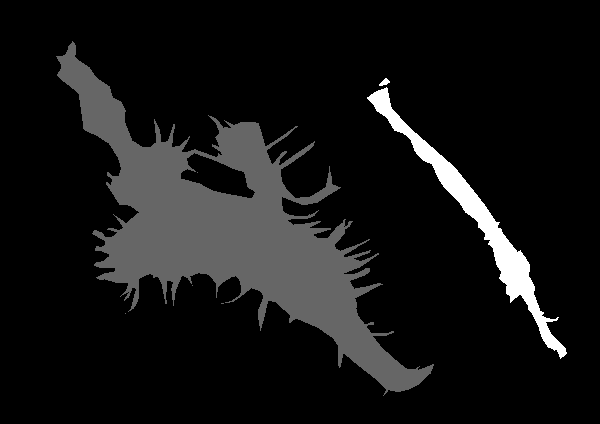}}&
    {\includegraphics[height=0.153\linewidth]{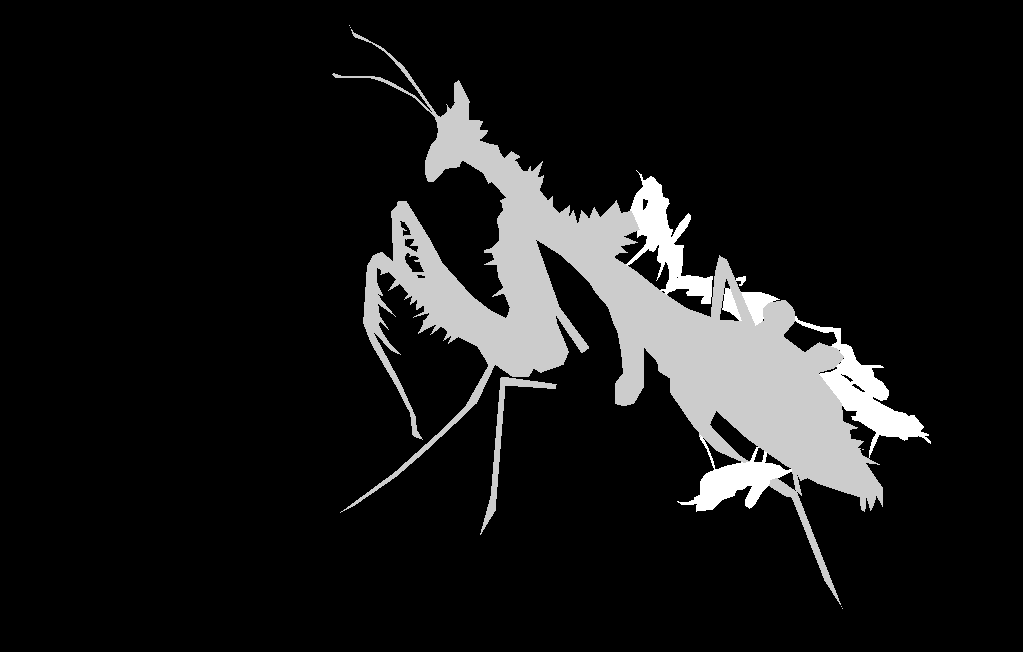}}&
    {\includegraphics[height=0.153\linewidth]{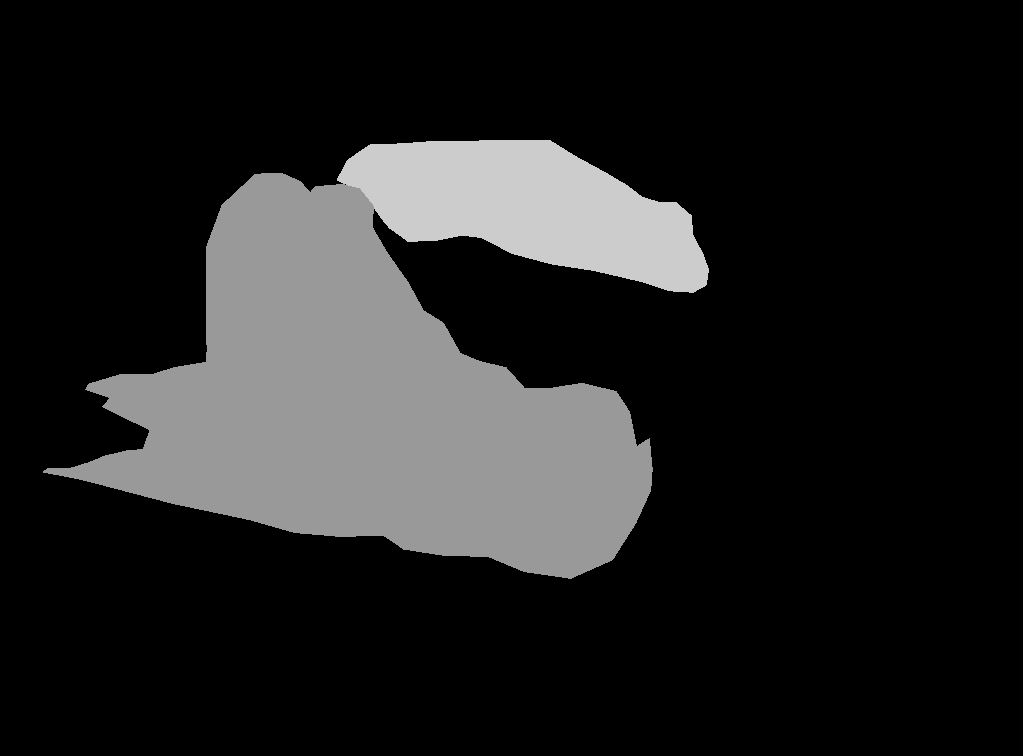}}\\
  \end{tabular}
  \end{center}
  \caption{Visualization of annotations in CAM-LDR dataset. From top to bottom: the original images, binary camouflaged object annotations,
  instance annotations, the camouflaged object localization map,
  the gray-scale ranking annotations.}
\label{fig:annotation_types}
\end{figure}

\noindent\textbf{Training and Testing Pipeline:}
Given the input image $x$, the ranking model with parameters $\theta^{(2)}$ is updated with the weighted rank loss $\mathcal{L}'_{rank}$. We define the \enquote{camouflage localization and segmentation loss} $\mathcal{L}_{dualT}$, which is then used to update $\{\theta^{(1)},\theta^{(3)},\theta^{(4)}\}$, and is defined as:
\begin{equation}
    \label{final_loss_cod_col}
    \mathcal{L}_{dualT} = \mathcal{L}_{dual} + \beta(\mathcal{L}_2(s_l^{\text{ref}},y_l) + \mathcal{L}_c(s_b^{\text{ref}},y_b)),
\end{equation}
where $\beta$ is used to control the contribution of the refined predictions, and empirically we set $\beta=1$.
At test time, our model can directly output $\{s_{\text{ins}},s_r\}$ from the ranking model, and the refined camouflage localization map $s_l^{\text{ref}}$ and segmentation map $s_b^{\text{ref}}$ from the \enquote{COR \& COL} module and \enquote{COR \& COD} module respectively.

\section{CAM-LDR Dataset}
In order to fully understand the mechanisms of camouflage, especially the inconsistent difficulty of detectability within a camouflaged instance and across the different camouflaged instances, we present a new dataset named \enquote{CAM-LDR} for camouflaged object localization and camouflaged object ranking (see Fig.~\ref{fig:annotation_types}). Our basic assumption is that the longer it takes for the viewer to find the camouflaged instance, the higher the level (rank) of the camouflaged instance \cite{troscianko2009camouflage}. With this, we use an eye tracker to record the detection delay for each camouflaged instance, which serves as an indicator of the ranking of the camouflaged object. 


Specifically, we use SMI RED250 eye tracker to obtain the camouflaged object localization map.
SMI RED250 provides three sampling rates, 60Hz, 120Hz, and 250Hz, representing the accuracy of the recorded detection delay. We use the 250Hz sampling rate in our experiment. The operating distance is 60-80cm, which is the distance from observers to the camouflaged image. The movement range is 40cm in the horizontal direction and 20cm in the vertical direction, which is the range for the observers to move in order to discover the camouflaged objects.

With the existing camouflaged object detection training datasets (which is the combination of 3,040 images from the COD10K training dataset \cite{fan2020camouflaged} and 1,000 images from the CAMO training datasets \cite{le2019anabranch}), we invite six observers to find camouflaged instances within the image.
Note that we have multiple observers to robustly predict the level of camouflage
We define the median observation time across different observers as the detection delay for each camouflaged instance.
Specifically, we define the observation time for the $j$-th observer towards the $i$-th instance as:
\begin{equation}
\triangle t_{ij} = \mathrm{median}(\delta t_{ij}), 
\ \delta t_{ij}=\{t_{ij}^{k}-t_{j}^0\}_{k=1}^K
\end{equation}
$K$ is the number of fixation points on the instance, $t_{j}^0$ is the start time for observer $j$ to watch the image and $t_{ij}^{k}$ is the time of the $k$-th fixation point on the instance $i$ with observer $j$. To avoid the influence of extremely high or low fixation times, we use the median instead of the mean value:
\Rev{
\begin{equation}
\mathrm{median}(\boldsymbol{x})=\left\{\begin{array}{cc}x_{(n+1) / 2}, & \mathrm{n} \% 2 \neq 0 \\ \frac{x_{n / 2}+x_{(n / 2)+1}}{2}, & \mathrm{n} \% 2=0\end{array}\right.
\end{equation}}
where $\boldsymbol{x}=\{x_l\}_{l=1}^n$ is a set indexed in ascending order.  
Considering different perception ability of observers,
we define the final detection delay for instance $i$ as the median across the six observers: $\triangle t_i = \mathrm{median}_j(\triangle t_{ij})$, based on which we
obtain our ranking based dataset
(see Fig.~\ref{fig:annotation_types}).

There exist two different cases that result in no fixation points within the camouflaged instance region.
The first is caused by
a mechanical error of the eye tracker or incorrect operation of the observers. The second is caused by the higher level of camouflage, which makes it difficult to detect the camouflaged object.
For the one with tracker failure across most of the observers, we
consider it as a hard sample and the search time is set to 1 (after normalization).
Otherwise, if the majority of the observers (4 observers in our setting) can reach an agreement in identifying the camouflaged instance, while some other observers failed to localize the camouflaged instance, we define this type of tracker failure as an operation error, and the observation time of this instance is decided by the majority observers that reach a consensus of camouflage. 
\Rev{With the observation time of instances, we rearrange the instances by increasing order and then divide the instances equally into 5 subsets with 5 ranks, \ie~easy, medium1, medium2, medium3 and hard. }  

\noindent\textbf{Dataset information:} Our CAM-LDR dataset comprises 4,040 training images with fixation annotations and ranking annotations, 1000 from the CAMO training set and 3040 from the COD10K training set, and 2,026 testing images from the COD10K testing set.
In our
ranking data,
we have six different ranks, \ie~background (BG), easy (ES), medium1 (M1), medium2 (M2), medium3 (M3), and hard (HD).

\begin{table*}[t!]
  \centering
  \scriptsize
  \renewcommand{\arraystretch}{1.15}
  \renewcommand{\tabcolsep}{0.97mm}
  \caption{Performance comparison with benchmark COD models.
  }
  \begin{tabular}{l|c|c|c|cccc|cccc|cccc|cccc}
  \hline
  &&&&\multicolumn{4}{c|}{CAMO \cite{le2019anabranch}}&\multicolumn{4}{c|}{CHAMELEON \cite{Chameleon2018}}&\multicolumn{4}{c|}{COD10K \cite{fan2020camouflaged}}&\multicolumn{4}{c}{NC4K \cite{yunqiu_cod21}} \\
    Method &Backbone&Input Scale&Year& $S_{\alpha}\uparrow$&$F_{\beta}\uparrow$&$E_{\xi}\uparrow$&$\mathcal{M}\downarrow$& $S_{\alpha}\uparrow$&$F_{\beta}\uparrow$&$E_{\xi}\uparrow$&$\mathcal{M}\downarrow$& 
    $S_{\alpha}\uparrow$&$F_{\beta}\uparrow$&$E_{\xi}\uparrow$&$\mathcal{M}\downarrow$&
   $S_{\alpha}\uparrow$&$F_{\beta}\uparrow$&$E_{\xi}\uparrow$&$\mathcal{M}\downarrow$  \\
  \hline
  SINet \cite{fan2020camouflaged}& ResNet50 & $352\times 352$ & 2020 & .745 & .702 & .804 & .092 & .872 & .827 & .936 & .034 & .776 & .679 & .864 & .043 & .810 & .772 & .873 & .057 \\ 
  MGL~\cite{zhai2021Mutual}& ResNet50 & $473\times 473$ & 2021 & .775 & .726 & .812 & .088 & \cellcolor{Ocean}{\bf.893} & {\bf.834} & .918 & \cellcolor{Ocean}{\bf.030} & {\bf.814} & {\bf.711} & .852 & {\bf.035} & .833 & .782 & .867 & .052\\
  PFNet~\cite{mei2021Ming}& ResNet50 & $416\times 416$ & 2021 & .782 & {\bf.744} & \cellcolor{Ocean}{\bf.840} & .085 & .882 & .826 & .922 & .033 & .800 & .700 & {\bf.875} & .040 & .829 & .782 & {\bf.886} & .053 \\
  C2FNet \cite{sun2021c2fnet} & ResNet50 & $352 \times 352$ & 2021  & .611 & .481 & .672 & .147 & .791 & .704 & .860 & .069 & .638 & .438 & .718 & .089 & .681 & .570 & .744 & .110  \\
  ERRNet \cite{ji2022fast} & ResNet50 & $352 \times 352$ & 2022  & .690 & .599 & .730 & .112 & .825 & .756 & .888 & .047 & .715 & .572 & .795 & .053 & .764 & .697 & .833 & .071 \\
 TANet \cite{ren2021deep} & ResNet50  & $384\times 384$ &  2021 & \cellcolor{Ocean}{\bf.793} & .690& {\bf.834}& .083& {\bf.888}& .786& .911& .036& .803& .629& .848& .041 & - & - & - & - \\
  ZoomNet \cite{ZoomNet_CVPR2022} & ResNet50  & $384 \times 384$ &  2022 & {\bf.789} & .741 & .829 & \cellcolor{Ocean}{\bf.076} & .865 & .823 & \cellcolor{Ocean}{\bf.939} & {\bf.031} & \cellcolor{Ocean}{\bf.821} & \cellcolor{Ocean}{\bf.741} & .866 & \cellcolor{Ocean}\cellcolor{Ocean}{\bf.032} & {\bf.839} & {\bf.796} & .867 & \cellcolor{Ocean}{\bf.046}  \\
LSR \cite{yunqiu_cod21} & ResNet50 & $352 \times 352$ & 2021  & \cellcolor{Ocean}{\bf.793} & .725 & .826 & .085 & \cellcolor{Ocean}{\bf.893} & \cellcolor{Ocean}{\bf.839} & {\bf.938} & .033 & .793 & .685 & .868 & .041 & {\bf.839} & .779 & .883 & .053  \\
{\bf LSR+}  &ResNet50 & $352 \times 352$ & 2022 & {\bf.789} & {\cellcolor{Ocean}\bf.751} & \cellcolor{Ocean}{\bf.840} & {\bf.079} & .878 & .828 & .929 & .034 & .805 & {\bf.711} & {\cellcolor{Ocean}\bf.880} & .037 & \cellcolor{Ocean}{\bf.840} & \cellcolor{Ocean}{\bf.801} & {\cellcolor{Ocean}\bf.896} & {\bf.048} \\
  \hline\multicolumn{19}{c}{Models with other backbones or extra training data}\\ \hline
  UJSC \cite{li2021uncertainty} & ResNet50 & $352 \times 352$ & 2021 & .803 & .759 & .853 & .076 & {\bf.894} & .848 & \cellcolor{Ocean}\bf{.943} & {\bf.030} & .817 & .726 & .892 & .035 & .842 & .806 & .898 & .047 \\
  C2FNet-V2 \cite{sun2021c2fnet} & Res2Net50 \cite{res2net} & $352 \times 352$ & 2021  & .772 & .737 & .825 & .087 & {.889} & \cellcolor{Ocean}\bf{.853} & .941 & {\bf.030} & .807 & .719 & .883 & .036 & .837 & .805 & .894 & .049  \\
  SINet-V2~\cite{fan2021concealed} & Res2Net50 \cite{res2net} & $352\times 352$ & 2022 & {\bf.820} & .782 & {\bf.882} & {\bf.070} & .888 & .835 & {\bf.942} & {\bf.030} & .815 & .718 & .887 & .037 & .847 & .805& .903 & .048\\
  \Rev{\bf LSR+$^1$}  & \Rev{Res2Net50~\cite{fan2021concealed}} & \Rev{$352 \times 352$} & \Rev{2022} & \Rev{{\bf.820}} & \Rev{{\bf.794}} & \Rev{.874} & \Rev{.071} & \Rev{.888} & \Rev{.840} & \Rev{.931} & \Rev{.032} & \Rev{{\bf.831}} & \Rev{{\bf.748}} & \Rev{{\bf.898}} & \Rev{{\bf.032}} & \Rev{{\bf.859}} & \Rev{{\bf.823}} & \Rev{{\bf.911}} & \Rev{{\bf.042}} \\
  \Rev{{\bf LSR+$^{2}$}} & \Rev{SwinT~\cite{liu2021Swin}}  & \Rev{$384 \times 384$} & \Rev{2022}  & \cellcolor{Ocean}\Rev{\bf{.854}}  & \cellcolor{Ocean}\Rev{\bf{.839}}  & \cellcolor{Ocean}\Rev{\bf{.924}}  & \cellcolor{Ocean}\Rev{\bf{.049}}  & \cellcolor{Ocean}\Rev{\bf{.895}}  & \Rev{\bf{.849}}  & \cellcolor{Ocean}\Rev{\bf{.943}}  & \cellcolor{Ocean}\Rev{\bf{.025}}  & \cellcolor{Ocean}\Rev{\bf{.847}}  & \cellcolor{Ocean}\Rev{\bf{.775}}  & \cellcolor{Ocean}\Rev{\bf{.924}}  & \cellcolor{Ocean}\Rev{\bf{.028}}  & \cellcolor{Ocean}\Rev{\bf{.870}}  & \cellcolor{Ocean}\Rev{\bf{.845}}  & \cellcolor{Ocean}\Rev{\bf{.924}}  & \cellcolor{Ocean}\Rev{\bf{.036}}  \\
   \hline
  \end{tabular}
  \label{tab:benchmark_model_comparison}
\end{table*}

\begin{table}[t!]
  \centering
  \scriptsize
  \renewcommand{\arraystretch}{1.2}
  \renewcommand{\tabcolsep}{1.2mm}
  \caption{COL performance on the COD10K testing dataset \cite{fan2020camouflaged}.}
  \begin{tabular}{l|cccccc}
  \hline
  Method&$SIM\uparrow$ & $CC\uparrow$ & $EMD\downarrow$ & $KLD\downarrow$ & $NSS\uparrow$ & $AUC\_J\uparrow$   \\ \hline
   GazeGAN \cite{che2019gaze} &0.498 & 0.617 & \textbf{2.293} & 1.507 & 1.186 & 0.934   \\
   SalGAN \cite{pan2017salgan} &0.489 & 0.638 & 2.317 & 1.078 & \textbf{1.347} & 0.939   \\
   UAVDVSM \cite{he2019understanding} &0.344 & 0.400 & 3.447 & 1.437 & 0.821 & 0.888   \\
  SimpleNet \cite{reddy2020tidying} & 0.485 & 0.622 & 2.401 & 0.990 & 1.222 & 0.933   \\
  \Rev{DeepGaze~\cite{deepgazeiie}} &\Rev{0.476} & \Rev{0.632} & \Rev{2.469} & \Rev{1.303} & \Rev{0.983}& \Rev{0.945}   \\
\Rev{SalFBNet~\cite{ding2022salfbnet}} &\Rev{0.390} & \Rev{0.494} & \Rev{3.291} & \Rev{1.870} & \Rev{1.132}& \Rev{0.905}   \\
  \hline
   LSR+ &\textbf{0.572} & \textbf{0.767} & 2.408 & \textbf{0.791} & 1.265 & \textbf{0.971}   \\ \hline
  \end{tabular}
  \label{tab:fixation_baseline_cod10k}
\end{table}

\begin{table}[t!]
  \centering
  \scriptsize
  \renewcommand{\arraystretch}{1.2}
  \renewcommand{\tabcolsep}{4.0mm}
  \caption{Comparison of camouflage ranking methods.}
  \begin{tabular}{r|cccc}
  \hline
  Method & $MAE\downarrow$ & $r_{MAE}\downarrow$ & $Corr\uparrow$ \\\hline
  Baseline (MaskRCNN) & 0.050 & 0.242 & 0.448 \\
  SOLOv2 \cite{wang2020SOLOv2} & 0.051 & 0.402 & 0.438 \\
  MS-RCNN \cite{liu2018path} & 0.053 & 0.267 & 0.433\\
   PPA \cite{fang2021salient} & \textbf{0.042} & 0.288 & 0.410\\ \hline
   LSR+  & 0.046 & \textbf{0.227} & \textbf{0.465} \\
   \hline
  \end{tabular}
  \label{tab:ranking_comparison}
\end{table}

\section{Experimental Results}
\subsection{Setup}
\noindent\textbf{Dataset:}
The benchmark training dataset is the combination of 3,040 images from COD10K training dataset \cite{fan2020camouflaged} and 1,000 images from CAMO training dataset \cite{le2019anabranch}. 
We extend the benchmark training dataset with two extra types of annotation,
namely a camouflaged object localization map and a camouflaged object ranking map. We then test the performance of our model on three existing testing datasets, namely
the CAMO testing dataset \cite{le2019anabranch} (250), COD10K testing dataset \cite{fan2020camouflaged} (2,026), CHAMELEON \cite{Chameleon2018} (76) and our new testing dataset, NC4K (4,121). The number after each dataset indicates its size.

\noindent\textbf{Training details:}
A pretrained ResNet50 \cite{he2016deep} is employed as our backbone network.
During both training and testing, the input image is resized to $352\times 352$. For the camouflaged object ranking model, the candidate bounding boxes spanning three scales (4, 8, 16) and three aspect ratios (0.5, 1.0, 2.0) are selected for each position similar to \cite{he2017mask}.
In the RPN module,
the IoU threshold with the ground truth is set to 0.7, which is used to determine whether the candidate bounding box is positive \Rev{(IoU$>=$0.7)} or negative (IoU$<$0.7) in the next detection phase. The IoU threshold is set to 0.5 to determine whether the camouflaged instances are detected and only positive ones are sent to the segmentation branch. Our triple-task learning model in Fig.~\ref{fig:network_overview} is trained on one GPU (NVIDIA RTX 3090 GPU) for 80 epochs (12 hours) with a mini-batch of 10 images, using the Adam optimizer with an initial learning rate of 2.5e-5. 
\Rev{We use the linear decay strategy with decay rate 0.9 and the decay begins after $50$ epochs}.

\noindent\textbf{Evaluation metrics:}
For COD, we adopt four evaluation metrics, including
Mean Absolute Error, Mean F-measure, Mean E-measure \cite{fan2018enhanced} and S-measure \cite{fan2017structure}, denoted as $\mathcal{M}$, $F_\beta$, $E_\xi$, $S_{\alpha}$, respectively.


Different from the instance segmentation and camouflage object detection tasks, we should consider the progressive relationship between different ranks in the COR task, which has been explained in Sec.~3.3.2. Therefore, we introduce a metric $r_{MAE}$ specifically for COR task:
\begin{equation}
\label{eq:rmae}
    r_{MAE}=\frac{\sum_{i=1}^w\sum_{j=1}^h|r_{ij}-\hat{r}_{ij}|}{N},
\end{equation}
where $N$ is the number of pixels, $w$ and $h$ are the width and height of the image. $\hat{r}_{ij}$ and $r_{ij}$ are the predicted and ground truth ranks respectively with values $1,2,3,4,5,6$ corresponding to \enquote{ES}, \enquote{M1}, \enquote{M2}, \enquote{M3}, \enquote{HD}, and \enquote{BG}, respectively. If the prediction is consistent with the ground truth, their difference is supposed to be 0. In $r_{MAE}$, an \enquote{ES} sample is 
punished less when it is predicted as an \enquote{M1} sample than as an \enquote{HD} sample.

Although $r_{MAE}$ could evaluate the pixel-level ranking difference between the prediction and the ground truth, we should also consider the ranking in instance-level. Therefore, we also present a $Corr$ metric to evaluate the ability of an algorithm to produce the instance-level ranking. To begin with, five instances are sampled randomly from the testing dataset, each with a distinct rank. Each instance has a corresponding predicted bounding box with $IOU$ greater than a certain threshold (0.25 in our setting) and with the maximal rank score. In each sampling, we adopt Spearman’s Rank-Order Correlation to obtain the similarity score between the ground truth rank vector and the predicted rank vector of the 5 examples. 
The average similarity score of $N$ sampling iterations is then calculated. In order to alleviate the influence of the randomness, we compute the average similarity for $M$ iterations and use the mean value as the final $Corr$ score. The $Corr$ metric is written as follows:
\begin{equation}
    Corr=\frac{1}{M}\frac{1}{N}\sum_{l=1}^M\sum_{k=1}^{N}\rho_l(\boldsymbol{r}_k, \boldsymbol{\hat r}_k),
\end{equation}
where $\rho(\cdot)$ represents Spearman’s Rank-Order Correlation, $\boldsymbol{r}_k$ is the ground truth rank vector in the $k$-th sampling and $\boldsymbol{\hat r}_k$ is its corresponding predicted rank vector. 

\begin{figure}[!htp]
   \begin{center}
   \begin{tabular}{c@{ } c@{ } c@{ } c@{ } c@{ } c@{ }}
  {\includegraphics[width=0.150\linewidth]{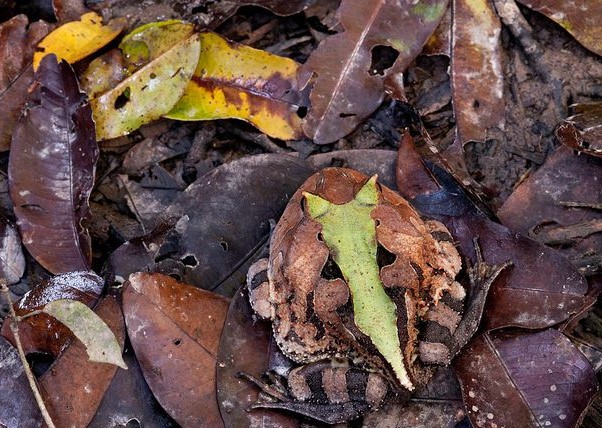}}&
    {\includegraphics[width=0.150\linewidth]{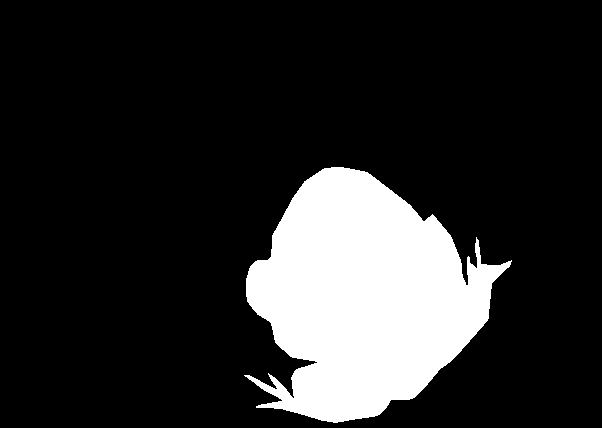}}&
    {\includegraphics[width=0.150\linewidth]{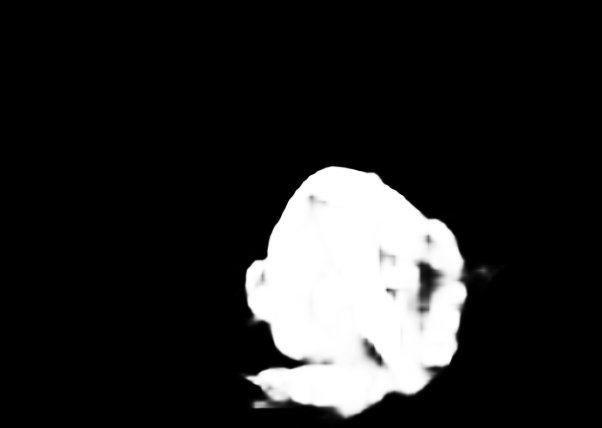}}&
    {\includegraphics[width=0.150\linewidth]{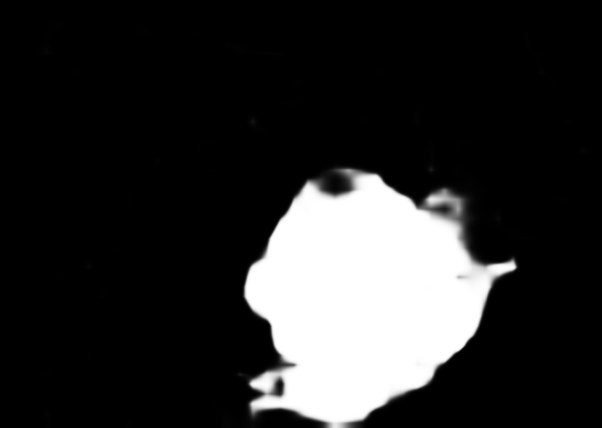}}&
    {\includegraphics[width=0.150\linewidth]{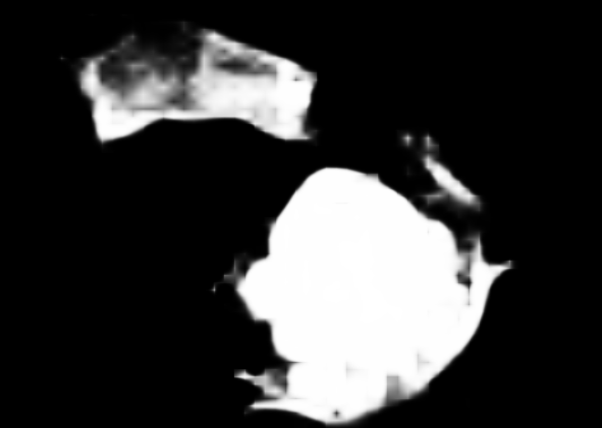}}&
    {\includegraphics[width=0.150\linewidth]{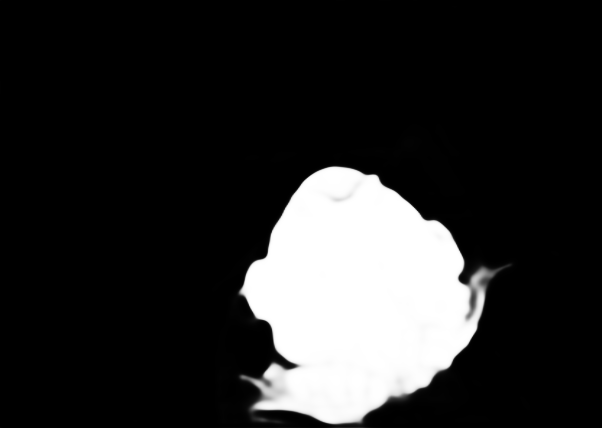}}\\
    {\includegraphics[width=0.150\linewidth]{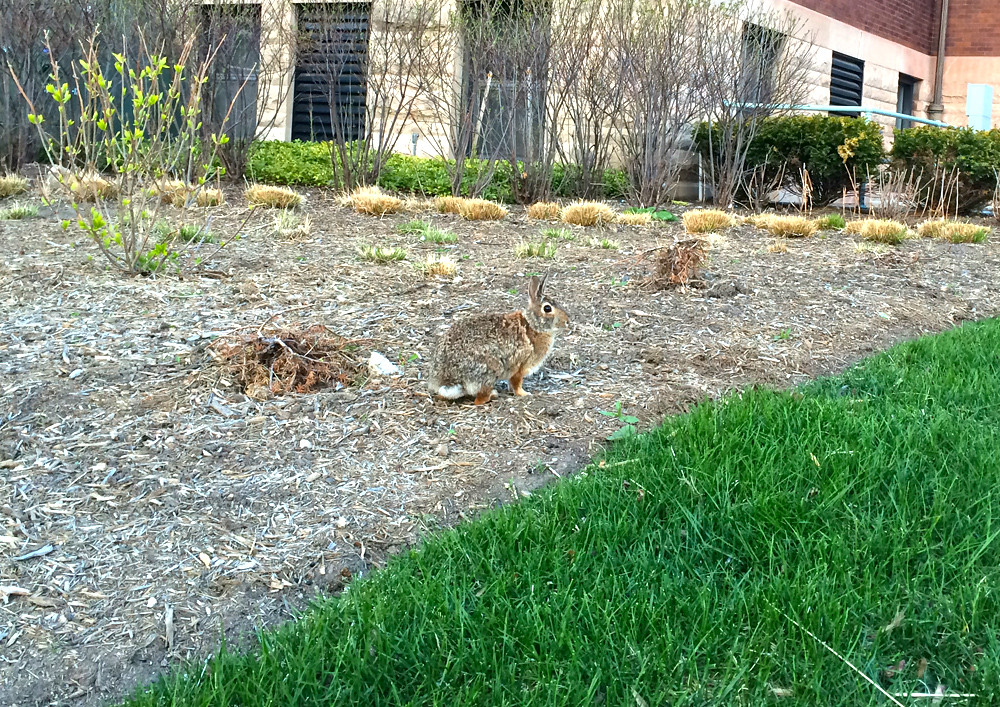}}&
    {\includegraphics[width=0.150\linewidth]{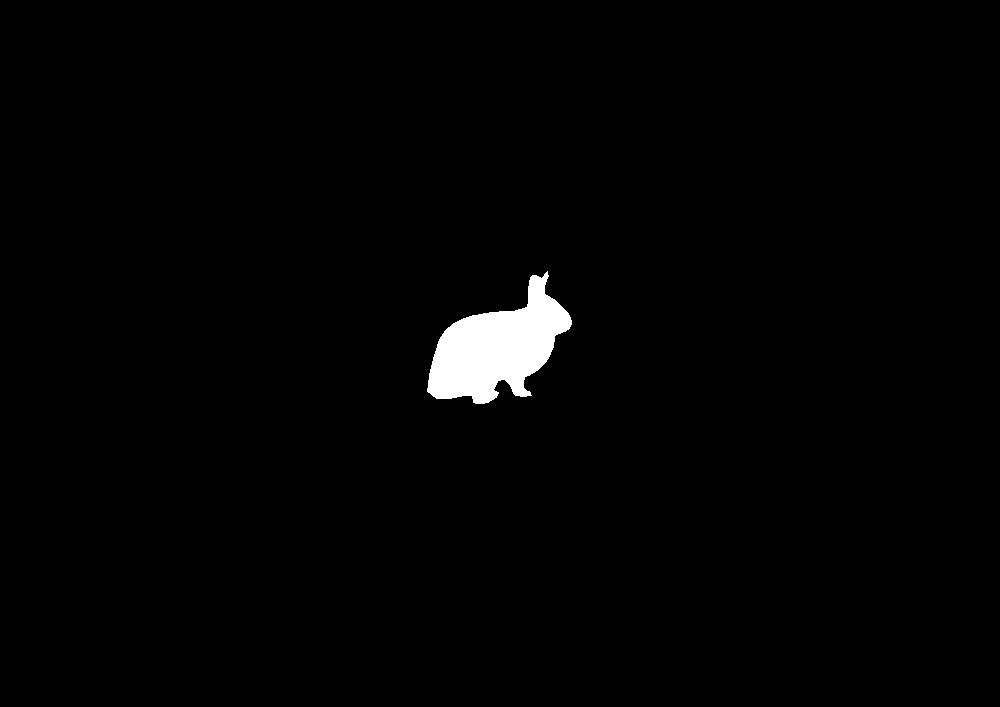}}&
    {\includegraphics[width=0.150\linewidth]{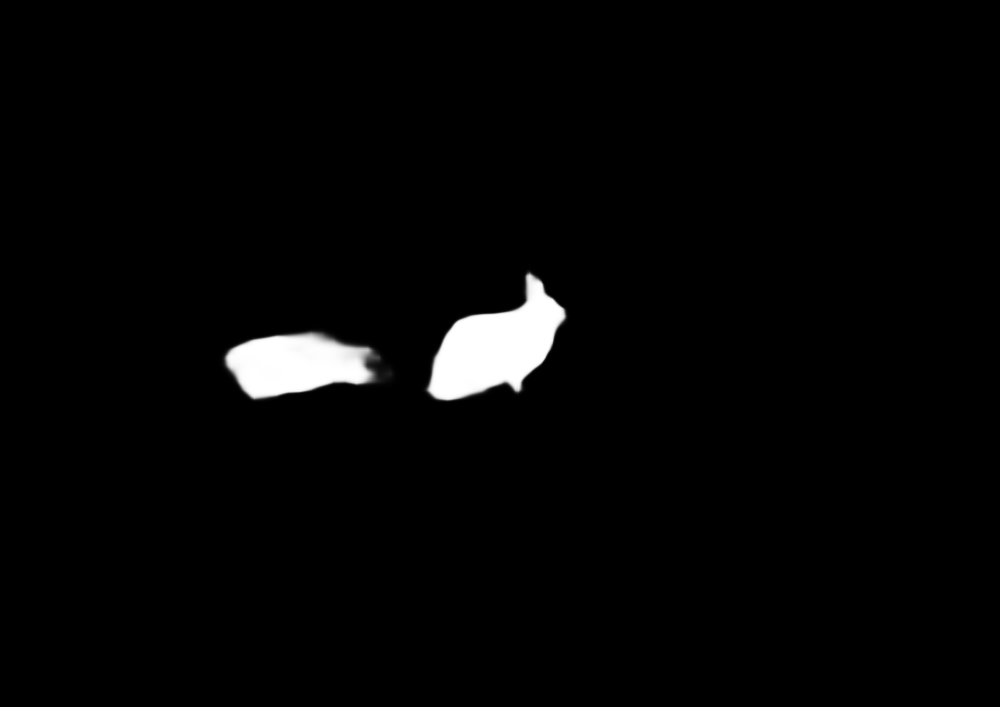}}&
    {\includegraphics[width=0.150\linewidth]{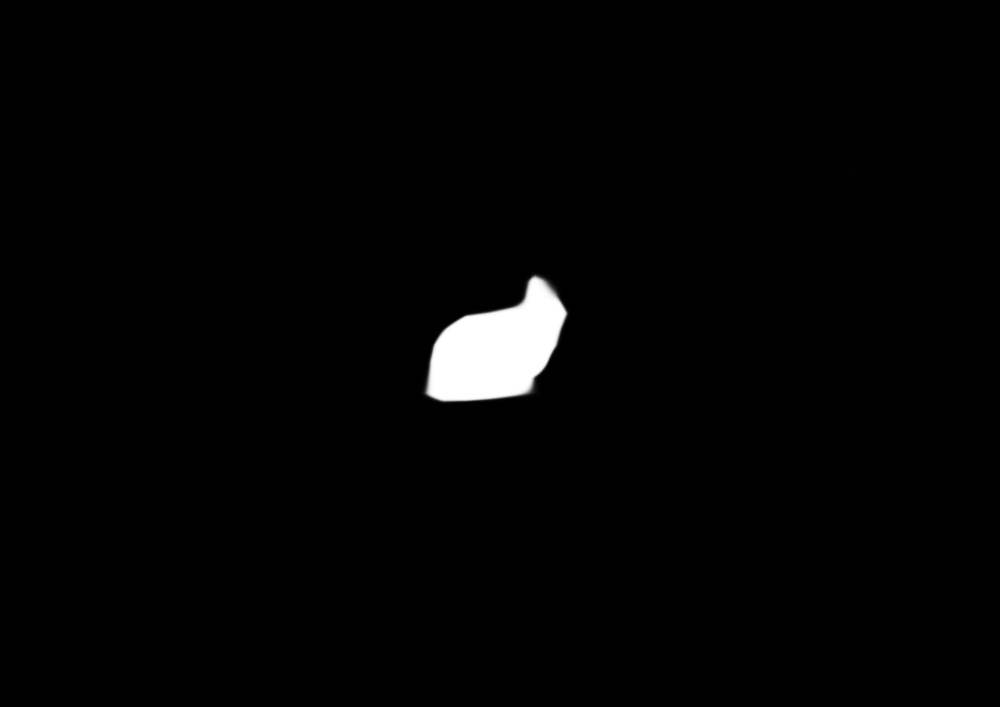}}&
    {\includegraphics[width=0.150\linewidth]{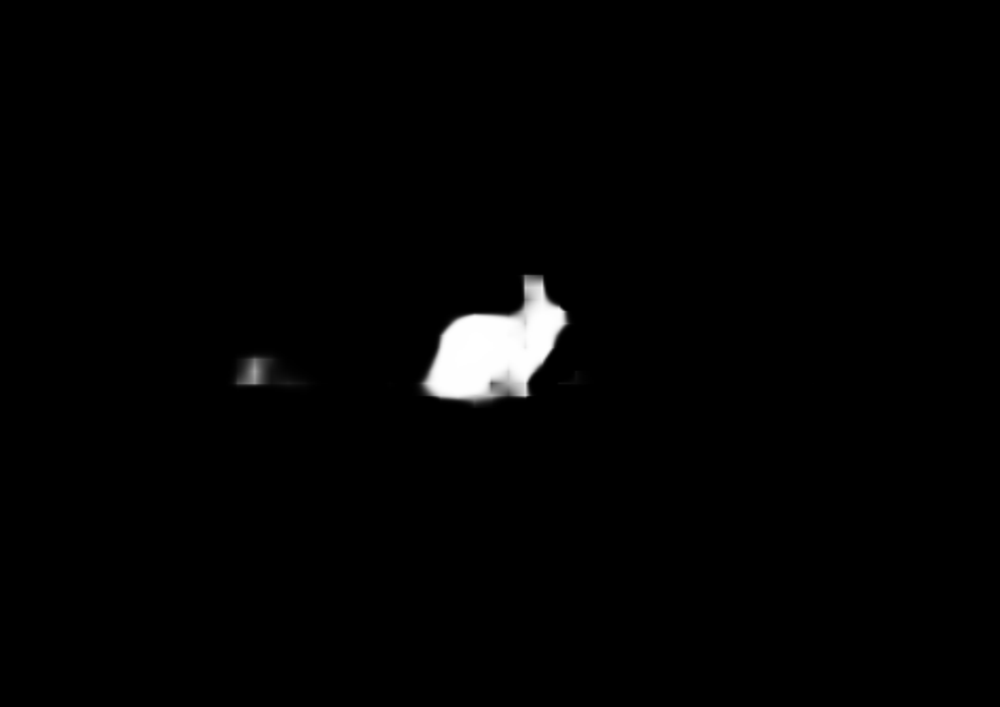}}&
    {\includegraphics[width=0.150\linewidth]{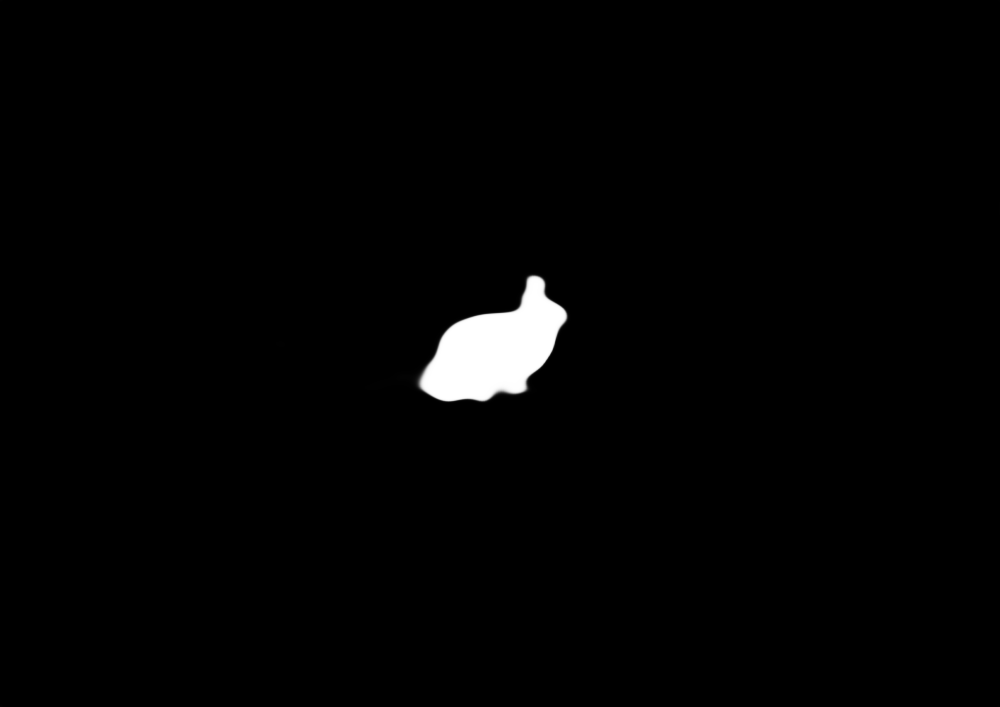}}\\
    \footnotesize{Image}& \footnotesize{GT}&\footnotesize{LSR~\cite{yunqiu_cod21}}&\footnotesize{\cite{fan2021concealed}}&\footnotesize{\cite{zhai2021Mutual}}&\footnotesize{LSR+}\\
   \end{tabular}
   \end{center}
   \caption{Visualization of predictions of our COD base model (\enquote{LSR+}) and benchmark COD models.}
\label{fig:cod_pred_visualization}
\end{figure}

\begin{figure}[!htp]
   \begin{center}
   \begin{tabular}{c@{ } c@{ } c@{ } c@{ } c@{ } c@{ }c@{ }}
      {\includegraphics[width=0.125\linewidth, height=0.100\linewidth]{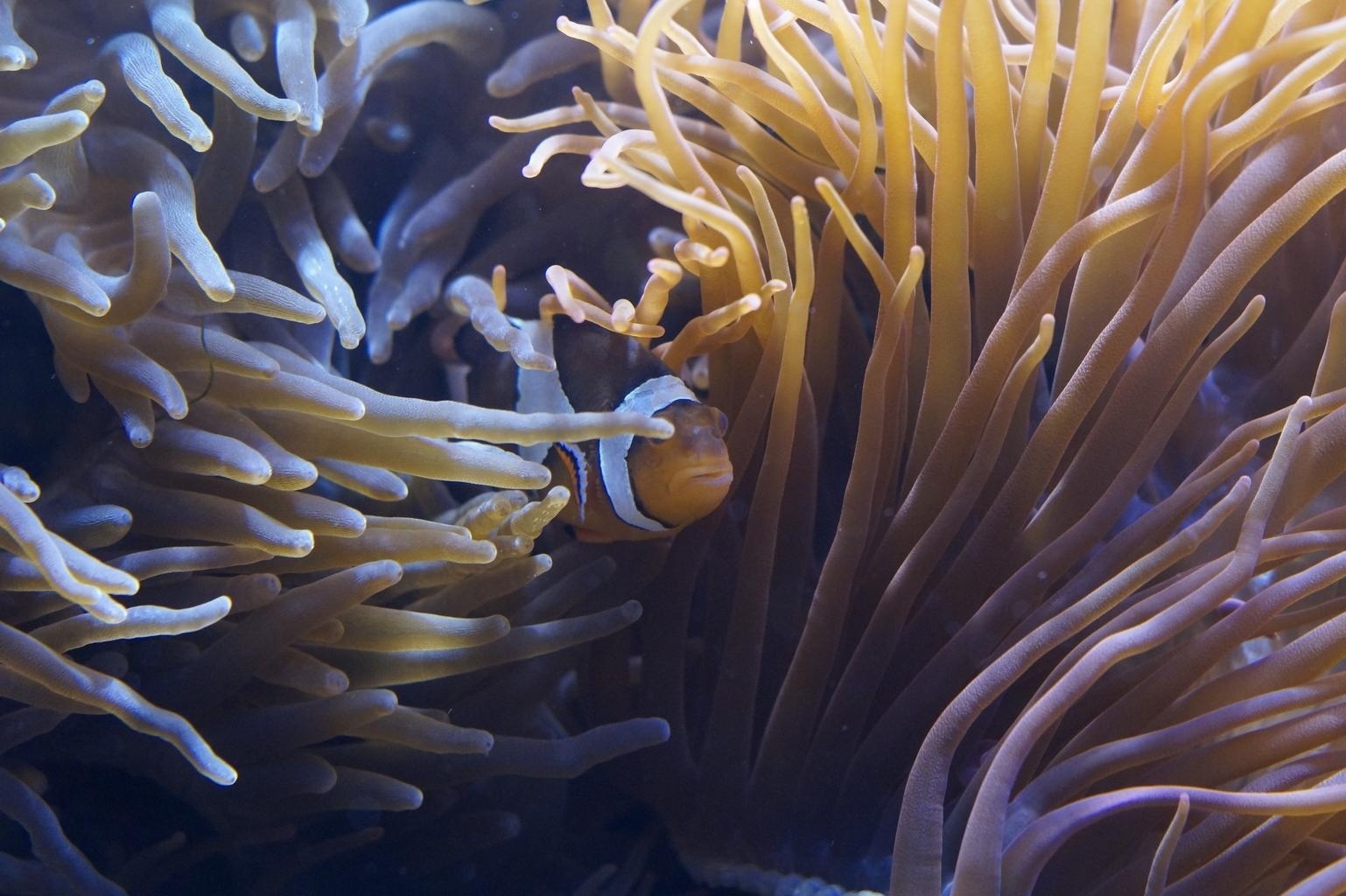}}&
    {\includegraphics[width=0.125\linewidth, height=0.100\linewidth]{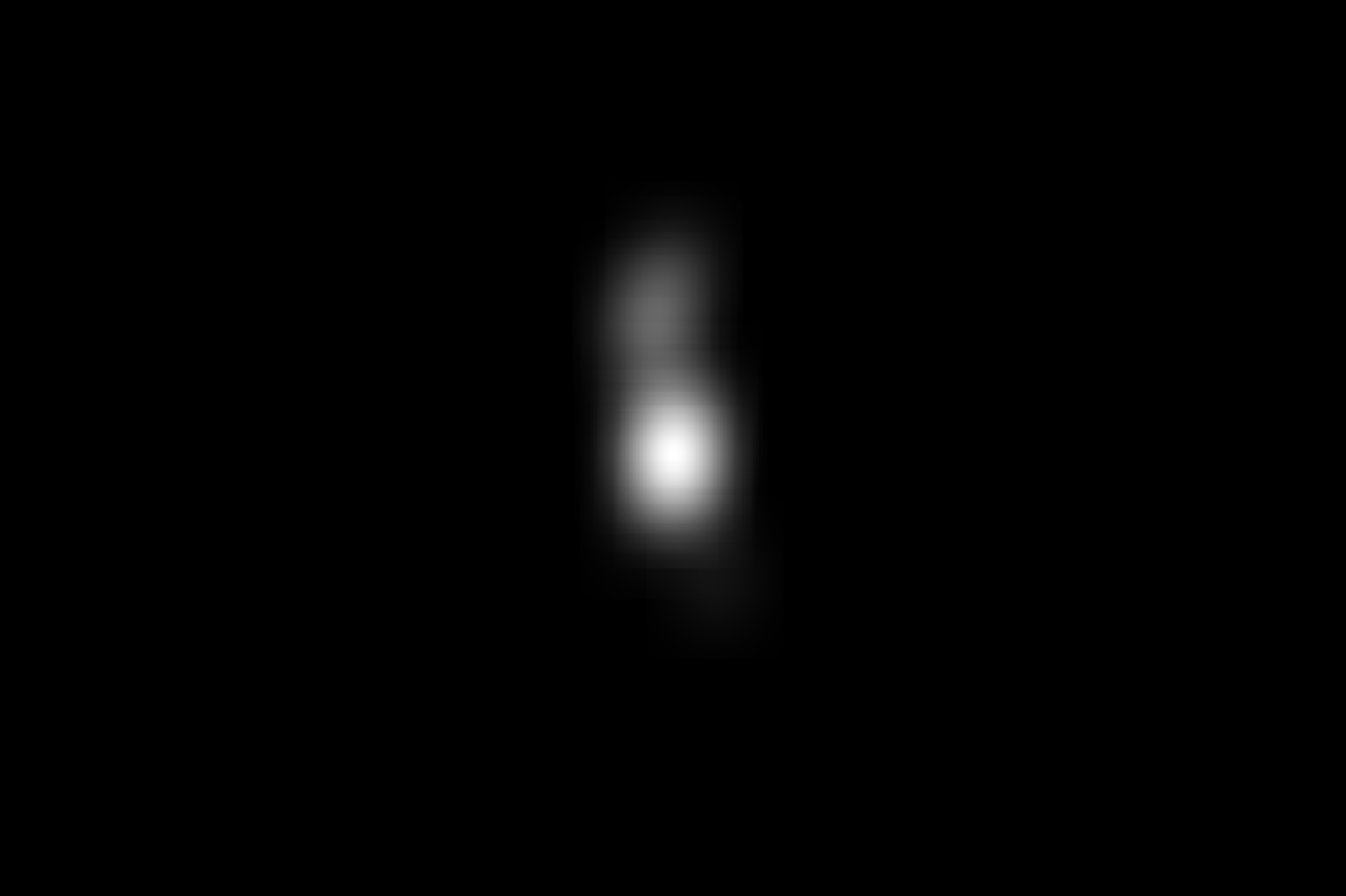}}&
    {\includegraphics[width=0.125\linewidth, height=0.100\linewidth]{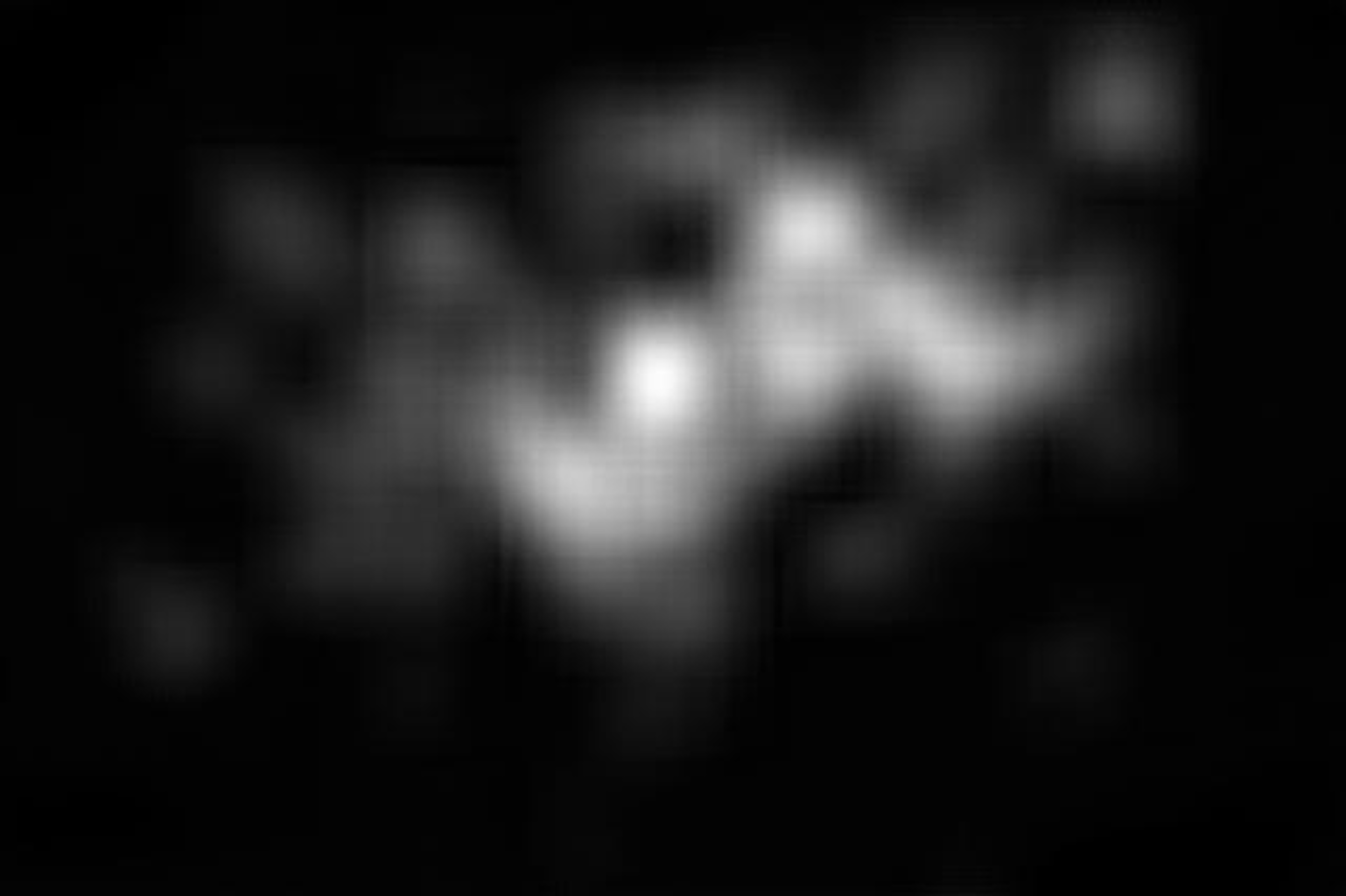}}&
    {\includegraphics[width=0.125\linewidth, height=0.100\linewidth]{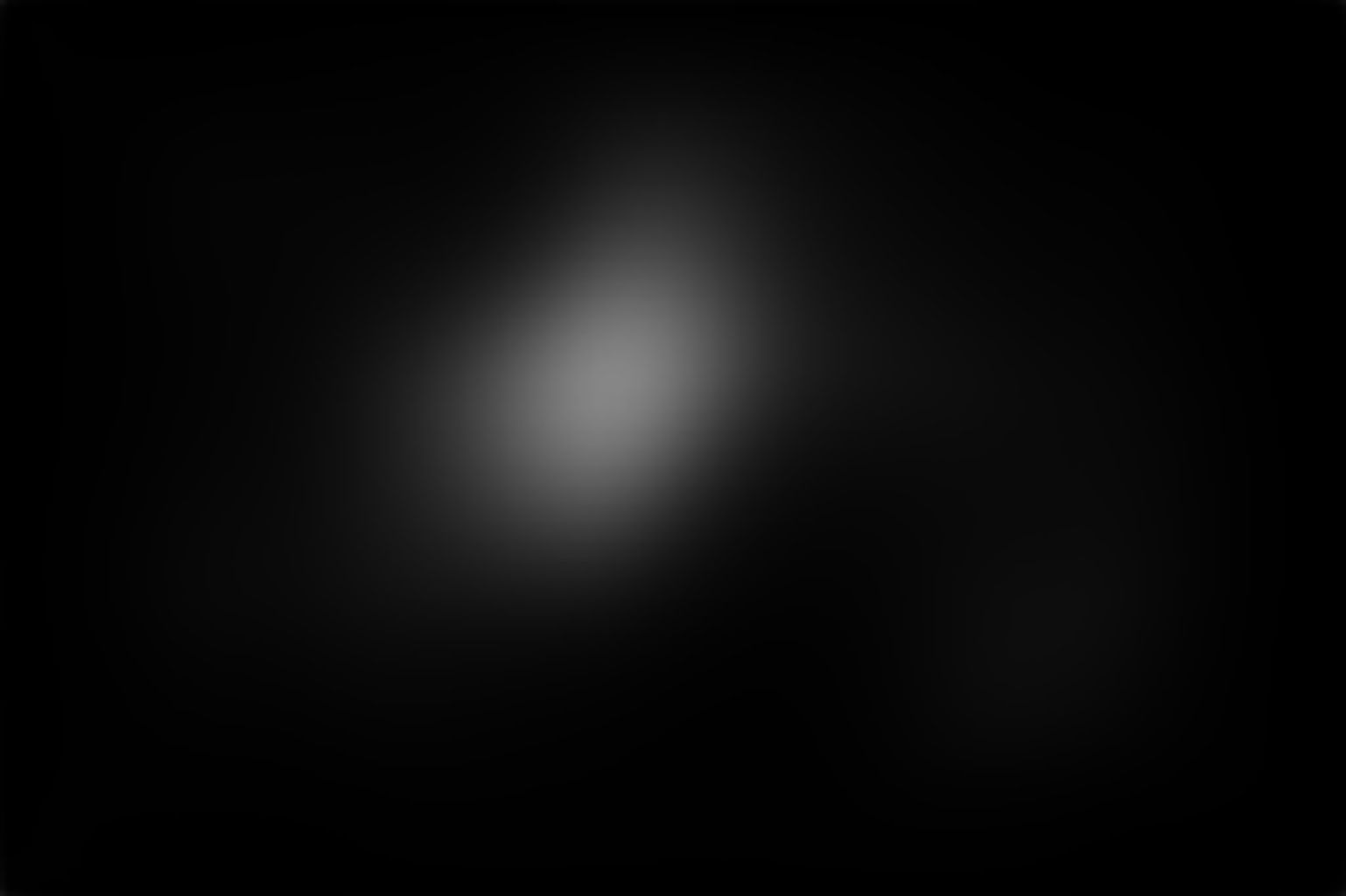}}&
    {\includegraphics[width=0.125\linewidth, height=0.100\linewidth]{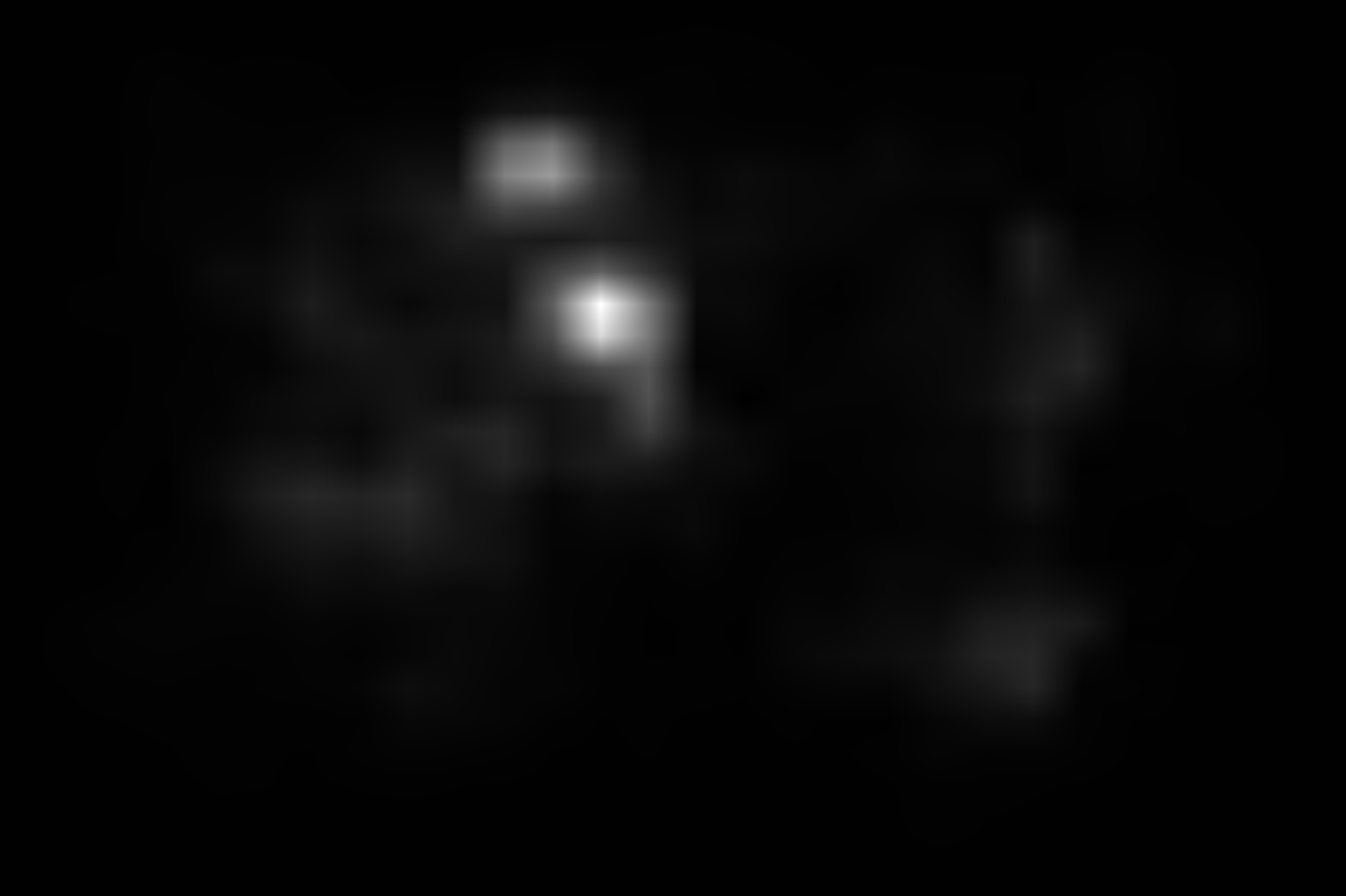}}&
    {\includegraphics[width=0.125\linewidth, height=0.100\linewidth]{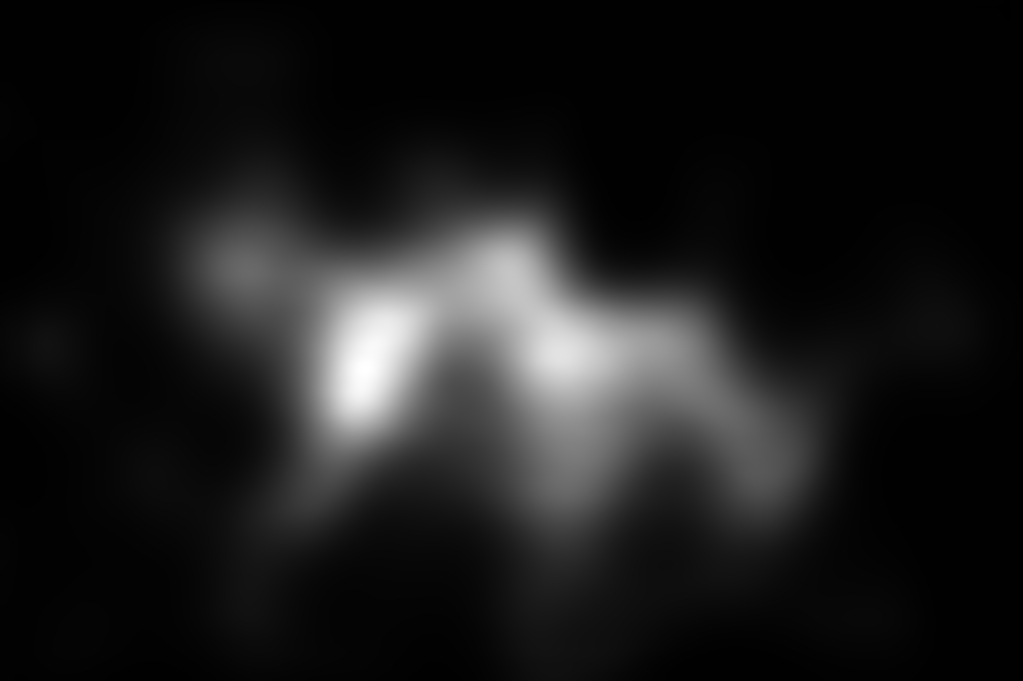}}&
    {\includegraphics[width=0.125\linewidth, height=0.100\linewidth]{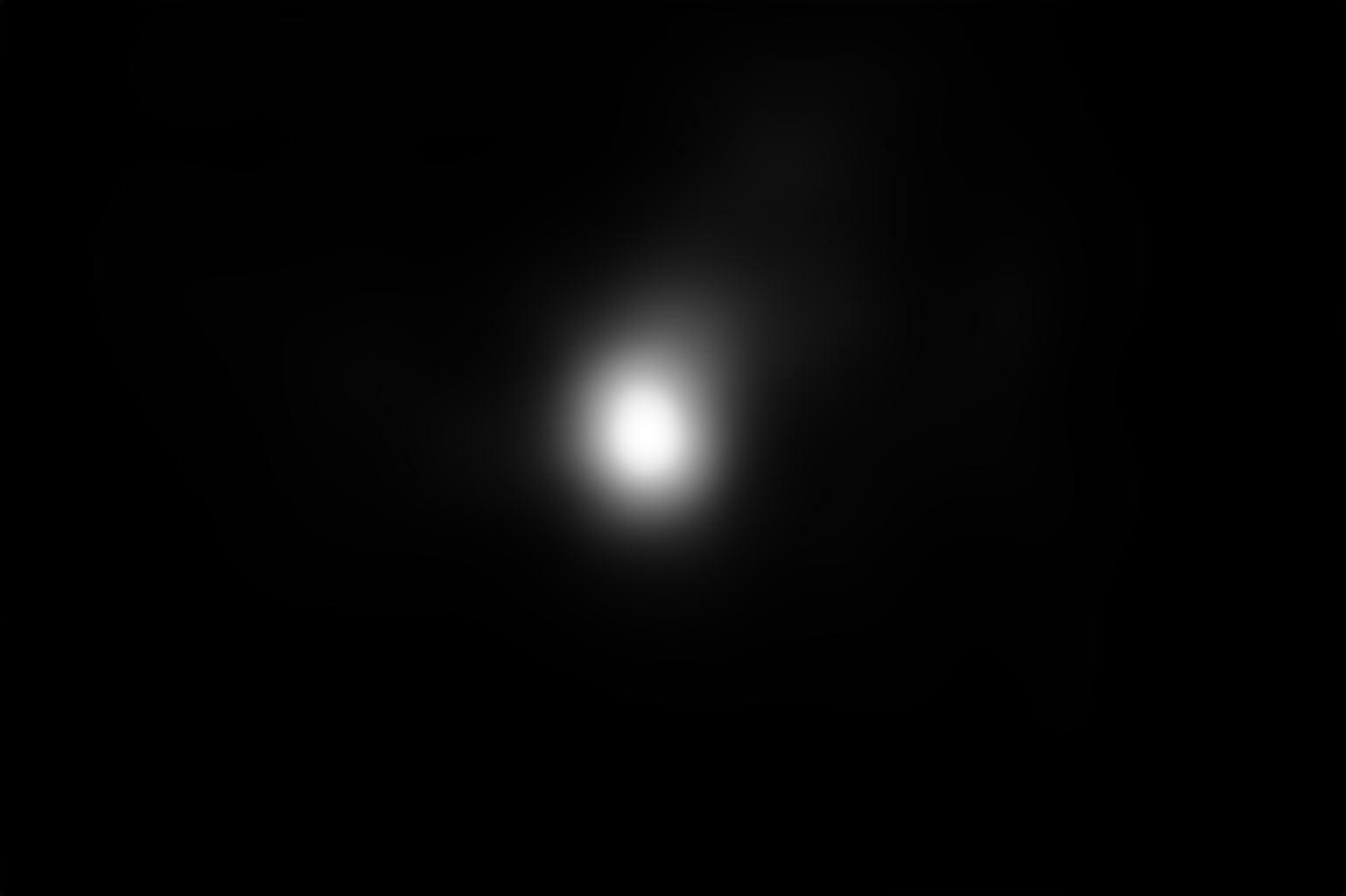}}\\
    {\includegraphics[width=0.125\linewidth, height=0.100\linewidth]{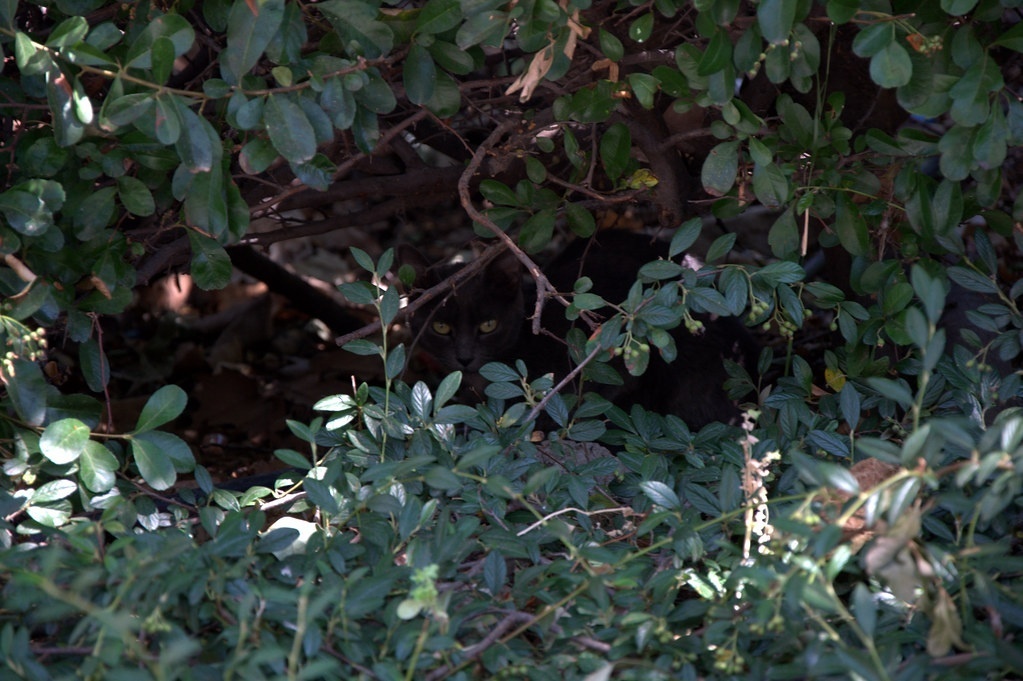}}&
    {\includegraphics[width=0.125\linewidth, height=0.100\linewidth]{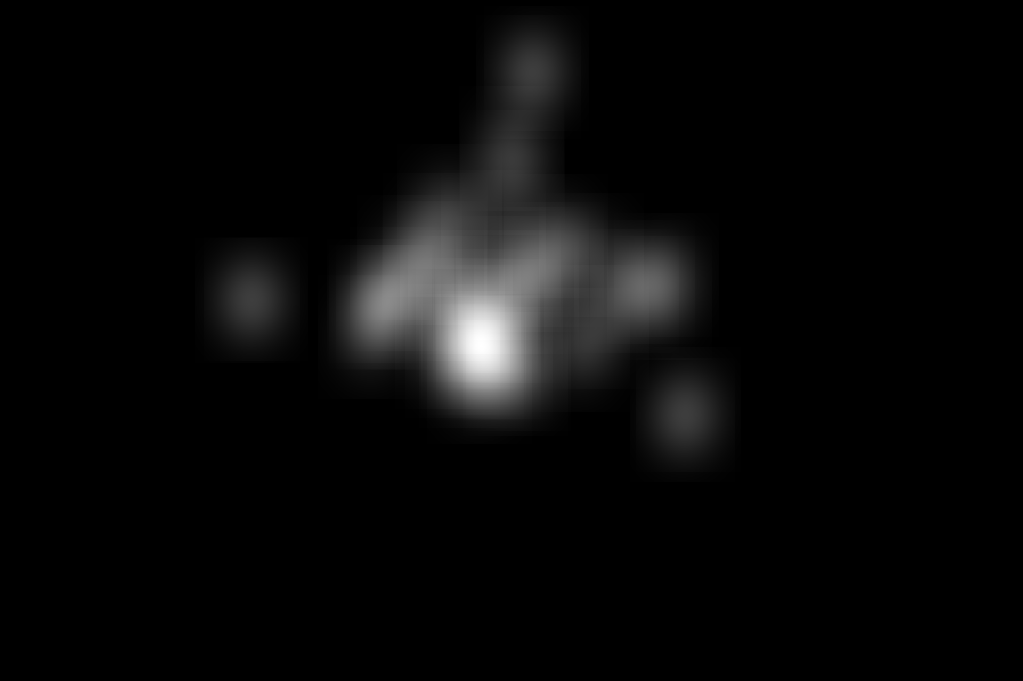}}&
    {\includegraphics[width=0.125\linewidth, height=0.100\linewidth]{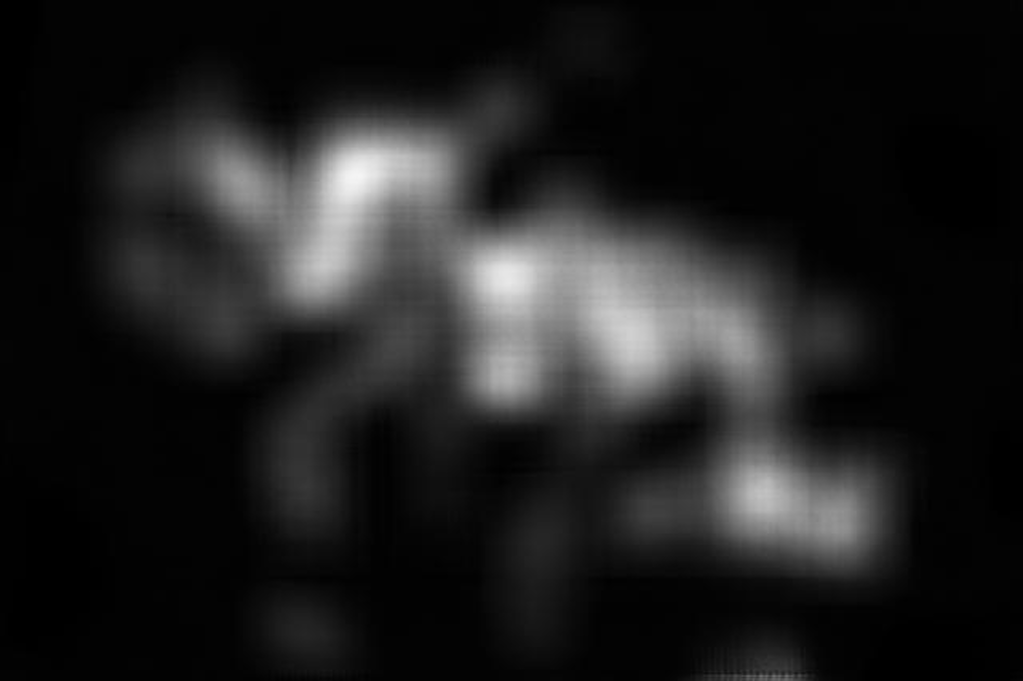}}&
    {\includegraphics[width=0.125\linewidth, height=0.100\linewidth]{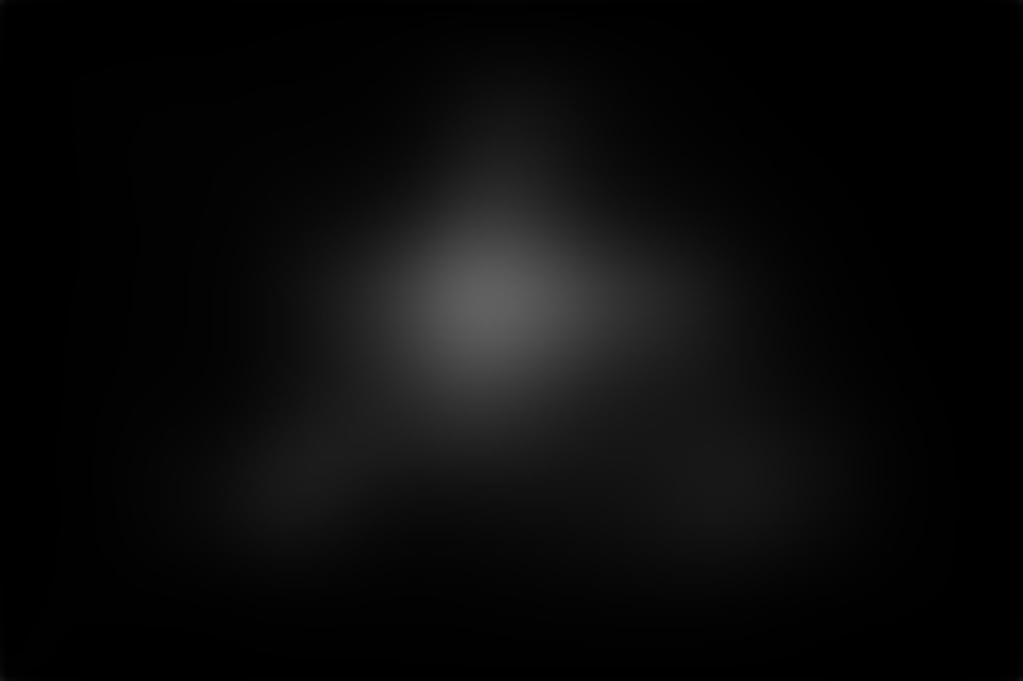}}&
    {\includegraphics[width=0.125\linewidth, height=0.100\linewidth]{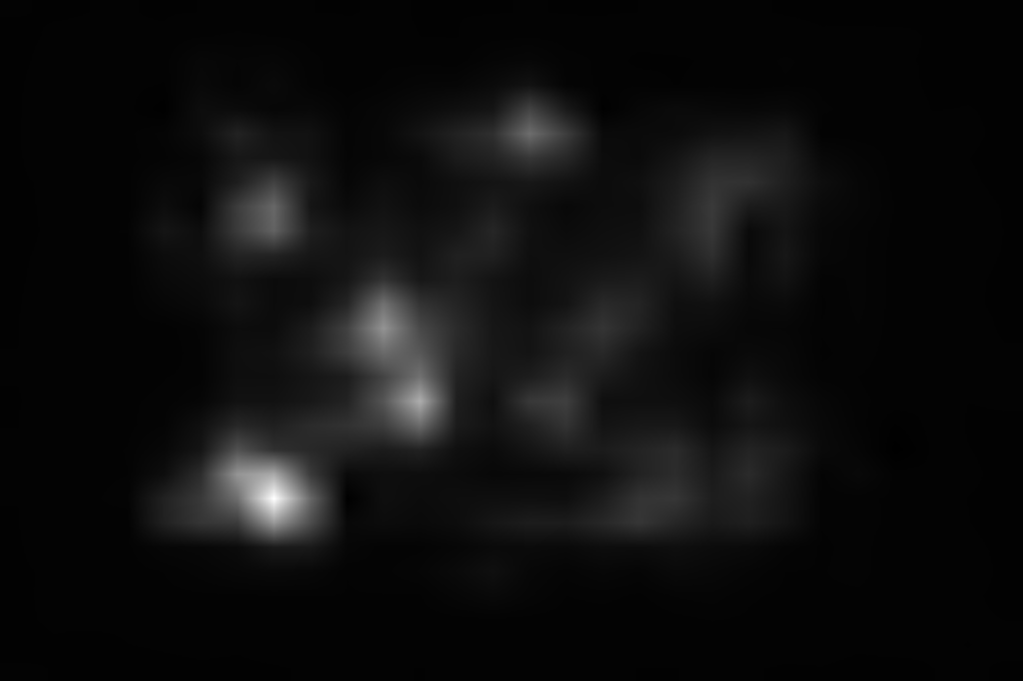}}&
    {\includegraphics[width=0.125\linewidth, height=0.100\linewidth]{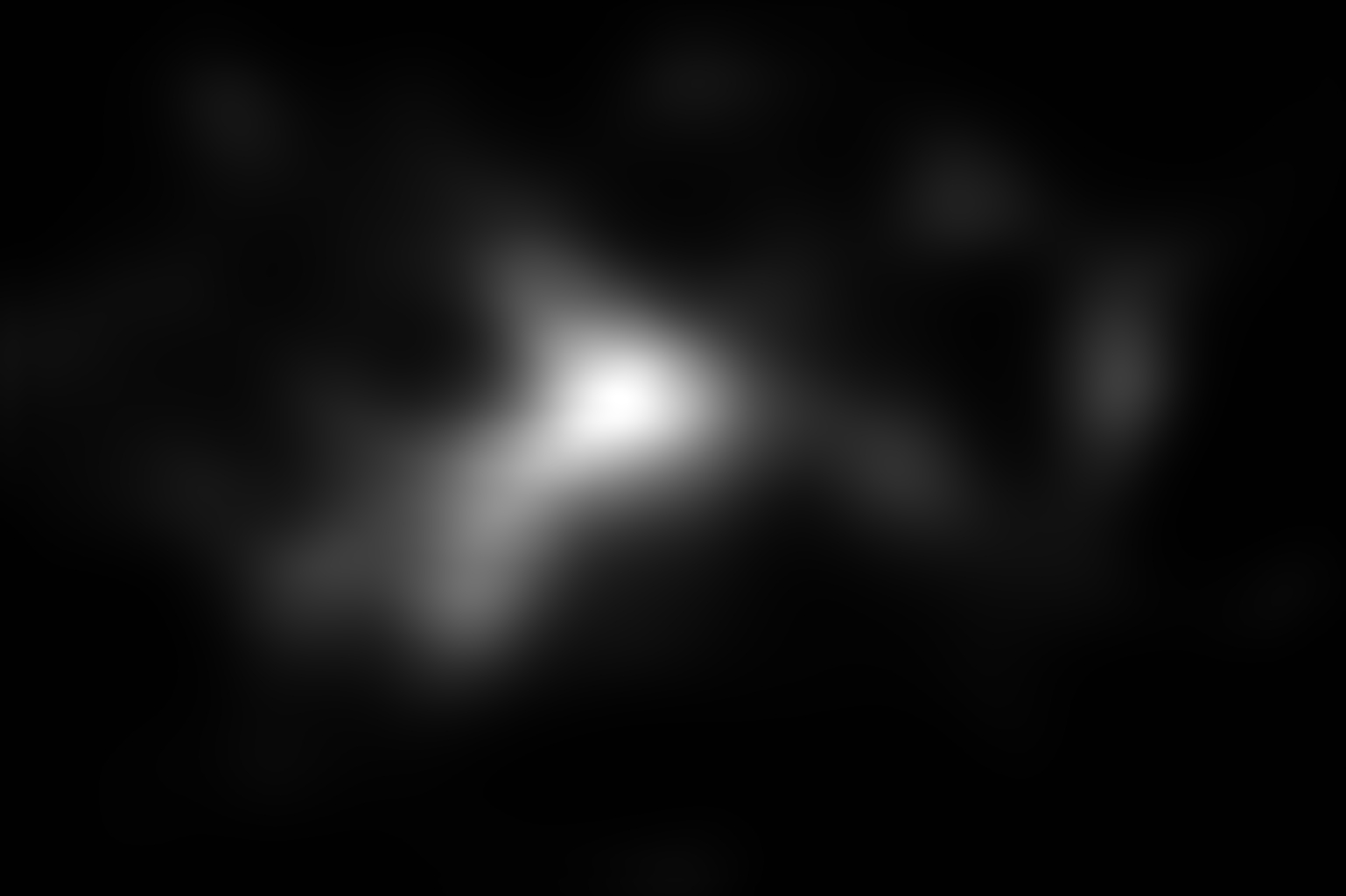}}&
    {\includegraphics[width=0.125\linewidth, height=0.100\linewidth]{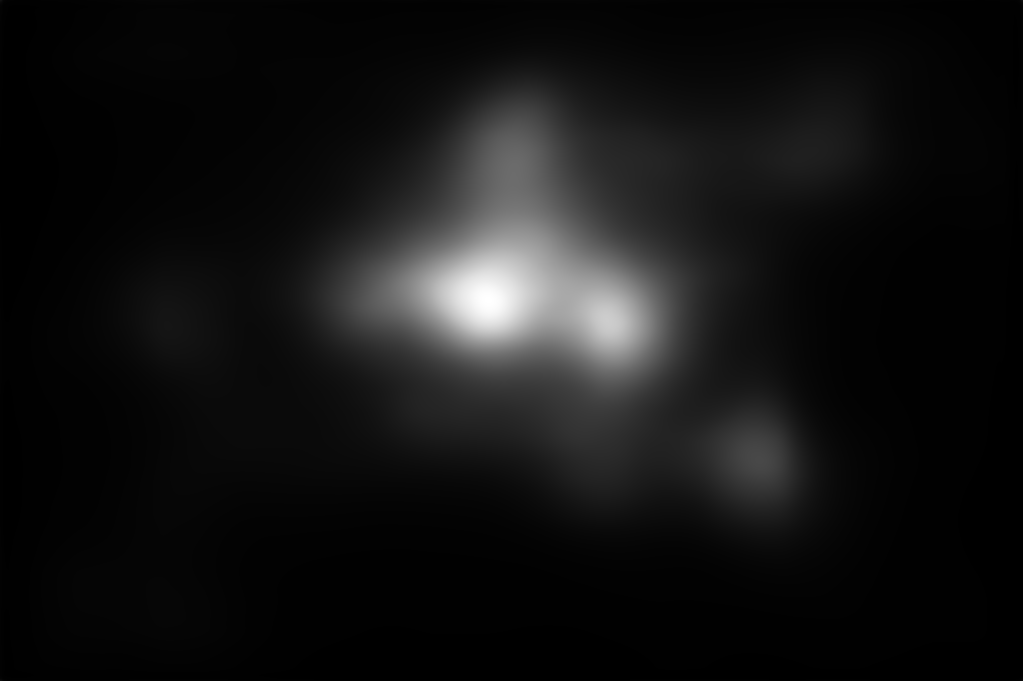}}\\
    \footnotesize{Image}& \footnotesize{GT}&\footnotesize{\cite{che2019gaze}}&\footnotesize{\cite{pan2017salgan}}&\footnotesize{\cite{he2019understanding}}&\footnotesize{\cite{reddy2020tidying}}&\footnotesize{LSR+}\\
   \end{tabular}
   \end{center}
   \caption{Visualization of predictions of the re-constructed benchmark COL models and our COL base model (\enquote{LSR+}).
   }
\label{fig:col_pred_visualization}
\end{figure}

For the ranking task, \cite{amirul2018revisiting} introduces the Salient Object Ranking (SOR) metric to measure ranking performance, which is defined as the Spearman’s Rank-Order Correlation between the ground truth rank order and the predicted rank order of salient objects. However, it cannot be used in our scenario because saliency ranking is concerned primarily about the construction of relative relationships of saliency degrees for the objects in the same image, while the relative relationship between camouflage levels for the objects should be evaluated over the entire dataset.

For discriminative region localization, we adopt the widely used fixation prediction evaluation metrics, including the similarity ($SIM$) metric \cite{judd2012benchmark},
the Linear Correlation Coefficient ($CC$) \cite{le2007predicting},
the normalized scanpath saliency ($NSS$) \cite{peters2005components},
the Earth Mover’s Distance ($EMD$) \cite{rubner2000earth},
Kullback–Leibler Divergence ($KLD$) \cite{kullback1951information},
and AUC\_Judd ($AUC\_J$) \cite{judd2009learning}.
\noindent\textbf{Benchmark models preparation:}
For the discriminative region localization task, as it has a
similar label generation process, we re-implement existing eye fixation prediction models \cite{jiang2015salicon}, which are then treated as benchmark for
camouflaged object localization.
As there are
no camouflaged object ranking models, we then implement three rank or instance based object segmentation methods for camouflage rank estimation, including
PPA \cite{fang2021salient}
for salient ranking prediction, SOLOv2 \cite{wang2020SOLOv2} and Mask Scoring-RCNN (MS-RCNN) \cite{liu2018path} for instance segmentation. 


\subsection{Base Models for Performance Comparison}
\label{subsec:base_models}
As existing COD models are trained with only COD training datasets, and the re-trained COL and COR models are trained only with the corresponding training dataset, for fair comparison, we report the base model performance (\enquote{LSR+}) of each task in Table \ref{tab:benchmark_model_comparison} (for COD), Table \ref{tab:fixation_baseline_cod10k} (for COL) and Table \ref{tab:ranking_comparison} (for COR) respectively.
To obtain the base models, given the backbone feature $\{s_k\}_{k=1}^4$, we adopt the decoder from Fig.~\ref{fig:joint_cod_fixation}, and design a COD model and COL model, where the former is trained with the pixel position aware loss ($\mathcal{L}_c$ in Eq.~\ref{dual_loss}) and the latter is trained with L2 loss ($\mathcal{L}_2$ in Eq.~\ref{dual_loss}). The COR model can then be trained with the weighted rank loss $\mathcal{L}'_{rank}$.

\begin{figure}[!htp]
   \begin{center}
   \begin{tabular}{c@{ } c@{ } c@{ } c@{ } c@{ } c@{ } c@{ }}
      {\includegraphics[width=0.125\linewidth, height=0.100\linewidth]{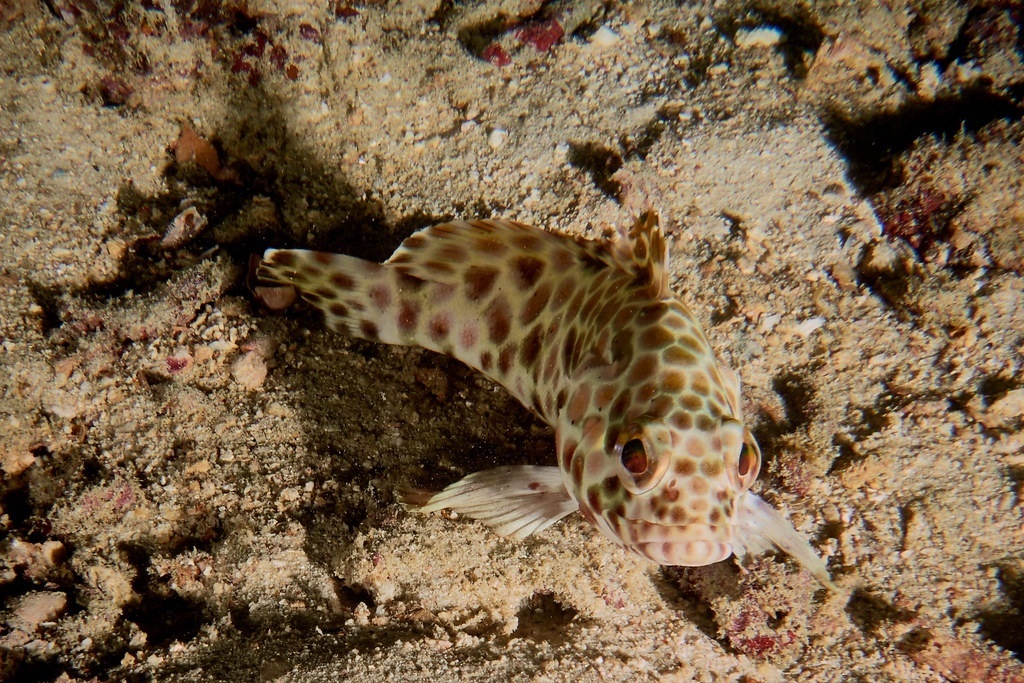}}&
      {\includegraphics[width=0.125\linewidth, height=0.100\linewidth]{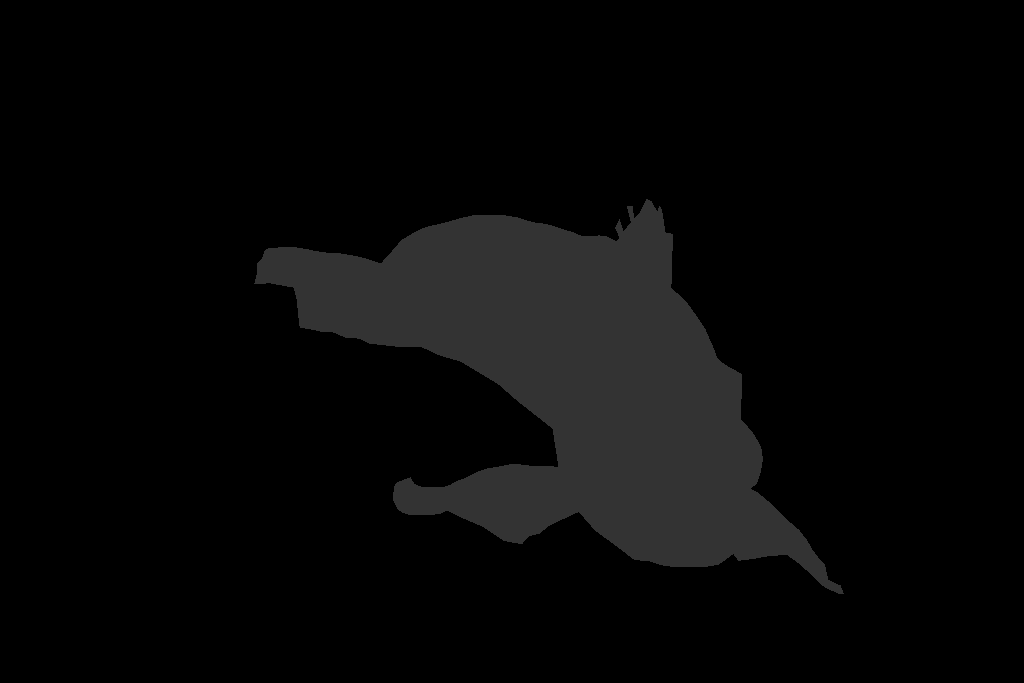}}&
    {\includegraphics[width=0.125\linewidth, height=0.100\linewidth]{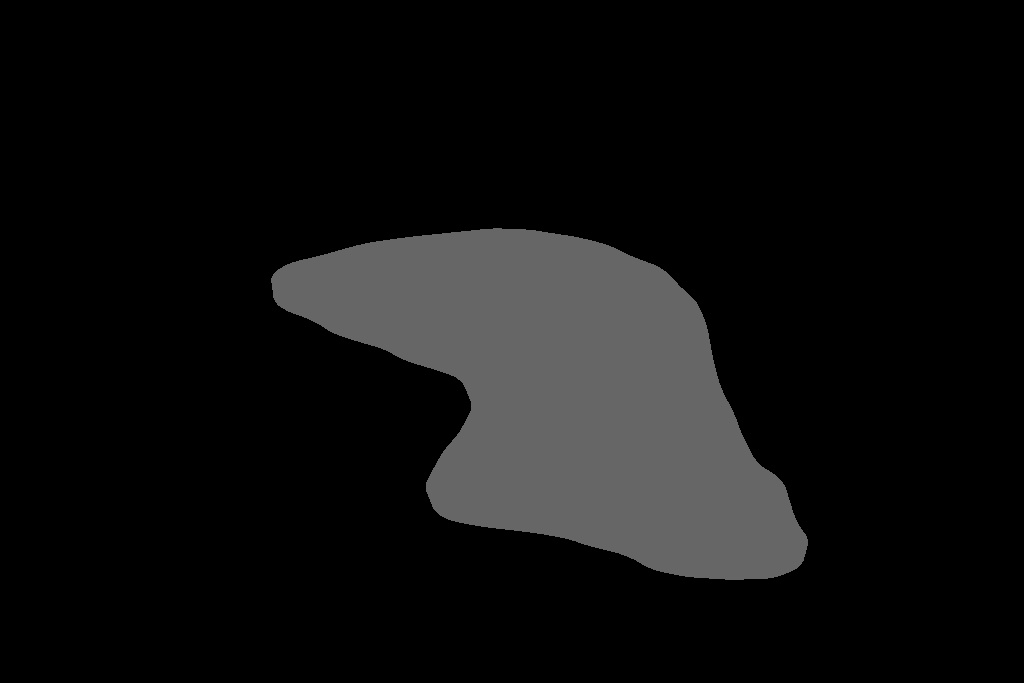}}&
    {\includegraphics[width=0.125\linewidth, height=0.100\linewidth]{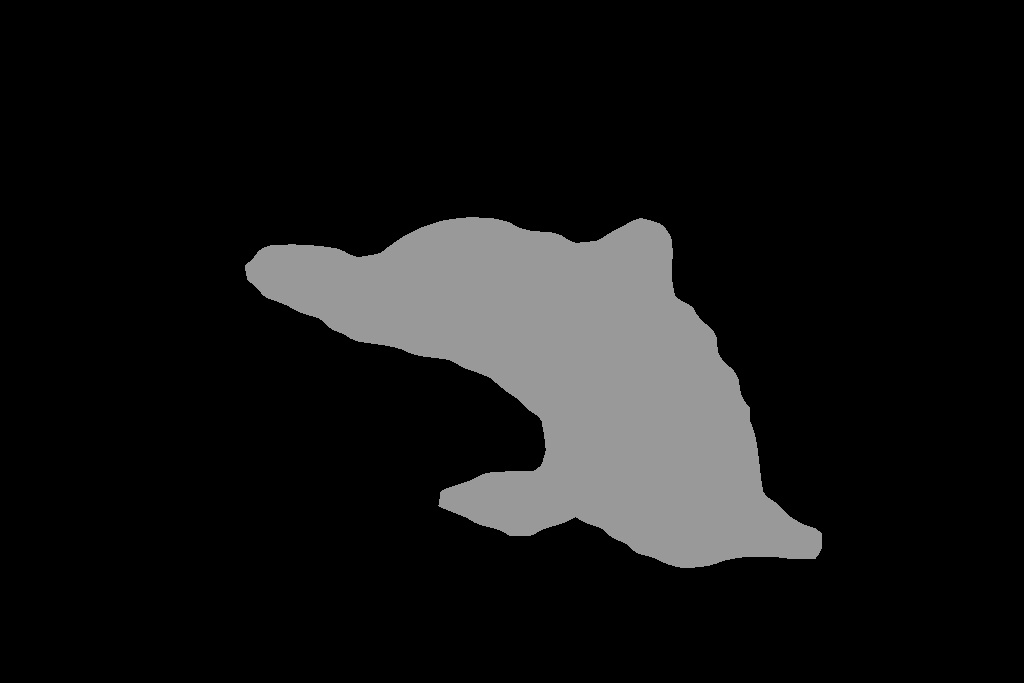}}&
    {\includegraphics[width=0.125\linewidth, height=0.100\linewidth]{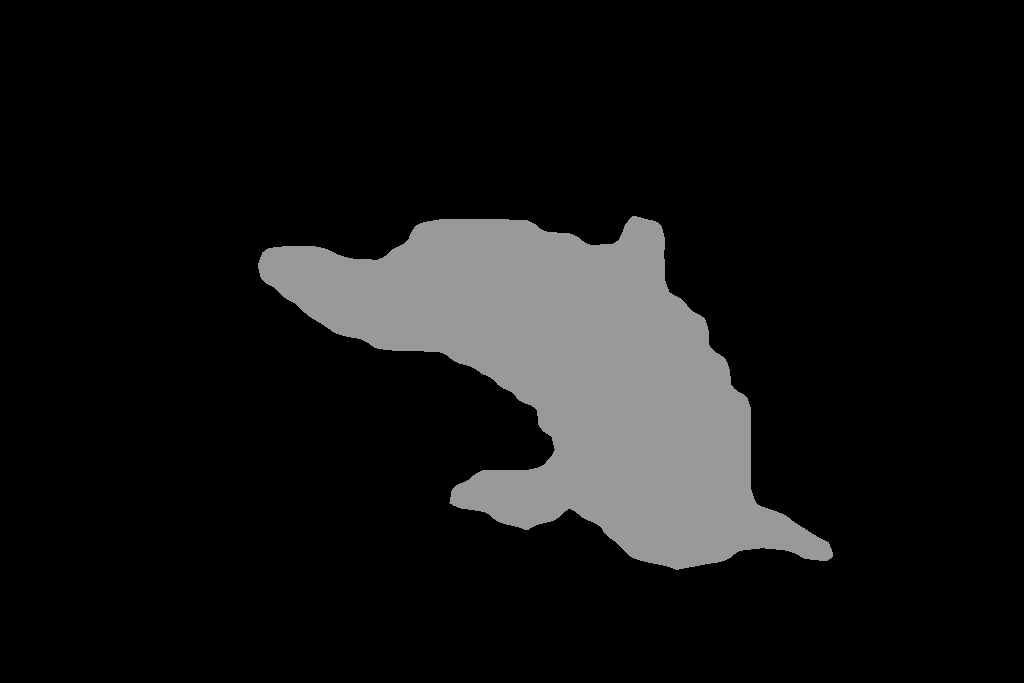}}&
    {\includegraphics[width=0.125\linewidth, height=0.100\linewidth]{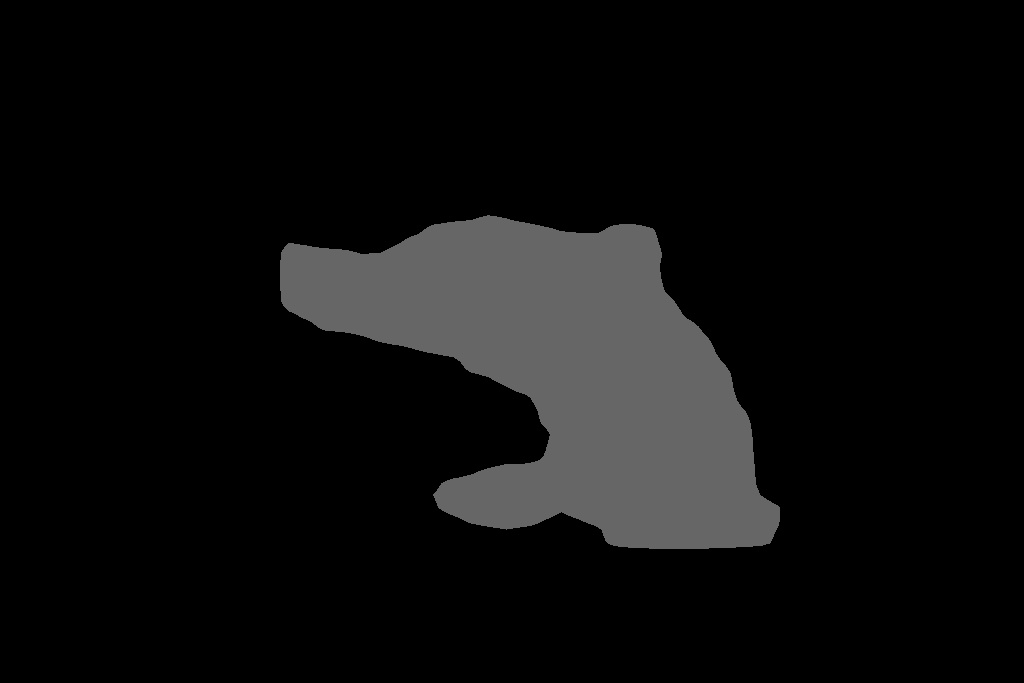}}&
    {\includegraphics[width=0.125\linewidth, height=0.100\linewidth]{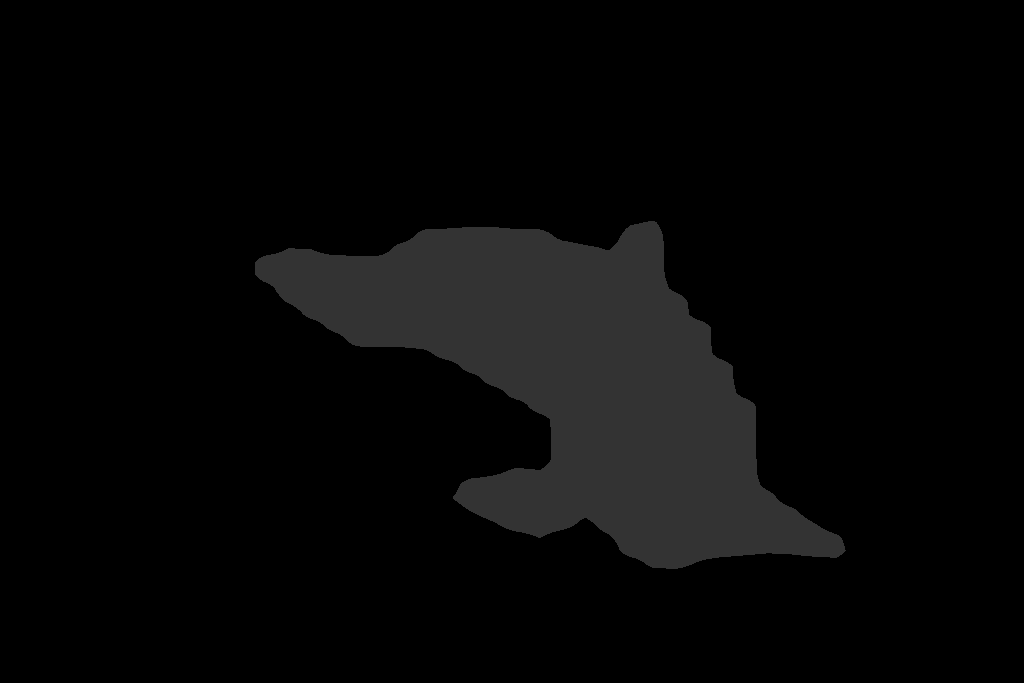}}\\
    {\includegraphics[width=0.125\linewidth, height=0.100\linewidth]{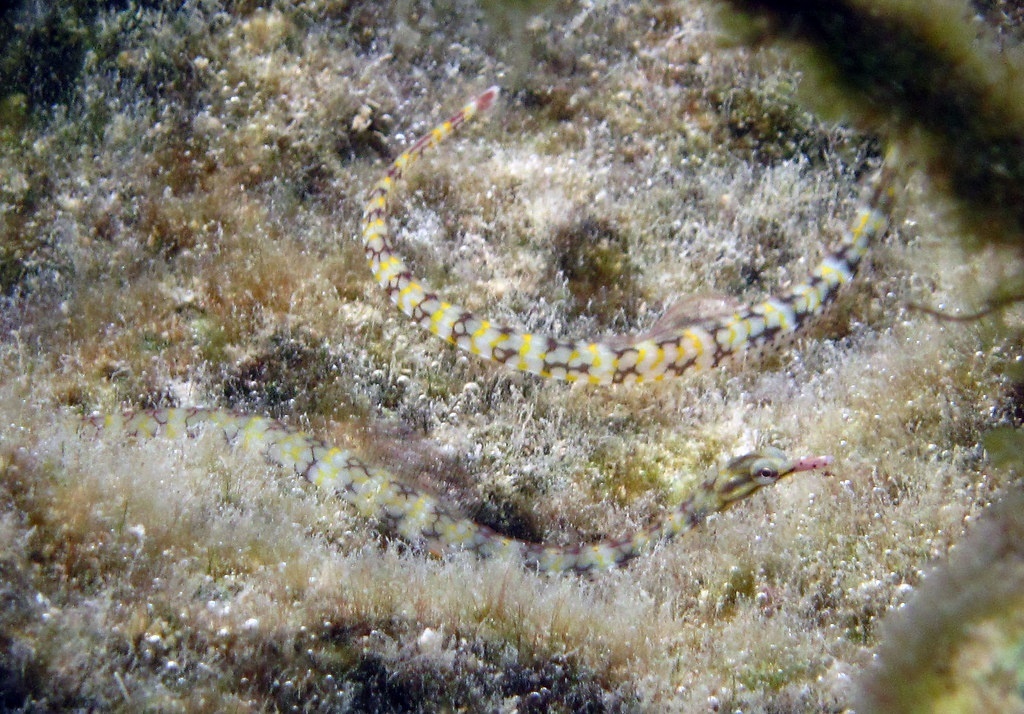}}&
    {\includegraphics[width=0.125\linewidth, height=0.100\linewidth]{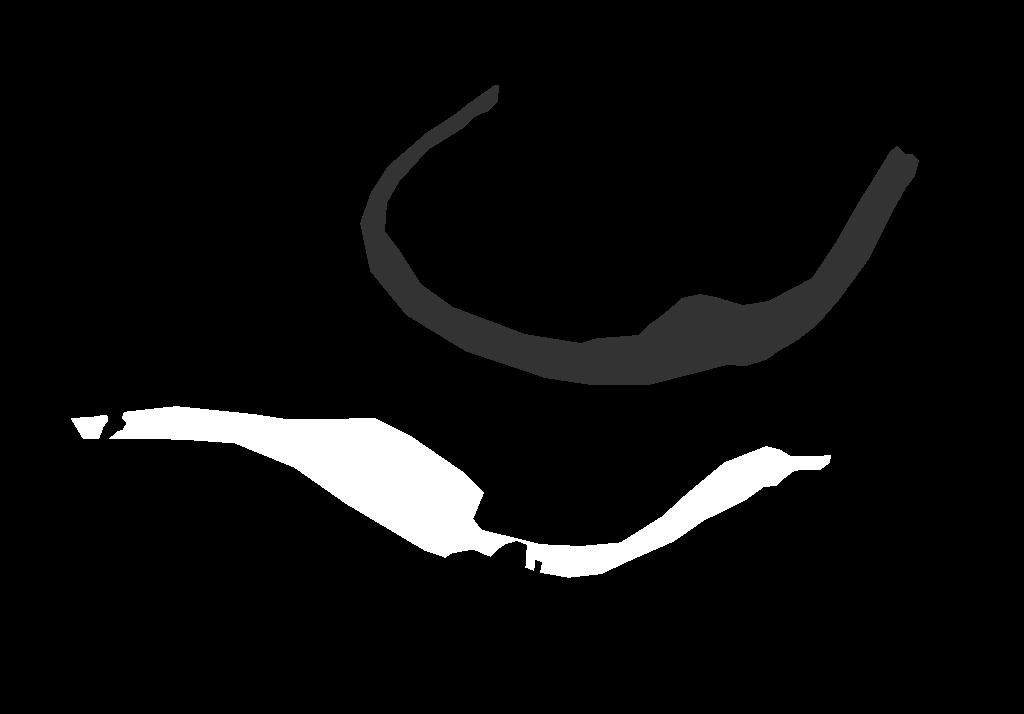}}&
    {\includegraphics[width=0.125\linewidth, height=0.100\linewidth]{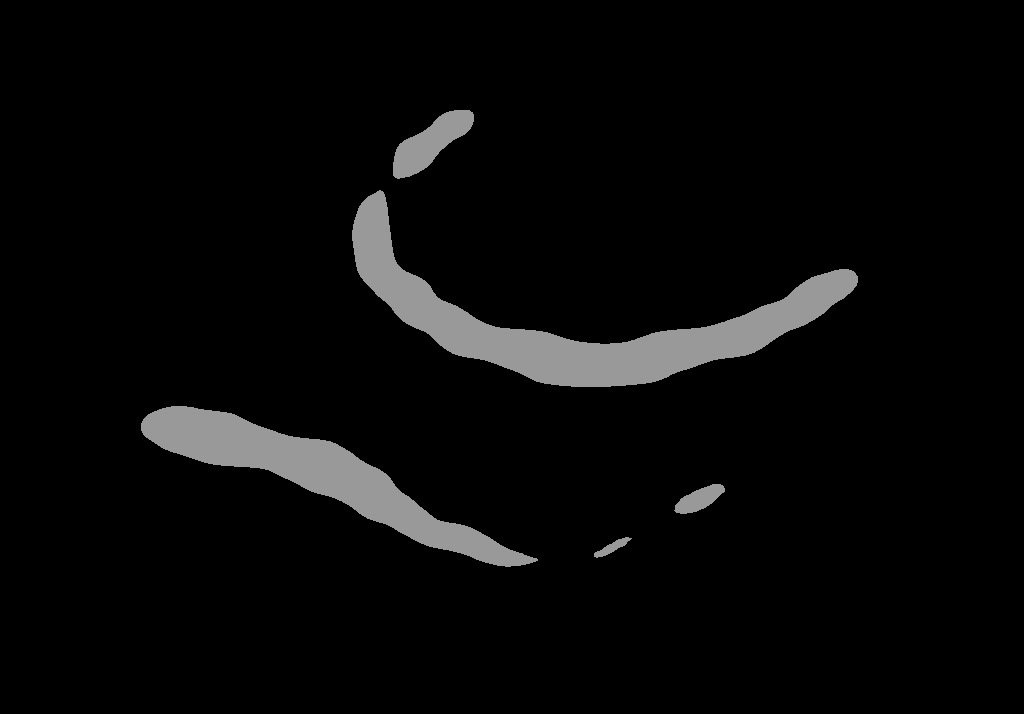}}&
    {\includegraphics[width=0.125\linewidth, height=0.100\linewidth]{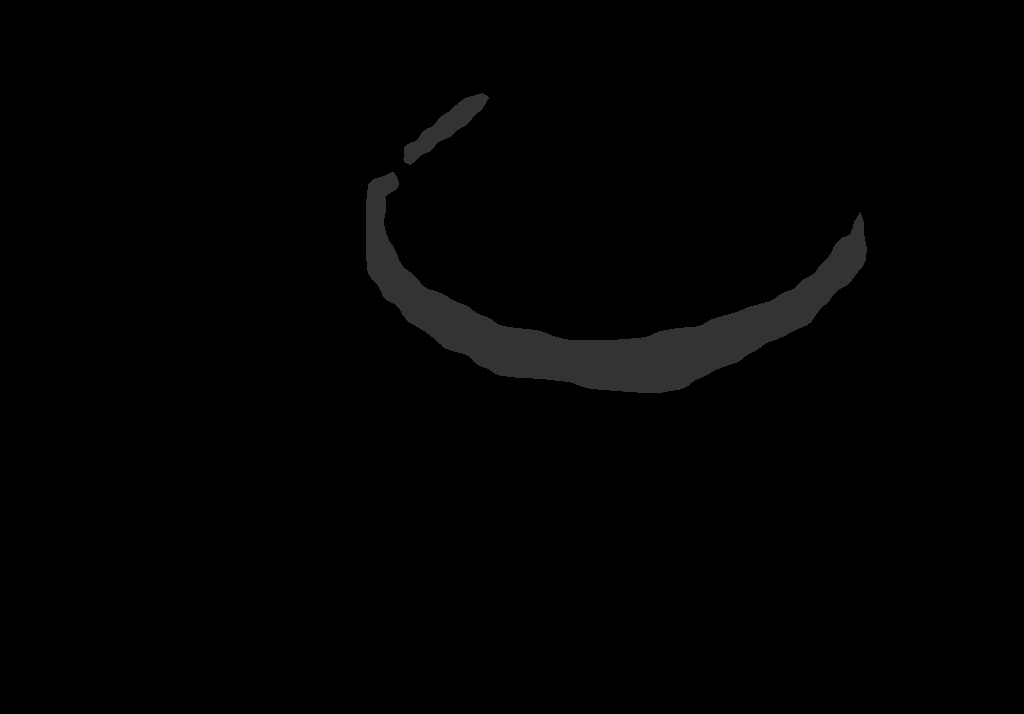}}&
    {\includegraphics[width=0.125\linewidth, height=0.100\linewidth]{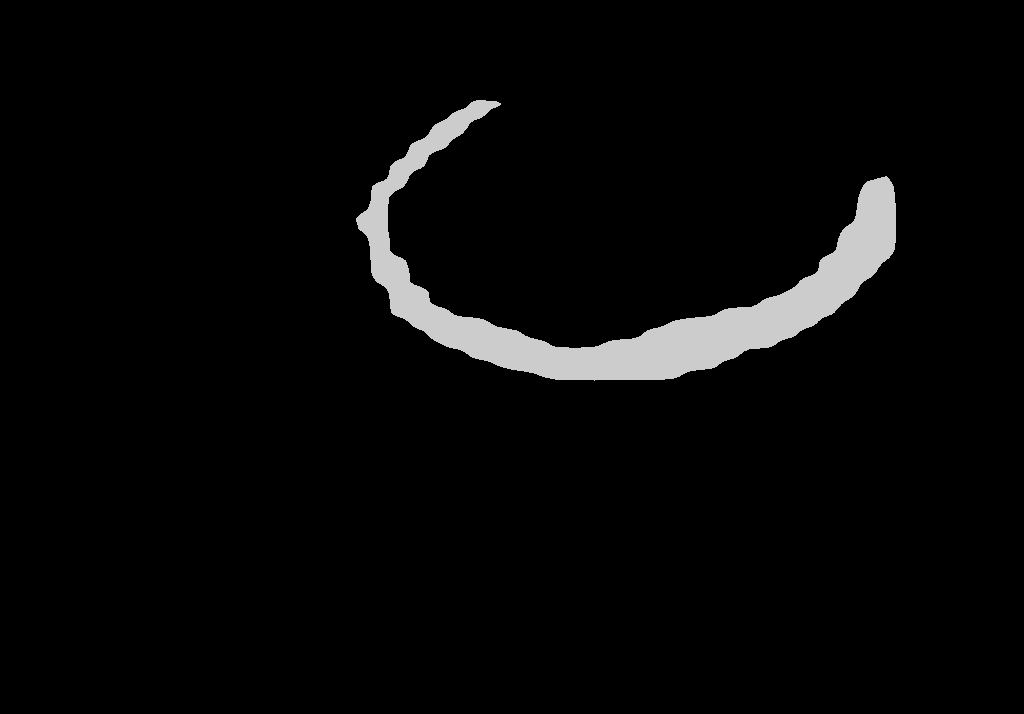}}&
    {\includegraphics[width=0.125\linewidth, height=0.100\linewidth]{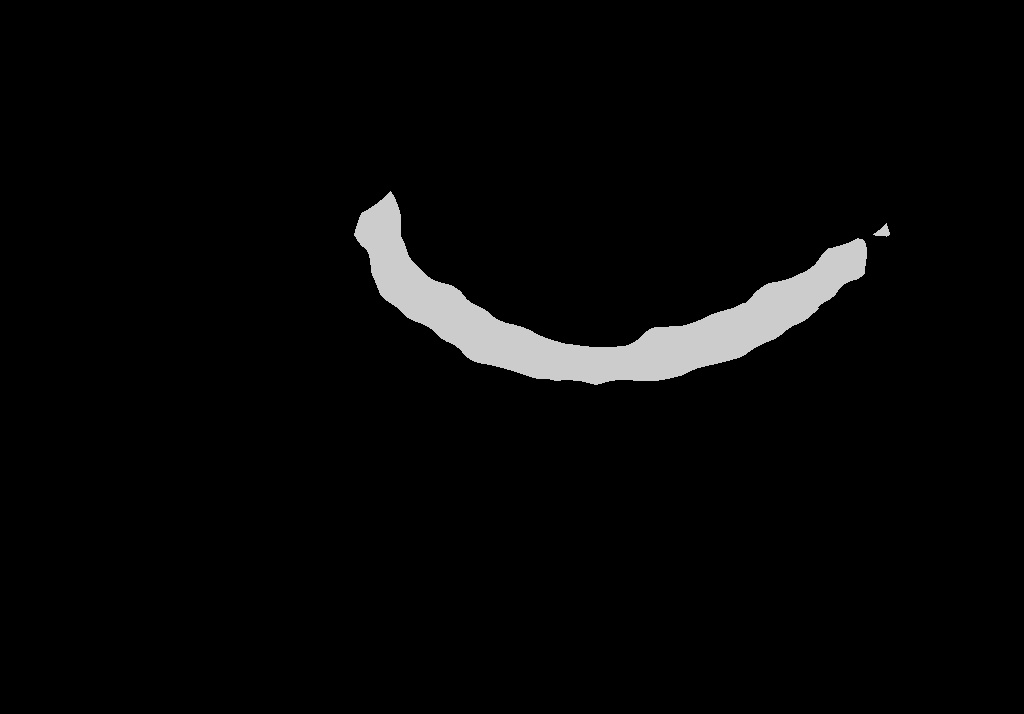}}&
    {\includegraphics[width=0.125\linewidth, height=0.100\linewidth]{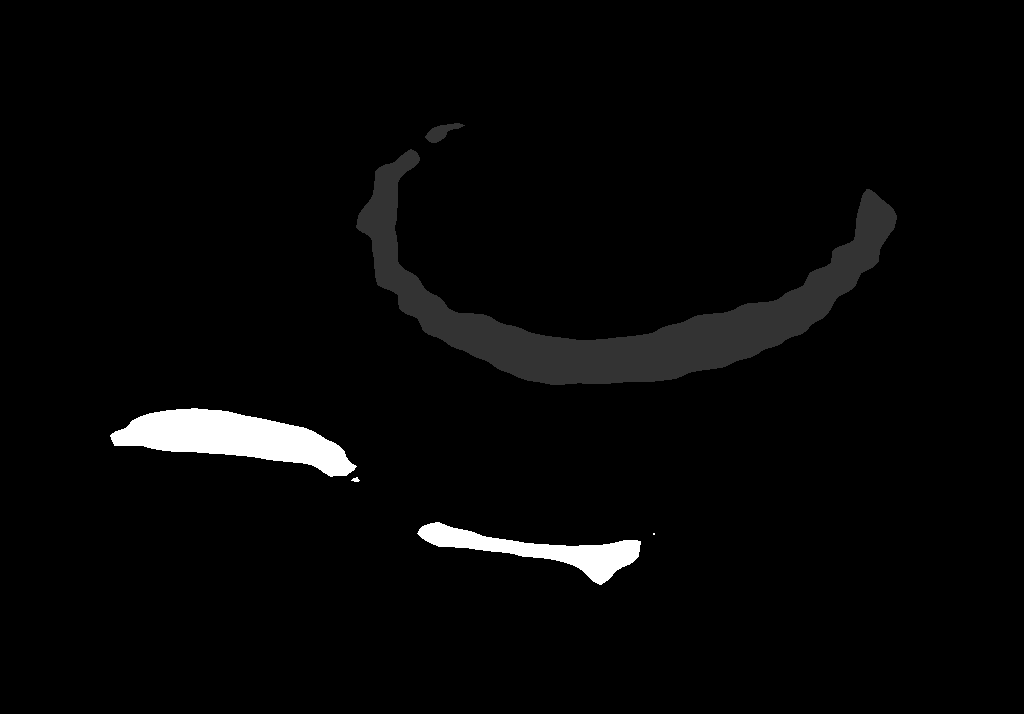}}\\
    \footnotesize{Image}& \footnotesize{GT}&\footnotesize{\cite{wang2020SOLOv2}}&\footnotesize{\cite{liu2018path}}&\footnotesize{\cite{he2017mask}}&\footnotesize{\cite{fang2021salient}
    }&\footnotesize{LSR+}\\
   \end{tabular}
   \end{center}
   \caption{Visualization of predictions of the re-constructed benchmark COR models and our COR base model (\enquote{LSR+}).}
\label{fig:cor_pred_visualization}
\end{figure}

\subsection{Performance Comparison}
Note that our performance in this section is based on the base models discussed in Section \ref{subsec:base_models}.

\noindent\textbf{Camouflaged object detection:}
We construct the benchmark COD models in Table~\ref{tab:benchmark_model_comparison}, which includes two parts: 1)
models transferred from SOD tasks, and 2)
existing COD models. For the transferred models, we re-train them with benchmark COD training dataset, and keep the other parts of the original model unchanged. For the existing COD models, we use their produced camouflage map directly for performance evaluation\footnote{We retrain ZoomNet \cite{ZoomNet_CVPR2022} with the conventional ResNet50 initialization weights as the conventional methods for fair comparison.}.
Table~\ref{tab:benchmark_model_comparison} shows improved performance of our extension (\enquote{LSR+}) compared with the preliminary version (\enquote{LSR}). Note that the base model of \enquote{LSR} contains 58M parameters, and to achieve efficient model training and testing for the new triple-task learning framework, the base model of \enquote{LSR+} contains 48M parameters. The overall better performance of \enquote{LSR+} compared with \enquote{LSR} indicates effectiveness of our new base COD model.
Further, the competing performance of our base model with existing benchmark models of larger training/testing image sizes validates the effectiveness of our base model. We argue that the resolution of the training dataset is especially important for camouflaged object detection, which will be further explained in Section \ref{subsub:model_analysis}.
We also show the qualitative comparison of our model and existing models in Fig.~\ref{fig:cod_pred_visualization}, \Rev{we use background shadow and bold text to highlight the best result and use only bold text to emphasize the second best result. Compared with other model uses the ResNet50 backbone, we can see that our ``LSR+'' model could achieve the best or the second best performance in in most metrics, which is competitive to the state-of-the-art ZoomNet~\cite{ZoomNet_CVPR2022} model. We also add the experiments with different backbone Res2Net50~\cite{res2net} and SwinTransformer~\cite{liu2021Swin}. Compared with the model with other stronger backbones or extra training data (UJSC), our LSR+ model with the swinTransformer backbone achieves the best performance. }

\noindent\textbf{Discriminative region localization:}
We introduce the first camouflaged object discriminative region localization task. Considering the same ground truth acquisition process of our task and the widely studied eye fixation prediction task \cite{jiang2015salicon}
(where both ground truth maps are obtained with eye trackers), we re-train existing eye fixation prediction models (GazeGAN \cite{che2019gaze}, SalGAN \cite{pan2017salgan}, UAVDVSM \cite{he2019understanding} and SimpleNet \cite{reddy2020tidying}) with our camouflaged object localization training dataset and construct the first camouflaged object localization benchmark models in Table
\ref{tab:fixation_baseline_cod10k}. The better performance of our COL model (\enquote{LSR+}) compared with the benchmark models validates the superiority of our solution.
We also visualize the produced localization maps of our method (\enquote{Ours}) and the benchmark methods in Fig.~\ref{fig:col_pred_visualization}, which clearly shows more accurate predictions of our model in localizing the discriminative region of the camouflaged instance.
\noindent\textbf{Camouflaged object ranking:}
We show the ranking predictions of our model and the constructed benchmark models in Table~\ref{tab:ranking_comparison} and Fig.~\ref{fig:cor_pred_visualization}. To adapt the COR task, the class labels are replaced with the camouflage ranks in Mask-RCNN \cite{he2017mask}, MS-RCNN \cite{liu2018path} and SOLOv2 \cite{wang2020SOLOv2} and the saliency ranks in PPA \cite{fang2021salient} are changed into camouflage ranks. As shown in Table~\ref{tab:ranking_comparison}, the proposed algorithm, denoted as \enquote{LSR+},
outperforms other algorithms in terms of $r_{MAE}$ and $Corr$. The framework in PPA \cite{fang2021salient} contains two branches, one for salient object detection and another for saliency ranking. Therefore, it achieves the best segmentation performance in terms of $MAE$. However, the saliency ranking is primarily concerned with the relative saliency of objects in the same image, while the COR models focus on the camouflage degree of the instance throughout the whole dataset. As a result, PPA performs poorly in camouflage ranking metrics $r_{MAE}$ and $Corr$. Although the proposed model is based on Mask-RCNN, directly applying it for the COR task (\enquote{Baseline (MaskRCNN)}) leads to inferior performance compared with our methods.

\begin{table*}[t!]
  \centering
  \scriptsize
  \renewcommand{\arraystretch}{1.1}
  \renewcommand{\tabcolsep}{2.2mm}
  \caption{Ablation study on camouflaged object detection (COD).
  }
  \begin{tabular}{l|cccc|cccc|cccc|cccc}
  \hline
  &\multicolumn{4}{c|}{CAMO \cite{le2019anabranch}}&\multicolumn{4}{c|}{CHAMELEON \cite{Chameleon2018}}&\multicolumn{4}{c|}{COD10K \cite{fan2020camouflaged}}&\multicolumn{4}{c}{NC4K \cite{yunqiu_cod21}} \\
    Method & $S_{\alpha}\uparrow$&$F_{\beta}\uparrow$&$E_{\xi}\uparrow$&$\mathcal{M}\downarrow$& $S_{\alpha}\uparrow$&$F_{\beta}\uparrow$&$E_{\xi}\uparrow$&$\mathcal{M}\downarrow$& 
    $S_{\alpha}\uparrow$&$F_{\beta}\uparrow$&$E_{\xi}\uparrow$&$\mathcal{M}\downarrow$&
   $S_{\alpha}\uparrow$&$F_{\beta}\uparrow$&$E_{\xi}\uparrow$&$\mathcal{M}\downarrow$  \\
  \hline
  \multicolumn{17}{c}{Module Analysis} \\ \hline
    Base& .789 & .751 & .840 & .079 & .878 & .828 & .929 & .034 & .805 & .711 & .880 & .037 & \textbf{.840} & .801 & .896 & .048 \\
   JCOD & .790 & .753 & .850 & .078 & \textbf{.883}& .830 & .927 & .034 & .805 & .713 & .882 & \textbf{.036} & .837 & .803 & .897 & \textbf{.047}  \\
  TCOD & .788 & .759 & .848 & .078 & .876 & .825 & .931 & .035 & \textbf{.806} & \textbf{.721} & \textbf{.889} &\textbf{.036} & .838 & \textbf{.806}& .898 & .048  \\
  \textbf{PTCOD} & \textbf{.792} & \textbf{.763} & \textbf{.850} & \textbf{.076} & .881 & \textbf{.833} & \textbf{.935} & \textbf{.032} & .805 & .719 & .888 & \textbf{.036} & .839 & \textbf{.806} & \textbf{.899} & \textbf{.047}  \\
  \hline
  \multicolumn{17}{c}{Backbone/Resolution Analysis} \\ \hline
  
 STCOD & .857 & .839 & .927 & .050 & .893 & .850 & .945 & .025 & .846 & .777 & .924 & .027 & .874 & .846 & .928 & .036  \\
  UCOD & .799 & .772 & .857 & .078 & .886 & .843 & .938 & .032 & .818 & .740 & .896 & .035 & .841 & .811 & .898 & .049   \\
  DCOD & .769 & .726 & .825 & .081 & .867 & .811 & .930 & .036 & .777 & .672 & .866 & .042 & .821 & .779 & .887 & .052  \\
   \hline
  \end{tabular}
  \label{tab:ablation_study_cod}
\end{table*}

\subsection{Model Analysis}
\label{subsub:model_analysis}
We integrate three different tasks in our framework to achieve simultaneous discriminative region localization, camouflaged object detection and camouflaged object ranking. In this section, we extensively analyse the proposed triple-task learning framework
to explain its effectiveness. We show performance of the COD, COL, and COR related models in Table \ref{tab:ablation_study_cod}, Table \ref{tab:ablation_study_col} and Table \ref{tab:ablation_study_cor} respectively. For easier reference, we show the base models of training each task separately (as discussed in Section \ref{subsec:base_models}) as \enquote{Base} (corresponding to \enquote{LSR+} in Table \ref{tab:benchmark_model_comparison}, \ref{tab:fixation_baseline_cod10k} and \ref{tab:ranking_comparison}) in each corresponding table. 


\noindent\textbf{Residual learning based joint COD \& COL:} As both COD and COL can be achieved with FCNs,
we jointly train the two tasks with the proposed residual learning based joint COD \& COL model in Fig.~\ref{fig:joint_cod_fixation}, where we adopt the dual task loss function in Eq.~\ref{dual_loss} for model parameters updating.
We show performance of $s_l^{\text{init}}$ and $s_b^{\text{init}}$ as \enquote{JCOL} and \enquote{JCOD} in the corresponding tables. We observed improved performance of \enquote{JCOL} and \enquote{JCOD} compared with the base models,
validating effectiveness of the proposed joint COD \& COL learning framework.

\noindent\textbf{Triple-task learning framework:} Based on the reverse attention based joint learning framework, we can add a camouflaged object ranking module (\enquote{COR} in Fig.~\ref{fig:network_overview}) to it and achieve the triple-task learning framework. In this way, we define prediction of COD, COL, and COR as $s_b^{\text{init}}$, $s_l^{\text{init}}$ and $\{s_{\text{init}},s_r\}$, and the corresponding performance is shown as \enquote{TCOD}, \enquote{TCOL}, and \enquote{TCOR} respectively. We observe similar performance of \enquote{TCOD} (\enquote{TCOL}) compared with \enquote{JCOD} (\enquote{JCOL}), explaining that introducing an extra ranking model to the joint learning framework without effective task relationship modeling brings no benefits to either of the two tasks. Further, the similar performance of \enquote{TCOR} and \enquote{Base} in Table \ref{tab:ablation_study_cor} also explains the ineffectiveness of the other two tasks contributing to the COR task.

\noindent\textbf{The proposed triple-task learning model:} Based on the above triple-task learning framework, we introduce task-interaction as shown in Fig.~\ref{fig:network_overview}, leading to the proposed triple-task learning model. We then show performance of COD, COL, and COR as \enquote{\textbf{PTCOD}}, \enquote{\textbf{PTCOL}}, and \enquote{\textbf{PTCOR}} respectively, representing the performance of $s_b^{\text{ref}}$, $s_l^{\text{ref}}$ and $\{s_{\text{ins}},s_r\}$ (see Fig.~\ref{fig:network_overview} and Section \ref{subsubsec:triple_task} for the generation of $s_b^{\text{ref}}$, $s_l^{\text{ref}}$ and $\{s_{\text{ins}},s_r\}$). The consistently better performance of \enquote{\textbf{PTCOD}}, \enquote{\textbf{PTCOL}}, and \enquote{\textbf{PTCOR}} compared with the naive triple task learning models (\enquote{TCOD}, \enquote{TCOL}, and \enquote{TCOR}) and the base models explains the effectiveness of the proposed task-interaction modeling strategies.

\begin{table}[t!]
  \centering
  \scriptsize
  \renewcommand{\arraystretch}{1.2}
  \renewcommand{\tabcolsep}{1.7mm}
  \caption{Ablation study on camouflaged object localization (COL).}
  \begin{tabular}{l|cccccc}
  \hline
  Method&$SIM\uparrow$ & $CC\uparrow$ & $EMD\downarrow$ & $KLD\downarrow$ & $NSS\uparrow$ & $AUC\_J\uparrow$   \\ \hline
  \multicolumn{7}{c}{Module Analysis} \\ \hline
   Base&0.572 & 0.767 & 2.408 & 0.791 & 1.265 & 0.971   \\
   JCOL&0.577 & 0.765 & 2.302 & \textbf{0.766} & \textbf{1.311} & 0.970   \\
    TCOL&\textbf{0.588} & 0.770 & 2.196 & 0.788 & 1.292 & 0.971  \\
     \textbf{PTCOL}&0.587 & \textbf{0.771} & \textbf{2.187} & 0.775 & 1.300 & \textbf{0.972} \\  \hline 
   \multicolumn{7}{c}{Backbone/Resolution Analysis} \\ \hline
   STCOL&0.592 & 0.781 & 2.166 & 0.799 & 1.306 & 0.975    \\  
   UCOL&0.588 & 0.774 & 2.216 & 0.764 & 1.287 & 0.973  \\
   DCOL&0.572 & 0.752 & 2.324 & 0.778 & 1.315 & 0.968  \\ \hline
  \end{tabular}
  \label{tab:ablation_study_col}
\end{table}

\begin{table}[t!]
  \centering
  \scriptsize
  \renewcommand{\arraystretch}{1.2}
  \renewcommand{\tabcolsep}{6mm}
  \caption{Ablation study on camouflaged object ranking (COR).}
  \begin{tabular}{r|cccc}
  \hline
  Method & $MAE\downarrow$ & $r_{MAE}\downarrow$ & $Corr\uparrow$ \\\hline
  \multicolumn{4}{c}{Module Analysis} \\ \hline
  Base & 0.046 & 0.227 & 0.465 \\
  TCOR & \textbf{0.042} & 0.219 & 0.462 \\
  \textbf{PTCOR} & \textbf{0.042} & \textbf{0.217} & \textbf{0.467} \\ \hline
  \multicolumn{4}{c}{Backbone/Resolution Analysis} \\ \hline
  STCOR & 0.043 & 0.223 & 0.482 \\
  UCOR & 0.042 & 0.216 & 0.452\\
  DCOR & 0.047 & 0.226 & 0.472 \\
   \hline
  \end{tabular}
  \label{tab:ablation_study_cor}
\end{table}

\noindent\textbf{Triple-task learning with different backbones:} Given the proposed triple-task learning framework in Fig.~\ref{fig:network_overview}, in this paper, we adopt the ResNet50 backbone as it is widely used in existing models. We further analyse model performance with different backbones. Specifically, we keep our overall framework in Fig.~\ref{fig:network_overview} unchanged, and use the Swin transformer \cite{liu2021Swin} backbone for the backbone of the joint COD \& COL module.
The performance of the three tasks is then shown as
\enquote{STCOD}, \enquote{STCOL}, and \enquote{STCOR} respectively.
The significant better performance of both the COD and COL task with the Swin transformer \cite{liu2021Swin} backbone compared with using ResNet50 backbone (\enquote{PTCOD} and \enquote{PTCOL}) validates the superiority of transformer backbone for COD and COL. Further, we observe slightly influenced performance of the COR task (\enquote{PTCOR}). Although the backbone of the COR task is still ResNet50, as a triple task learning framework, its performance is still improved with the improved performance of the other two tasks.

\noindent\textbf{Triple-task learning with different resolutions:} We find the resolution of training/testing images is critical for camouflaged object detection. To test how image resolution affects performance of each task, given the trained triple-task learning framework with ResNet50 backbone (the \enquote{PT*} models), we test model performance with both higher resolution ($448\times448$) and lower resolution ($256\times256$) than the training images ($352\times352$), and show model performance as \enquote{UCOD}, \enquote{UCOL}, and \enquote{UCOR} for the higher resolution results and \enquote{DCOD}, \enquote{DCOL}, and \enquote{DCOR} for the lower resolution results. Note that for the \enquote{PT*} models, we have the same resolution ($352\times352$) for the training dataset and testing dataset. Table \ref{tab:ablation_study_cod} shows that with larger testing image size (\enquote{UCOD}), the trained COD model can yield better performance, and the lower resolution testing images (\enquote{DCOD}) will significantly decrease model performance. However, for the COL task, although the lower resolution testing images can influence COL performance, its performance is still comparable with testing with larger resolution images. For the COR task, the higher resolution testing images (\enquote{UCOR}) lead to comparable performance to the raw model (\enquote{PTCOR}). However, the lower resolution images significantly decrease its performance. Considering the influence of upsampling and downsampling operations on the three tasks, we conclude that resolution is especially important for camouflage object detection, localization and ranking. The training dataset with one resolution leads to one specific camouflage distribution description. The model trained with this dataset may be incapable of operating on
lower resolution images following a different camouflage distribution. We will further investigate camouflage distribution with respect to testing image resolutions in the future.

\begin{figure}[!htp]
  \begin{center}
  \begin{tabular}{c@{ } c@{ } c@{ } c@{ } c@{ } c@{ } c@{ }}
    {\includegraphics[height=0.10\linewidth]{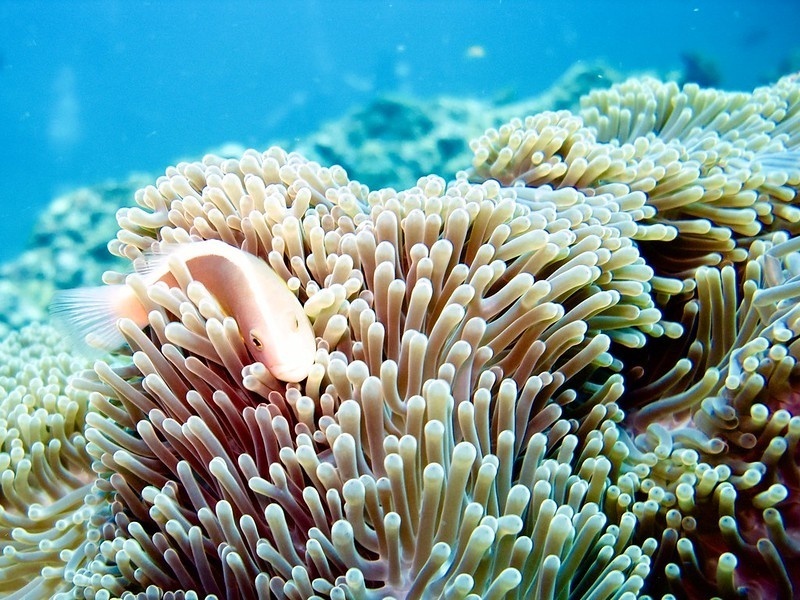}}&
    {\includegraphics[height=0.10\linewidth]{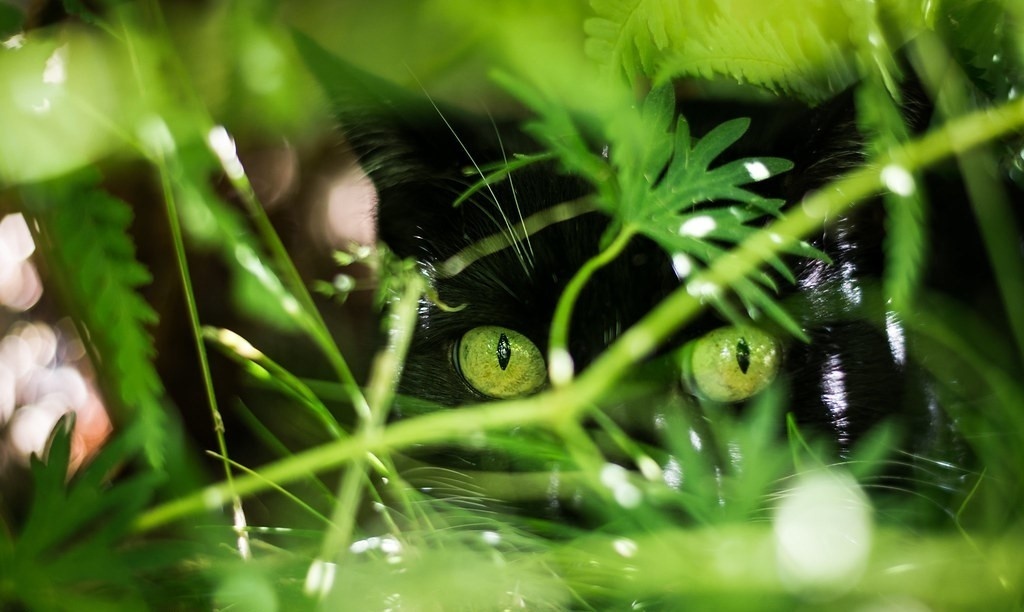}}&
    {\includegraphics[height=0.10\linewidth]{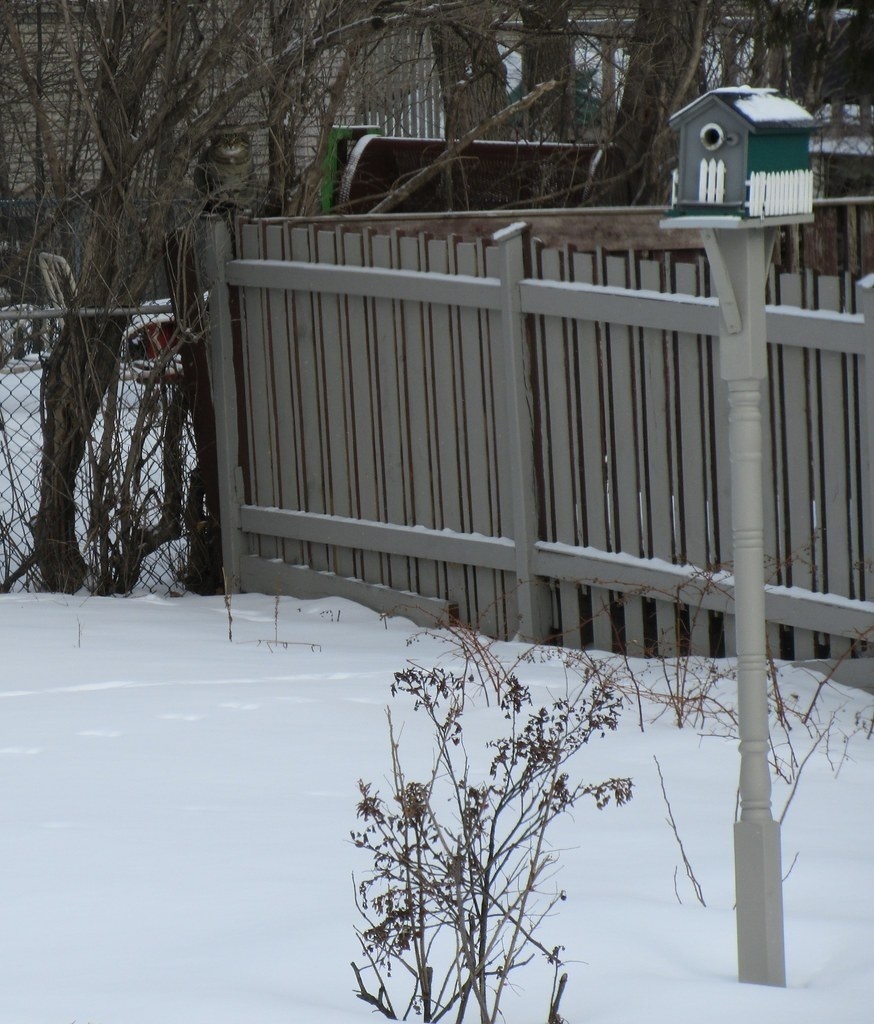}}&
    {\includegraphics[height=0.10\linewidth]{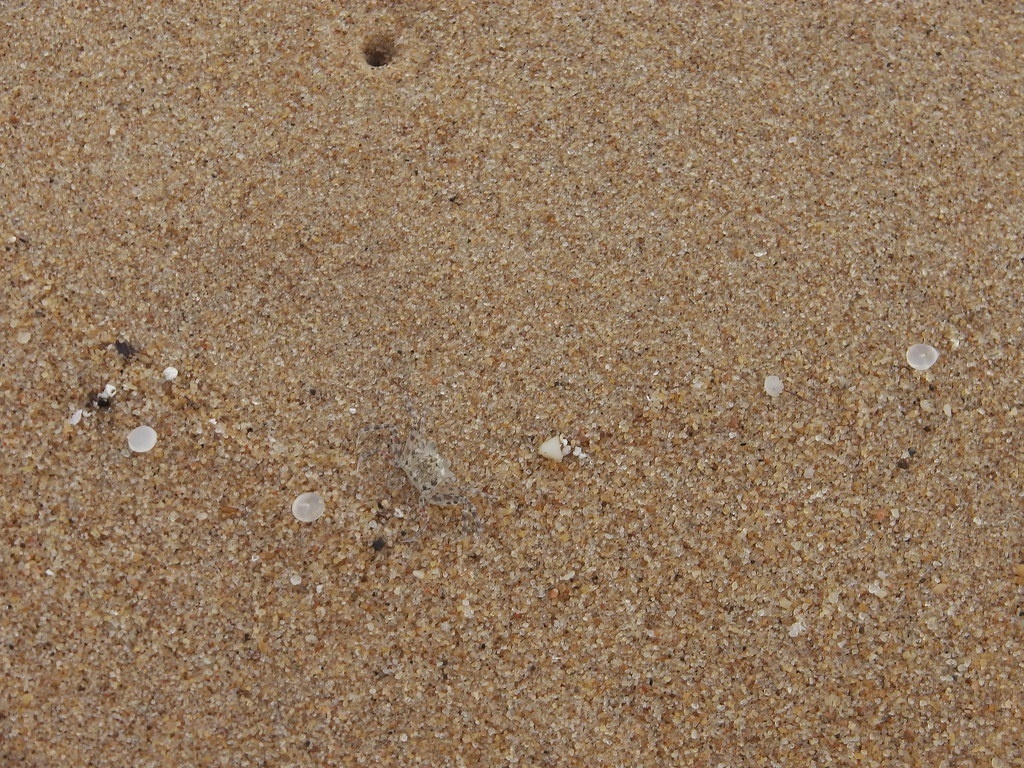}}&
    {\includegraphics[height=0.10\linewidth]{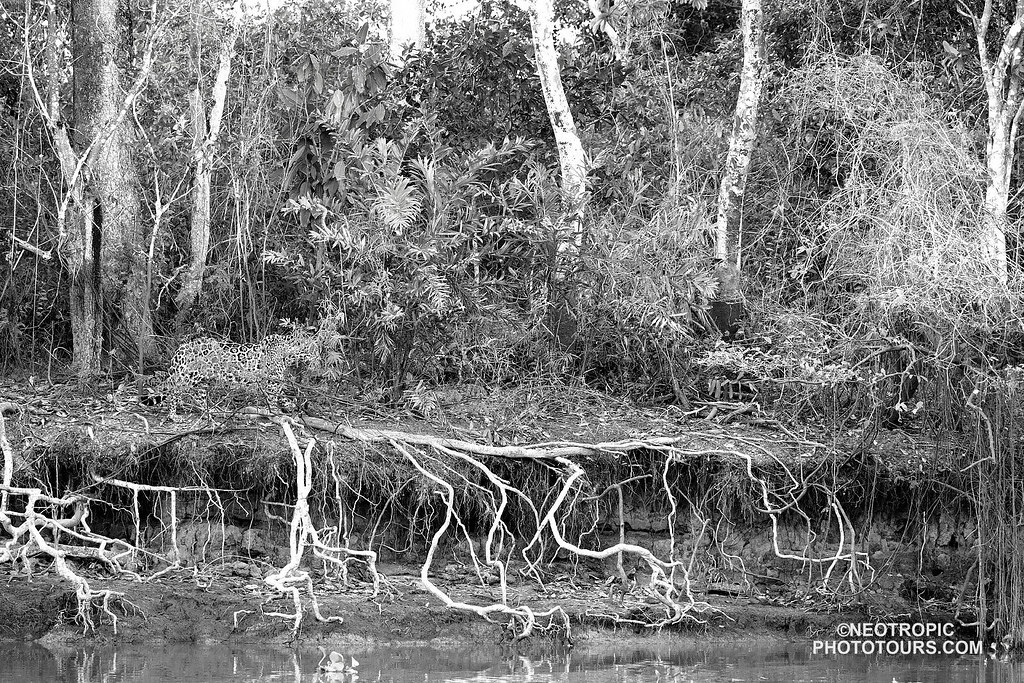}}&
    {\includegraphics[height=0.10\linewidth]{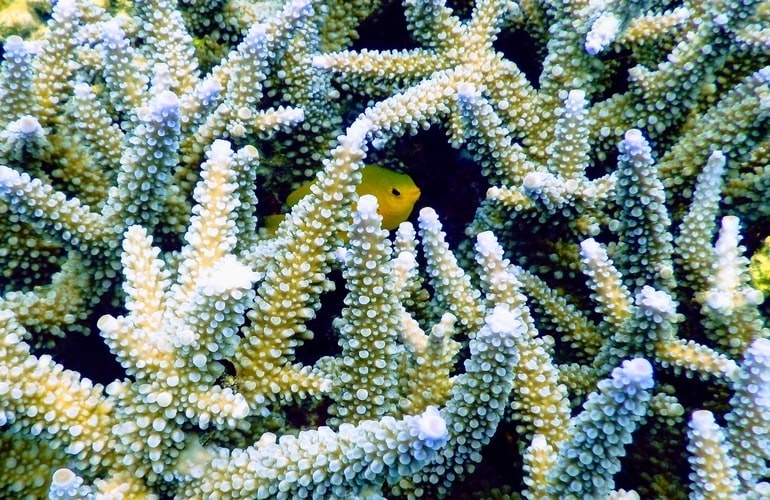}}&
    {\includegraphics[height=0.10\linewidth]{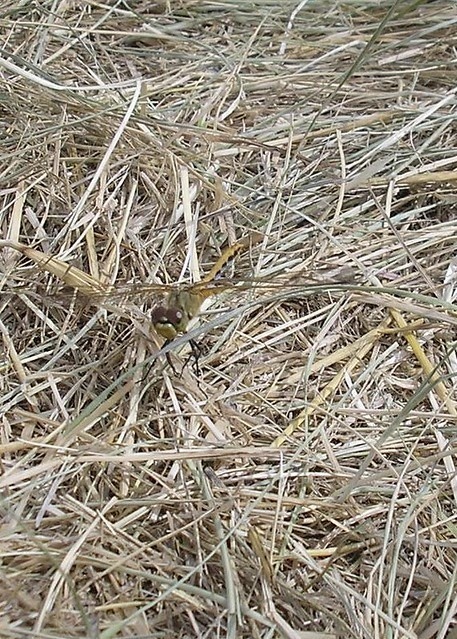}}
    \\
    {\includegraphics[height=0.10\linewidth]{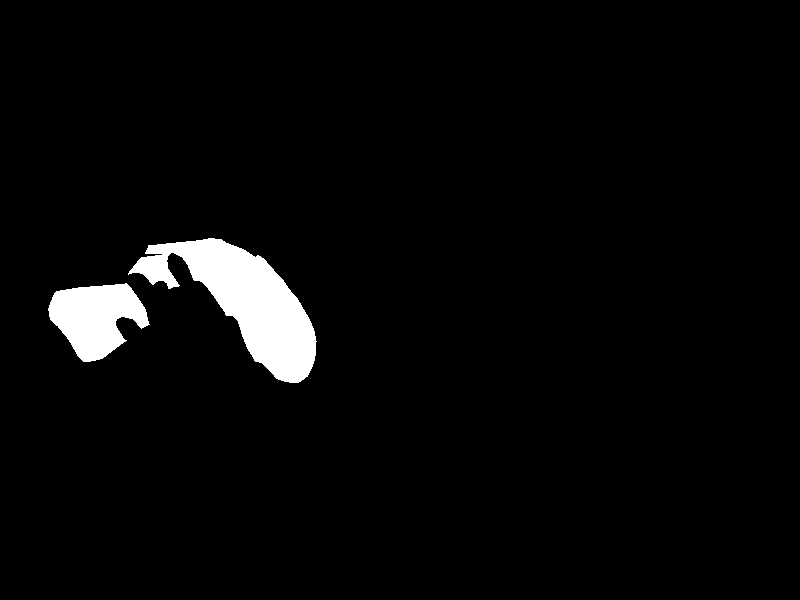}}&
    {\includegraphics[height=0.10\linewidth]{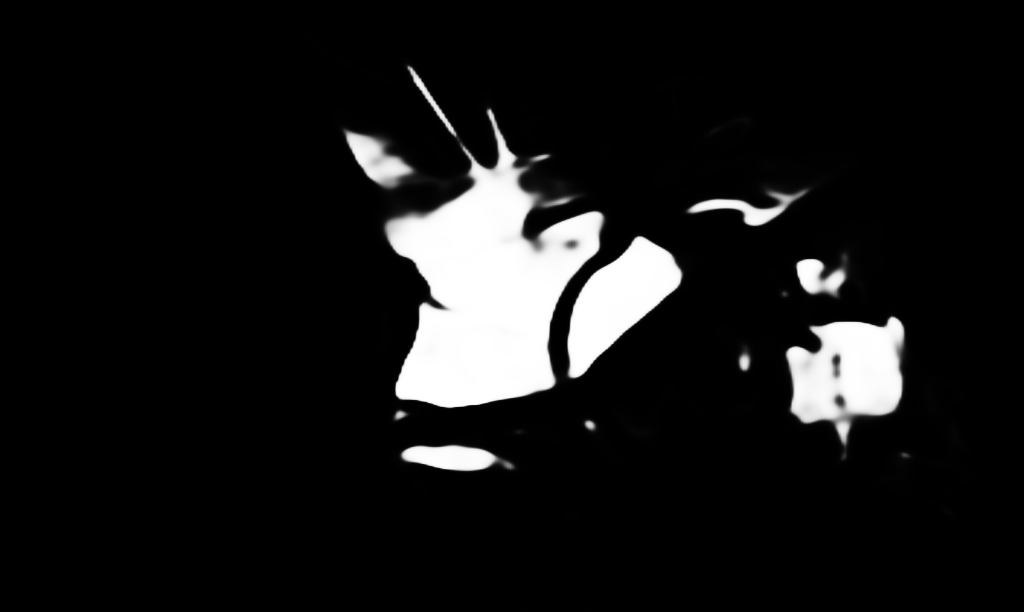}}&
    {\includegraphics[height=0.10\linewidth]{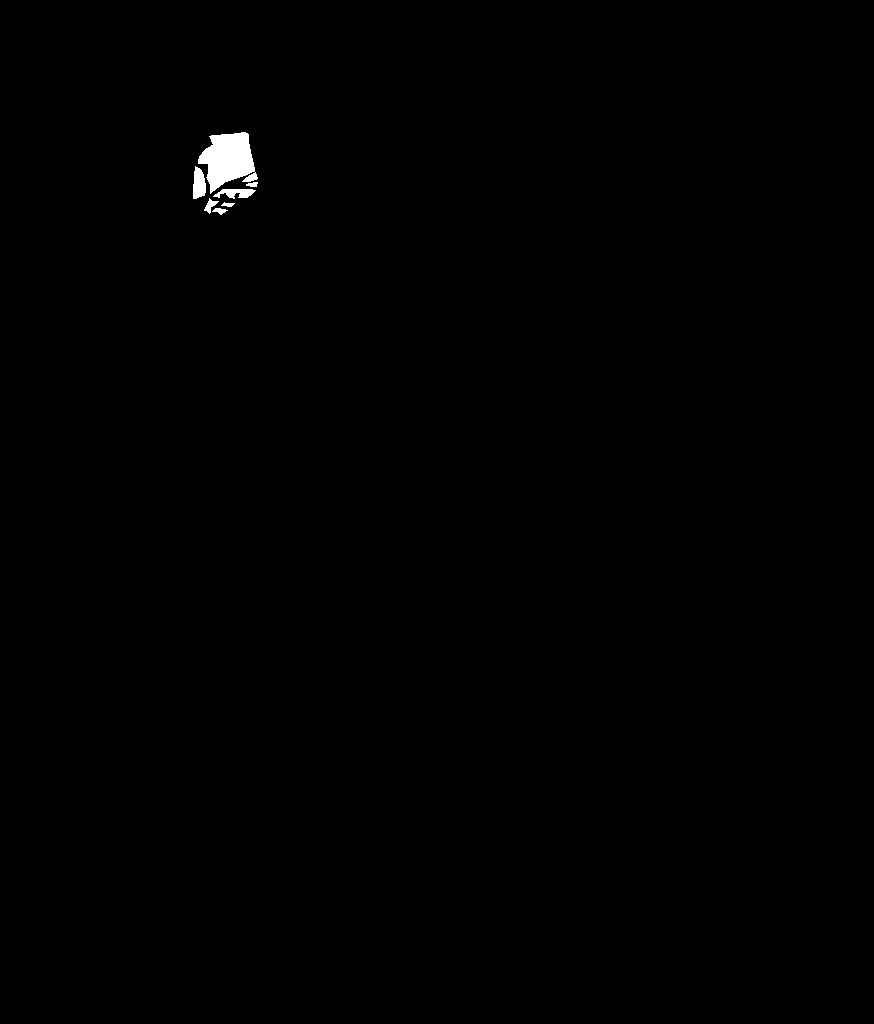}}&    
    {\includegraphics[height=0.10\linewidth]{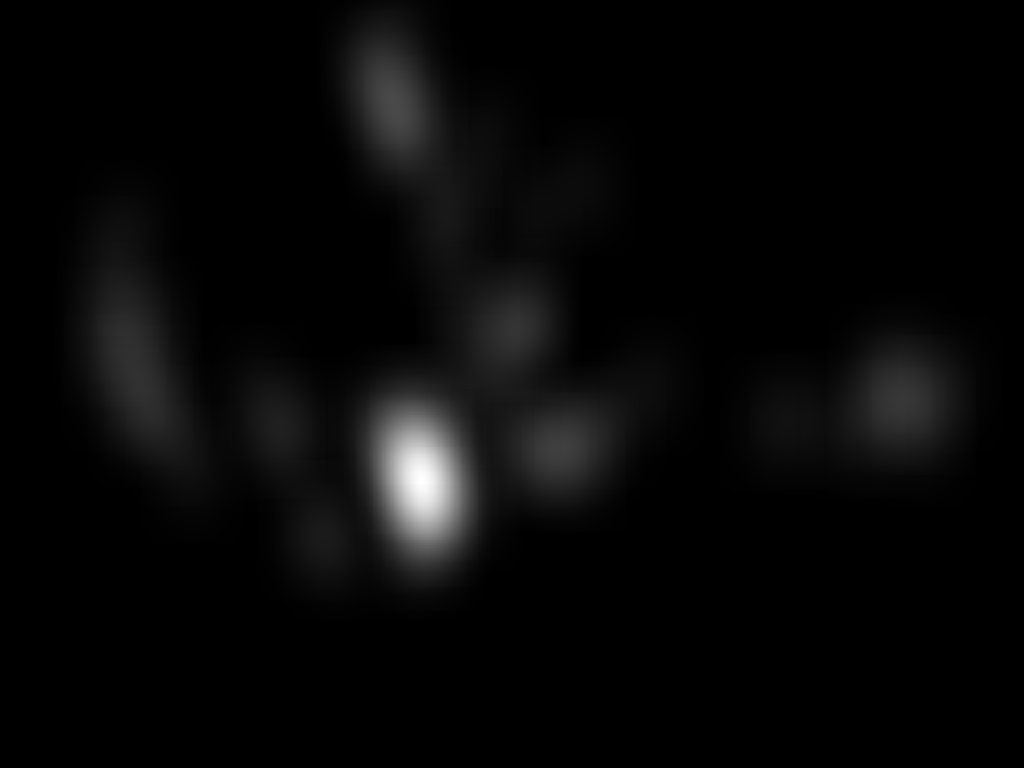}}&
    {\includegraphics[height=0.10\linewidth]{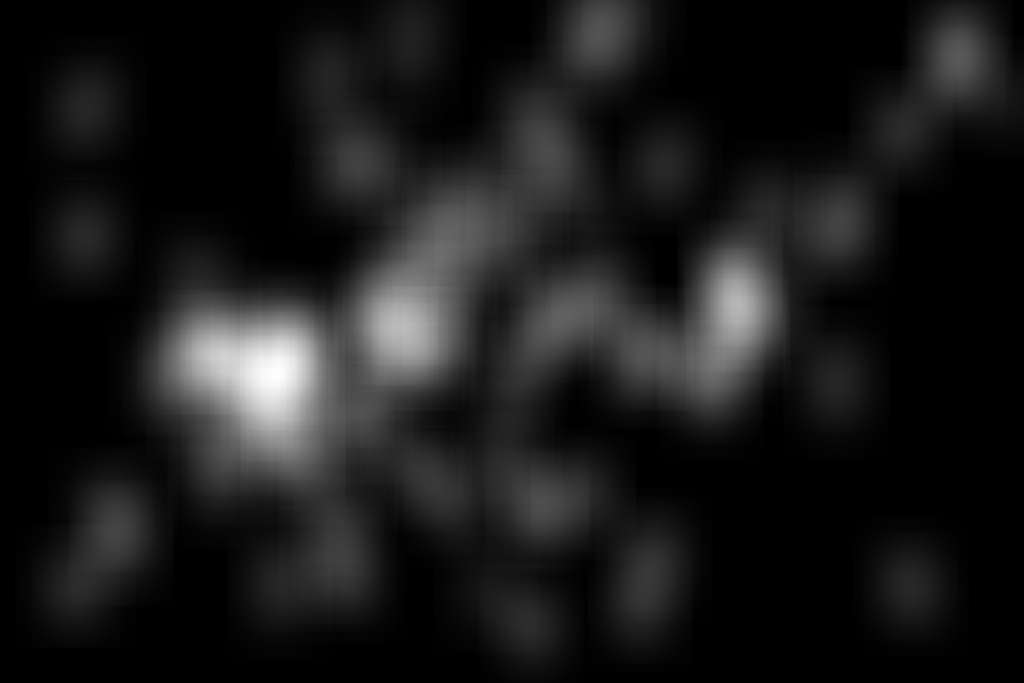}}&
    {\includegraphics[height=0.10\linewidth]{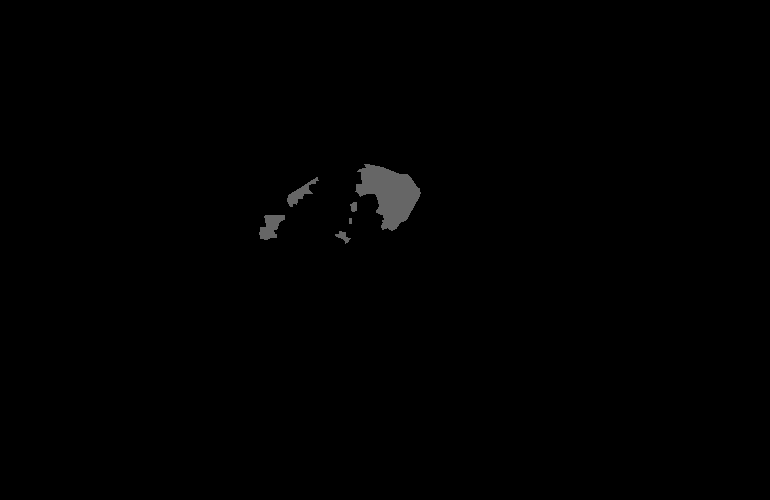}}&
    {\includegraphics[height=0.10\linewidth]{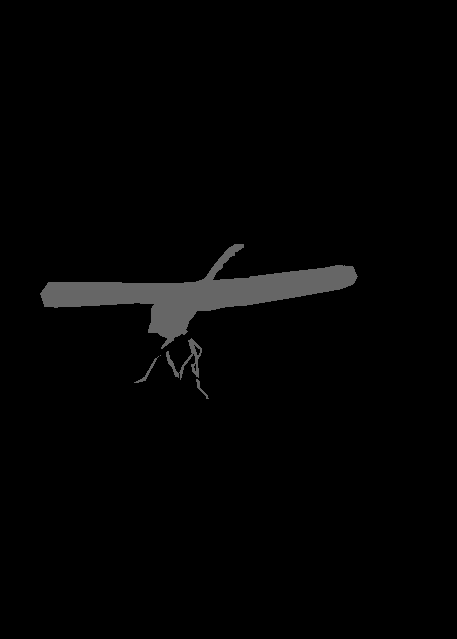}}\\
    {\includegraphics[height=0.10\linewidth]{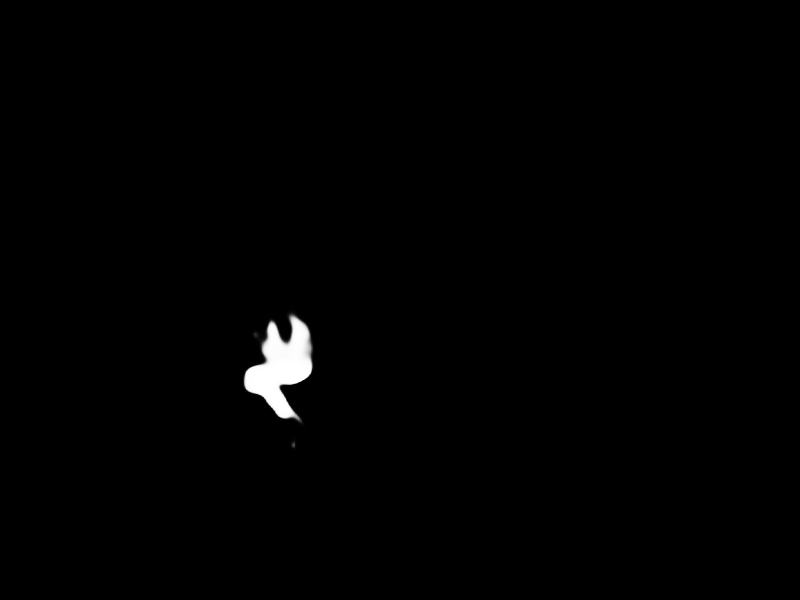}}&
    {\includegraphics[height=0.10\linewidth]{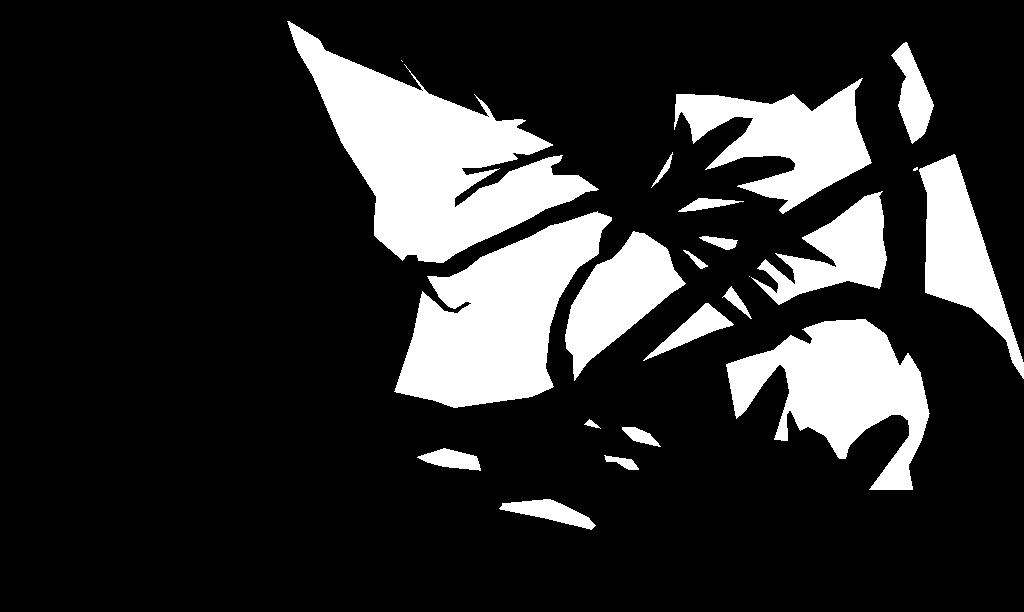}}&
    {\includegraphics[height=0.10\linewidth]{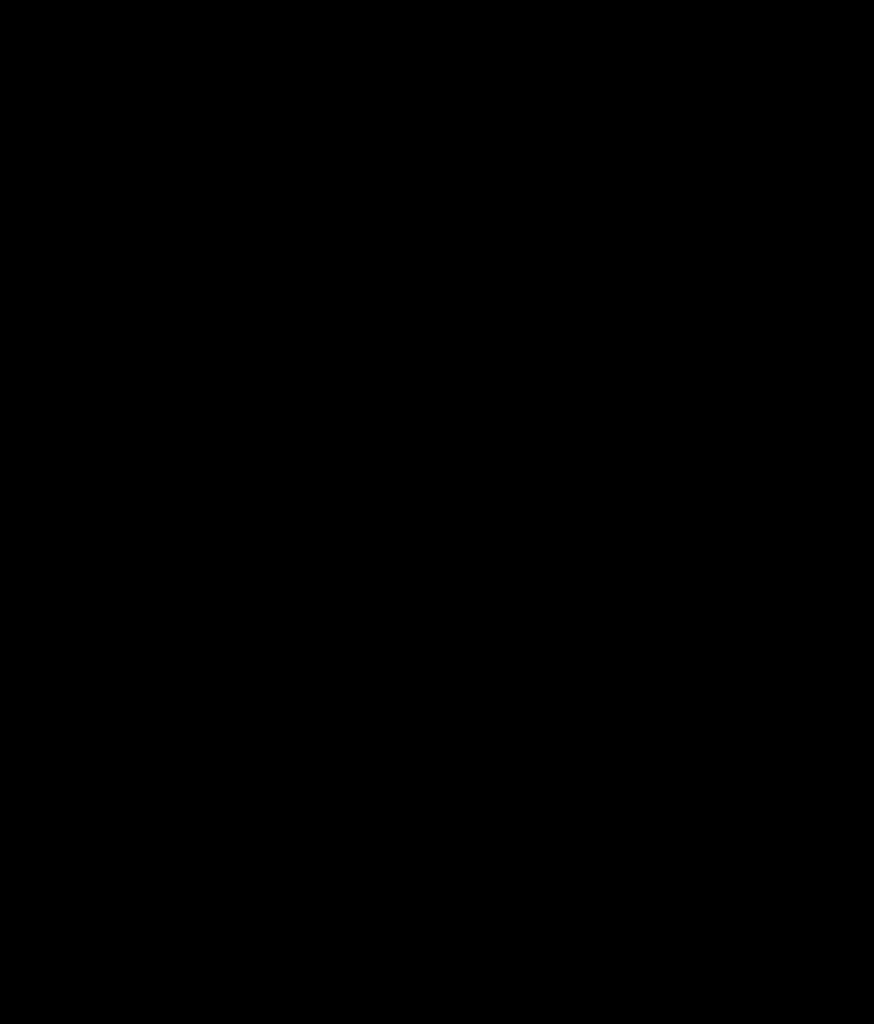}}&
    {\includegraphics[height=0.10\linewidth]{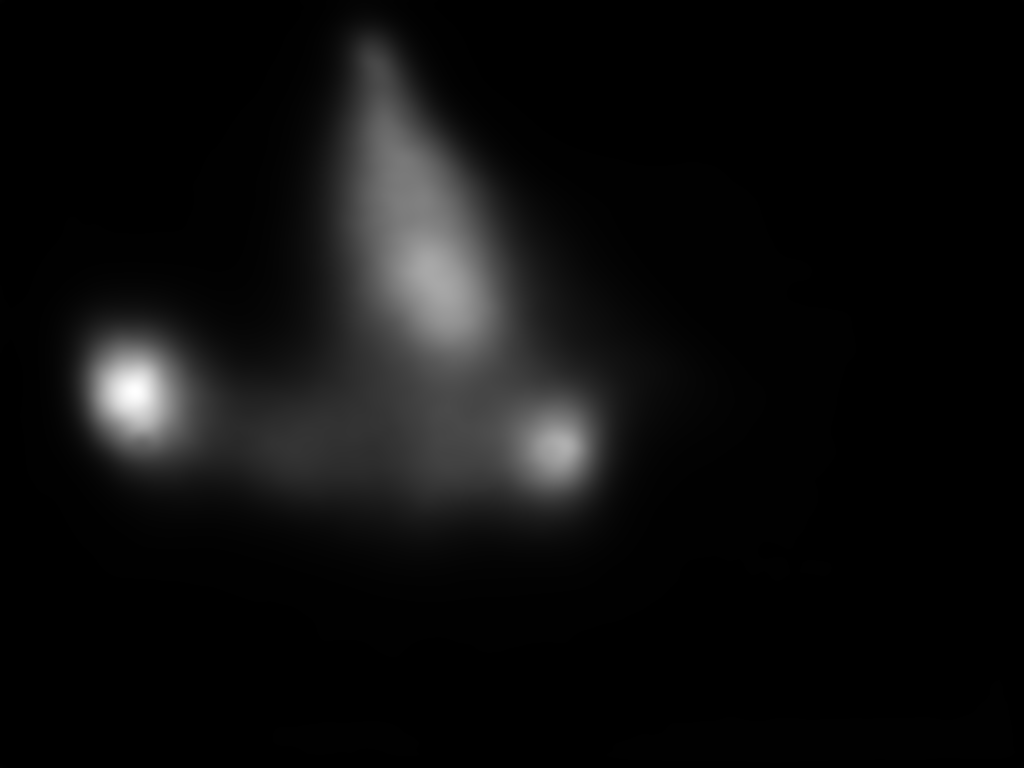}}&
    {\includegraphics[height=0.10\linewidth]{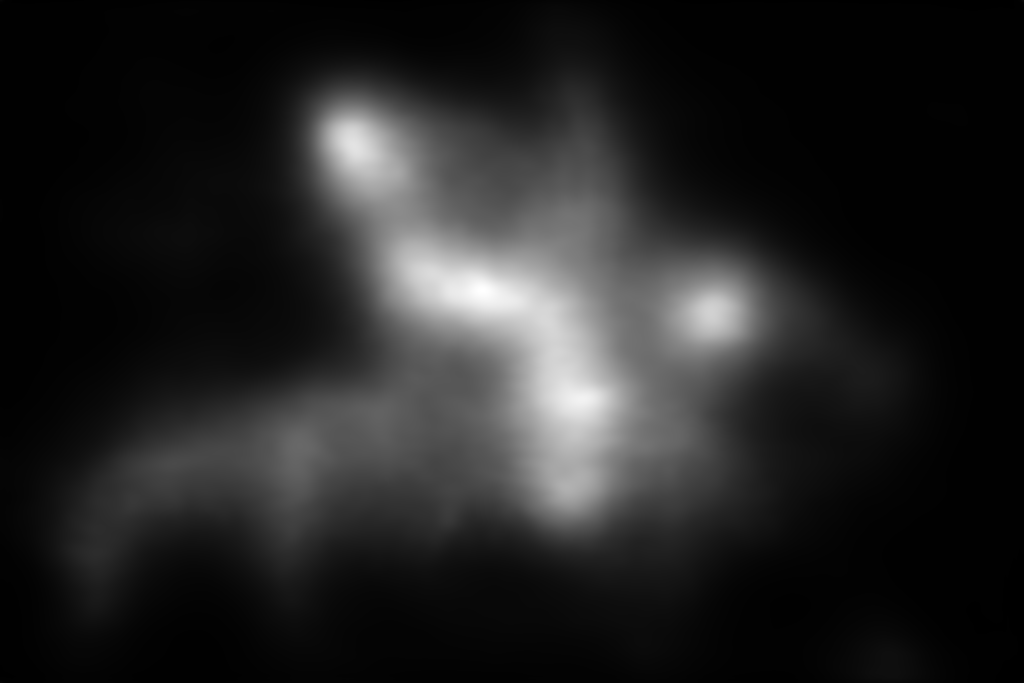}}&
    {\includegraphics[height=0.10\linewidth]{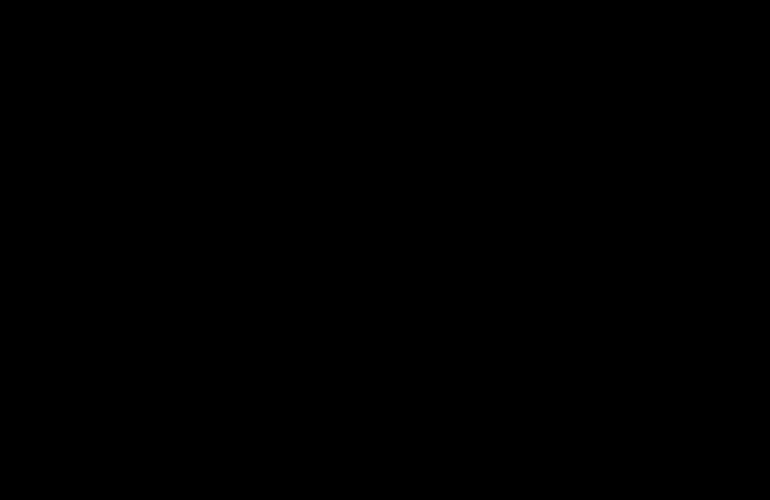}}&
    {\includegraphics[height=0.10\linewidth]{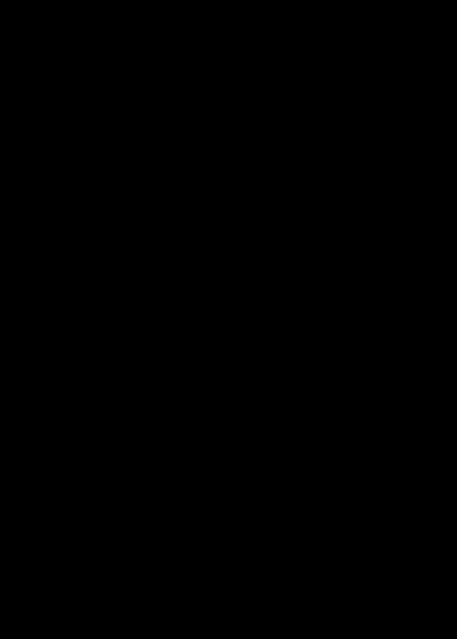}}
    \\
    \footnotesize{(a)}& \footnotesize{(b)} &\footnotesize{(c)}& \footnotesize{(d)}&\footnotesize{(e)} &\footnotesize{(f)}& \footnotesize{(g)}\\
  \end{tabular}
    \end{center}
  \caption{\Rev{Failure cases of our proposed method. From top to bottom: the original image, ground truth, prediction of our method. (a)-(c),(d)-(e) and (f)-(g) are failure cases of COD, COL and COR, respectively.}
  }
\label{fig:failure_cases}
\end{figure}

\noindent \Rev{ {\bf Failure cases:} The failure cases of our triple-task learning camouflaged object detection model PTCOD are shown in Fig.~\ref{fig:failure_cases}. As we can see from Fig.~\ref{fig:failure_cases}(a), we find that our model always fails when the object is extremely small. Another failure case is that the dataset contains similar foreground objects with the background. As in Fig.~\ref{fig:failure_cases}(b), the disruptive object in the background is similar with ``pipefish'' in the COD10K training dataset. Fig.~\ref{fig:failure_cases}(c) shows that the model sometimes cannot segment some occluded objects successfully. Fig.~\ref{fig:failure_cases}(d)-(e) illustrate that the hard instances also pose a great challenge for COL model. From Fig.~\ref{fig:failure_cases}(f)-(g), we can draw the conclusion that the detection is also important to ranking. }

\subsection{Task Exploration}
\subsubsection{Camouflaged Object Localization}
COL aims to capture the salient regions of the camouflaged object to generate a localization map, indicating the discriminative regions of the camouflaged object.
The ground truth map is generated with an eye tracker to record the raw fixation points, which are then Gaussian blurred to obtain our final localization map. Given the similar ground truth acquisition process, we find the existing fixation prediction models \cite{jiang2015salicon} can be used to train our COL task.
Most existing eye fixation prediction models are trained based on a fully convolutional neural network to achieve mapping from input RGB image to the corresponding blur version fixation map \cite{he2019understanding,che2019gaze,gan_raw}.

We find two main directions to further explore our COL model, including using more appropriate loss functions and changing the deterministic model to a stochastic model considering its inherent randomness attribute. In this paper, we adopt L2 loss as the loss function for the COL module, which is the most natural choice for a regression model. We find some eye fixation prediction techniques using the metric related loss functions, including KL divergence, NSS, SIM, and \etc. To test how our COL model performs with different loss functions, we train our base COL model (\enquote{Base} in Table \ref{tab:ablation_study_col}) with various types of loss functions, including the KL divergence loss (\enquote{KL loss}), the cross-entropy loss (\enquote{CE loss}), the correlation coefficient loss (\enquote{CC loss}), the similarity metric related loss (\enquote{SIM loss}). We then show performance in Table \ref{tab:col_wrt_loss_function_stomodel}. We observe relative stable performance of those models with different loss functions
except for the KLD measure. Further research can be conducted on choosing more suitable loss functions for our COL task.

\begin{figure}[!htp]
  \begin{center}
  \begin{tabular}{c@{ } c@{ } c@{ } c@{ } c@{ } c@{ }}
      {\includegraphics[width=0.149\linewidth]{fix_show/Img_COD10K-CAM-1-Aquatic-2-ClownFish-10.jpg}}&
    {\includegraphics[width=0.149\linewidth]{fix_show/GT_COD10K-CAM-1-Aquatic-2-ClownFish-10.png}}&
    {\includegraphics[width=0.149\linewidth]{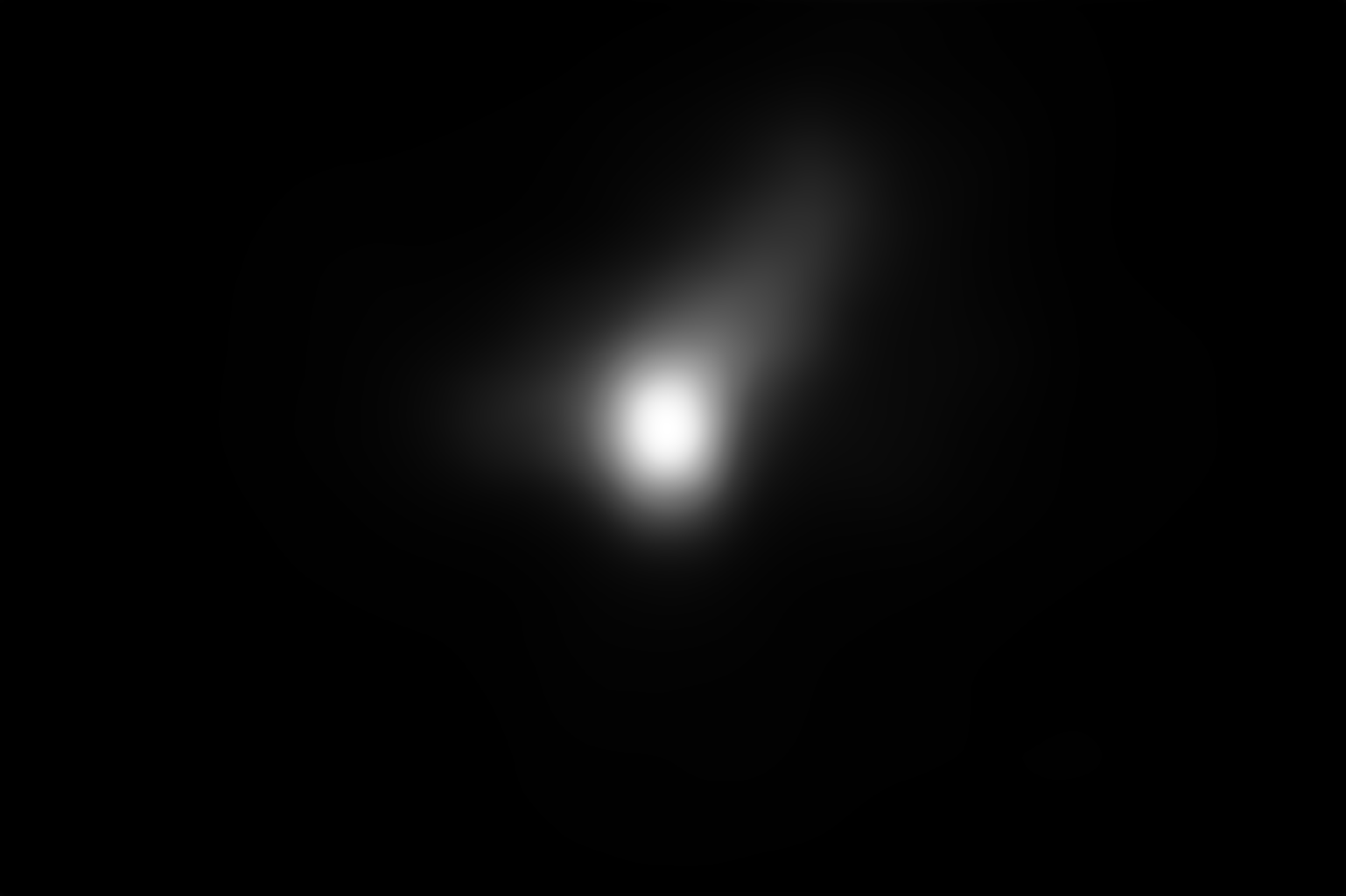}}&
    {\includegraphics[width=0.149\linewidth]{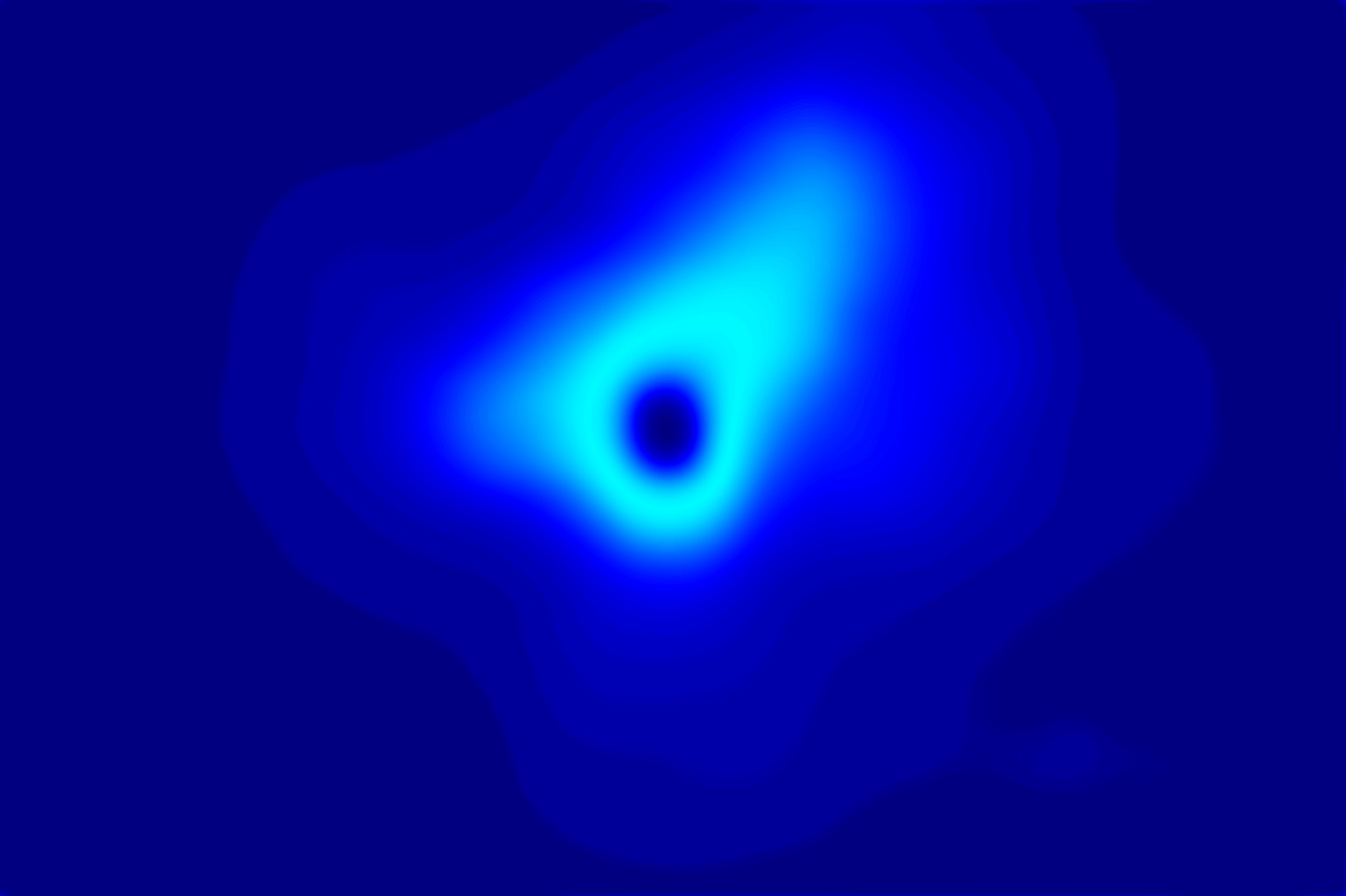}}&
    {\includegraphics[width=0.149\linewidth]{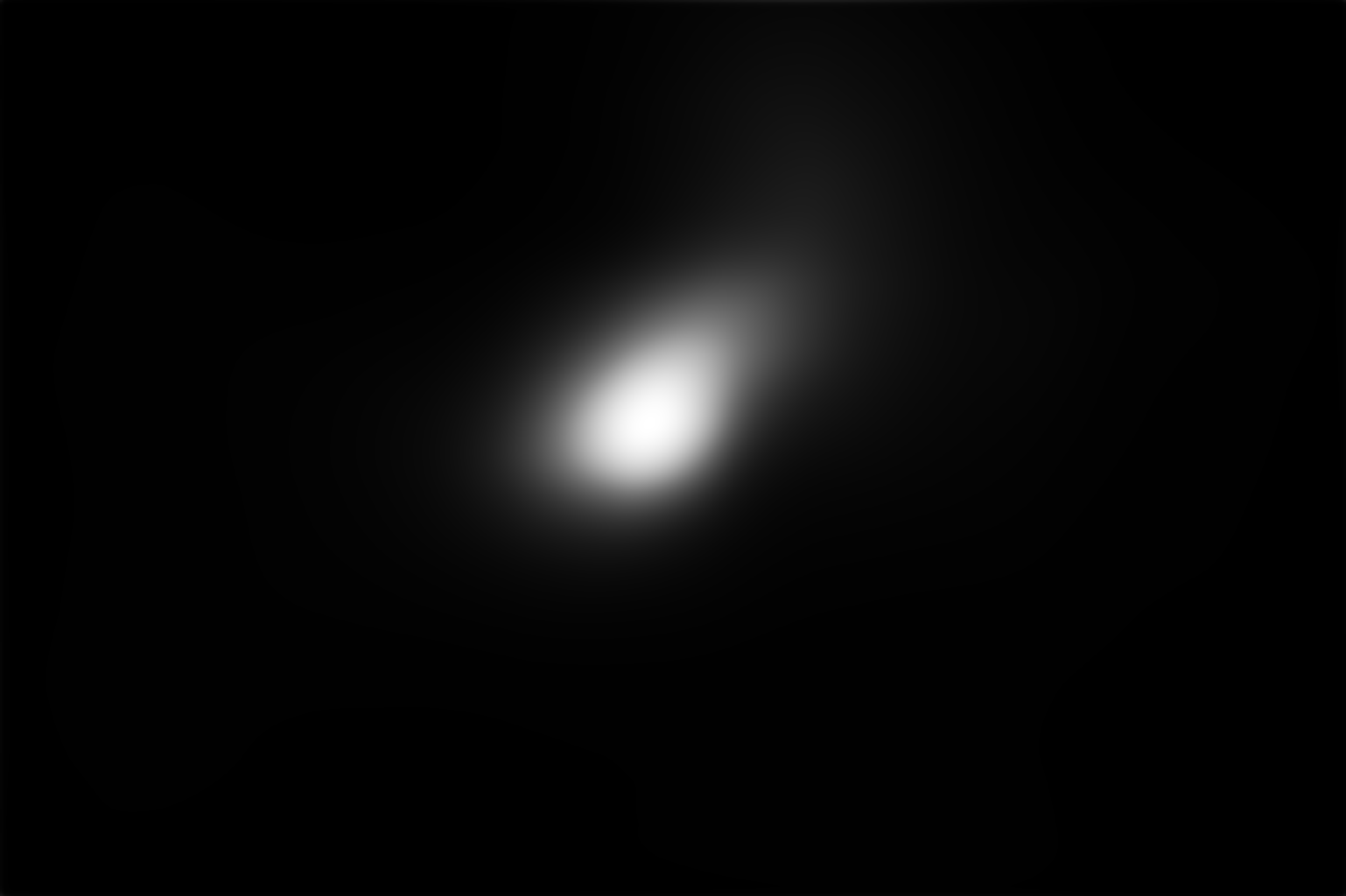}}&
    {\includegraphics[width=0.149\linewidth]{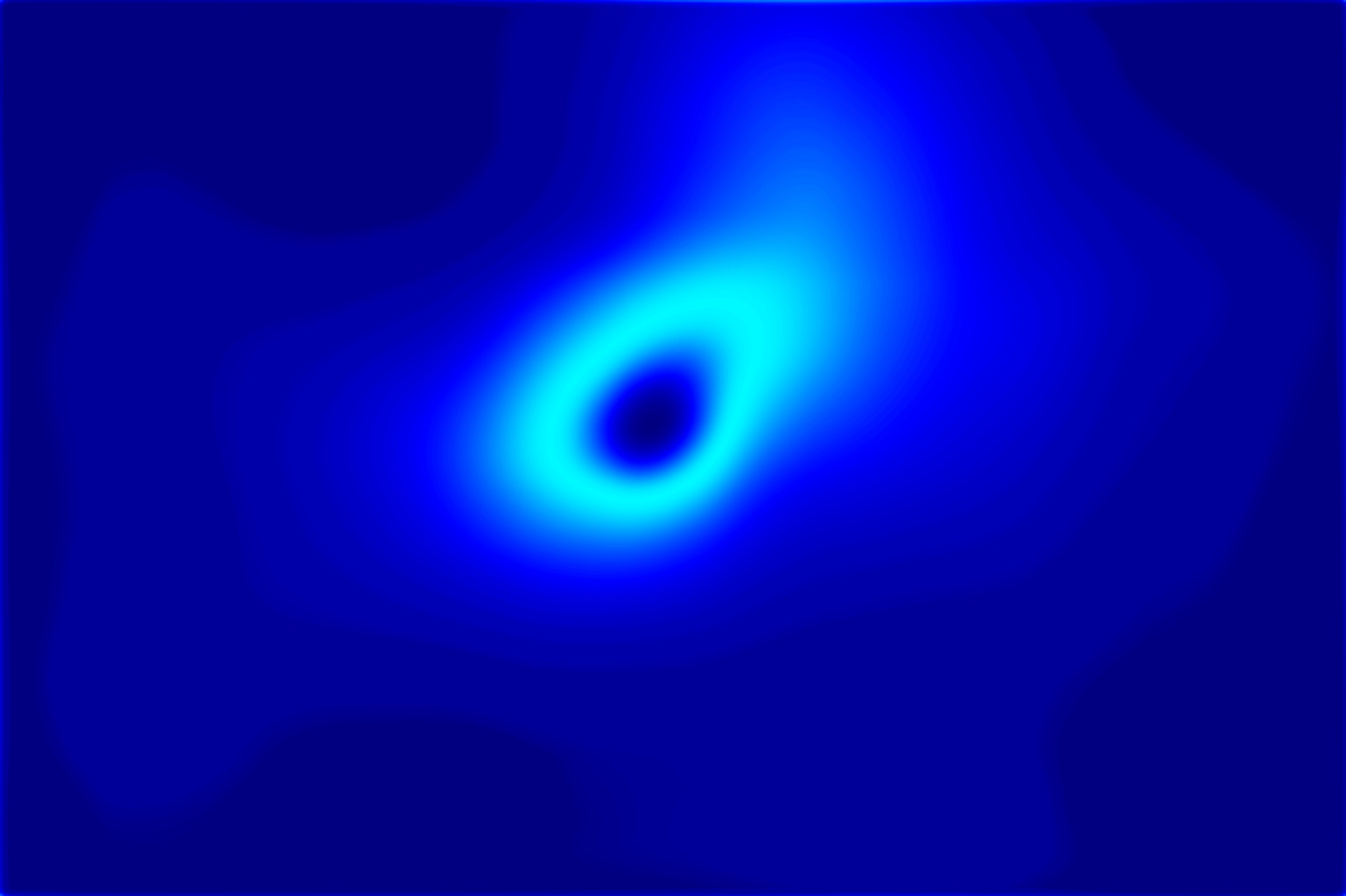}}\\
    {\includegraphics[width=0.149\linewidth]{fix_show/Img_COD10K-CAM-2-Terrestrial-23-Cat-1492.jpg}}&
    {\includegraphics[width=0.149\linewidth]{fix_show/GT_COD10K-CAM-2-Terrestrial-23-Cat-1492.png}}&
    {\includegraphics[width=0.149\linewidth]{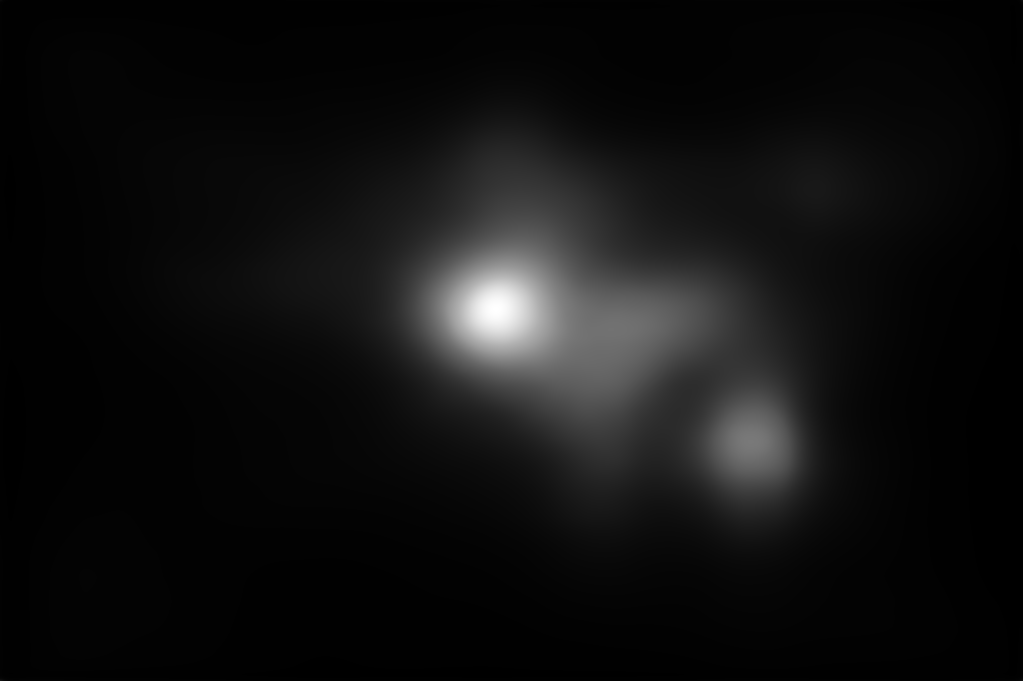}}&
    {\includegraphics[width=0.149\linewidth]{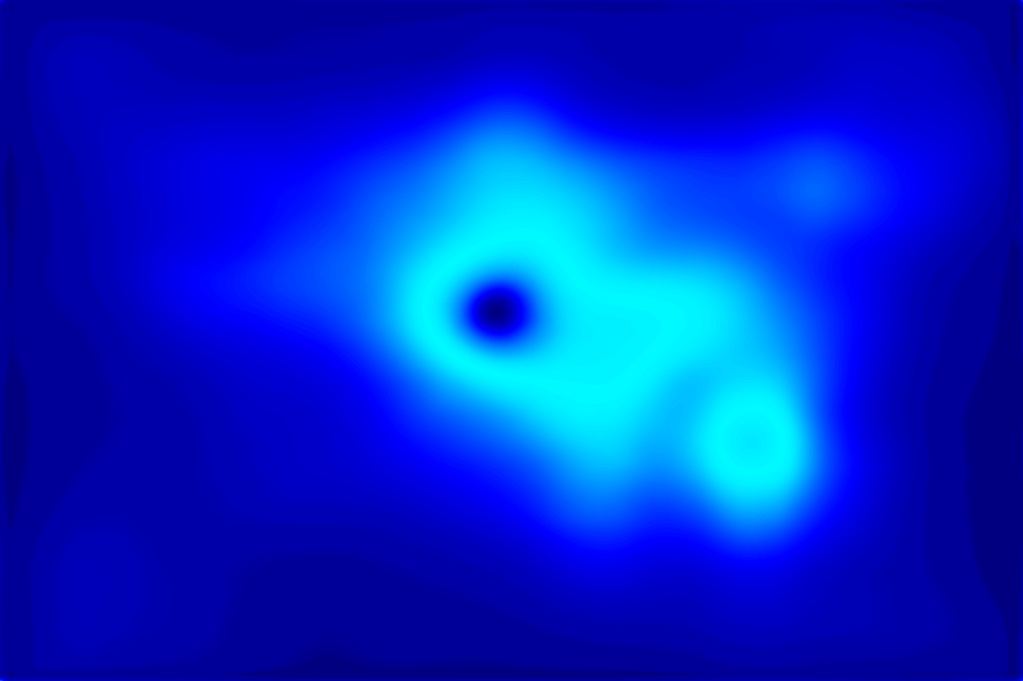}}&
    {\includegraphics[width=0.149\linewidth]{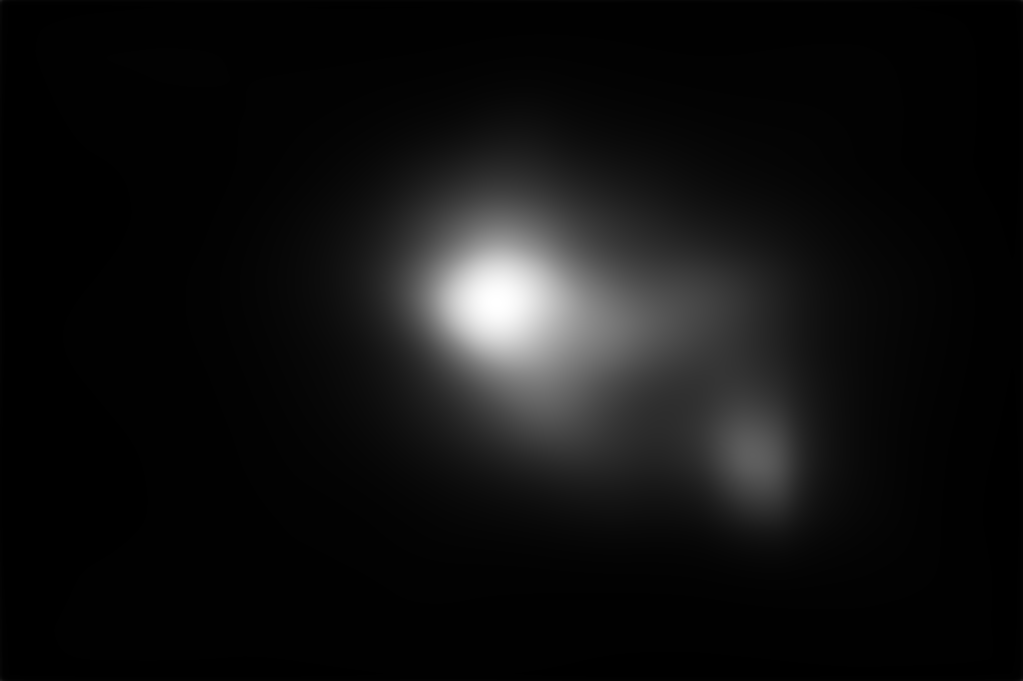}}&
    {\includegraphics[width=0.149\linewidth]{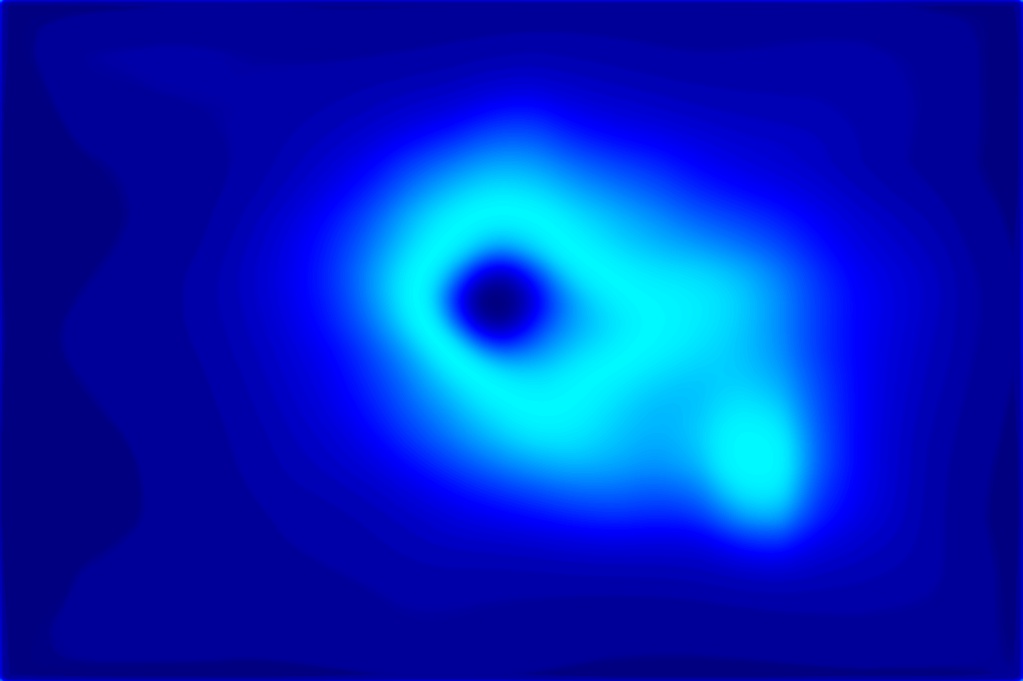}}\\
    \footnotesize{Image}& \footnotesize{GT}&\multicolumn{2}{c}{\footnotesize{GAN}}&\multicolumn{2}{c}{\footnotesize{VAE}}\\
  \end{tabular}
  \end{center}
  \caption{Visualization of the uncertainty area of the extended generative model based COL task, where the two images within each generative model block are the deterministic prediction and the uncertainty map.
  }
\label{fig:uncertainty_col}
\end{figure}

Further, considering the inherent randomness of the ground truth acquisition process of our COL task, we explore our COL task with latent variable models, including the
generative adversarial net (GAN) \cite{gan_raw} and the variational auto-encoder (VAE) \cite{vae_raw},
where the latent variables within the latent variable models aim to represent the \enquote{inherent randomness} of the task. Note that the L2 loss function is used in both models for the generator, and we adopt the GAN based framework as in \cite{li2021uncertainty} and the VAE based framework as in \cite{Zhang_2020_CVPR_UCNet}. The performance of each model is shown as \enquote{GAN} and \enquote{VAE} respectively in Table \ref{tab:col_wrt_loss_function_stomodel}. Note that the deterministic performance of the two latent variable model based COL is achieved by computing the performance of the mean predictions of multiple iterations (10 in this paper) of sampling from the latent space. The relatively consistent performance of our deterministic model with MSE loss and the two generative models explains that the extra random variable within the generative model will not deteriorate model deterministic performance.
We further show the uncertainty maps of the multiple predictions of each generative model in Fig.~\ref{fig:uncertainty_col}, where the uncertainty of each prediction is defined as variance of the multiple stochastic predictions.
Fig.~\ref{fig:uncertainty_col} shows that the extra uncertainty map leads to better understanding of the discriminative region, which serves as an effective tool to explain model prediction and
is critical for down-stream tasks.

\begin{table}[t!]
  \centering
  \scriptsize
  \renewcommand{\arraystretch}{1.2}
  \renewcommand{\tabcolsep}{1.7mm}
  \caption{Exploration of the COL model.}
  \begin{tabular}{l|cccccc}
  \hline
  Method&$SIM\uparrow$ & $CC\uparrow$ & $EMD\downarrow$ & $KLD\downarrow$ & $NSS\uparrow$ & $AUC\_J\uparrow$   \\ \hline
    \multicolumn{7}{c}{Loss function verification} \\ \hline
   KL loss &\textbf{0.614} & 0.769 & \textbf{1.994} & 4.282 & 1.148 & 0.963   \\
   CE loss &0.596 & 0.771 & 2.134 & 0.974 & 1.246 & 0.971   \\
   CC loss &0.579 & \textbf{0.773} & 2.313 & \textbf{0.717} & \textbf{1.293} & \textbf{0.973}   \\
   SIM loss &0.604 & \textbf{0.773} & 2.037 & 1.711 & 1.278 & \textbf{0.973}   \\ \hline
   \multicolumn{7}{c}{Generative model extension} \\ \hline
   GAN &0.575 & 0.766 & 2.373 & 0.804 & 1.271 & 0.972  \\
   VAE &0.571 & 0.767 & 2.424 & 0.757 & 1.253 & 0.971  \\
   \hline
   Base &0.572 & 0.767 & 2.408 & 0.791 & 1.265 & 0.971   \\ \hline
  \end{tabular}
  \label{tab:col_wrt_loss_function_stomodel}
\end{table}

\begin{table}[t!]
  \centering
  \scriptsize
  \renewcommand{\arraystretch}{1.2}
  \renewcommand{\tabcolsep}{6.3mm}
  \caption{The contribution analysis of COD and COL for COR.}
  \begin{tabular}{r|cccc}
  \hline
  Method & $MAE\downarrow$ & $r_{MAE}\downarrow$ & $Corr\uparrow$\\\hline
  JDR & 0.043 & 0.222 & 0.470 \\
  JLR & 0.043 & 0.223 & 0.468 \\
   \hline
  Base & 0.046 & 0.227 & 0.465 \\ 
  \hline
  \end{tabular}
  \label{tab:ranking_model_joint}
\end{table}

\subsubsection{Camouflaged Object Ranking}
Camouflaged object ranking evaluates the degrees of concealment of the camouflaged objects. As no benchmark models exist for this task, we resort to
salient object ranking, another ranking based task to generate our COR benchmark models.
The existing salient object ranking models \cite{amirul2018revisiting, siris2020inferring, liu2021instance, fang2021salient} aim to explore the correlation of salient instances within an image to decide the saliency level of each salient instance.
In this paper, we define camouflage ranking as at
dataset level. We record the observation time of each camouflage instance to decide the dataset-level camouflage ranking, which is reasonable in general as
camouflage ranking or the difficulty of camouflage is also global. We claim that the ranking of the camouflaged instance can provide a useful cue for camouflaged object detection and localization, leading to our final triple task learning pipeline in Fig.~\ref{fig:network_overview}. Considering the attention-shift based saliency ranking models, we aim to explore the contribution of the other two tasks for camouflage ranking. Towards this, we build two extra models, including a \enquote{Joint COD \& COR} model and a \enquote{Joint COL \& COR} model. We show the final ranking performance of these two models in Table \ref{tab:ranking_model_joint} as \enquote{JDR} and \enquote{JLR} respectively. The better performance of \enquote{JDR} and \enquote{JLR} compared with training only the individual COR model (\enquote{Base}) explains that the other two tasks can indeed be beneficial for COR. The slightly inferior performance of \enquote{JDR} and \enquote{JLR} compared with the COR prediction from our triple-task learning framework (\textbf{PTCOR} in Table \ref{tab:ablation_study_cor}) further explains the superiority of our triple-task learning framework.


\subsection{Camouflage Attributes Analysis}

We find the existing camouflaged object detection models are sensitive to the attributes of camouflaged instances. We then summarize the main attributes of camouflage.

\textit{Background Matching (BM)~\cite{background_matching}:} is the attribute when the foreground object and the surrounding background share
same color or texture.
\textit{Complex Background (CB):}
Images may have various levels of background complexity. We
compute image background complexity using \cite{Gen_TransformerSOD_NIPS}, and define images with a complexity measure higher than 0.12 as having a \enquote{Complex Background} attribute.
\textit{Corner Position (CP):} is a camouflage object location related attribute, which is used to decide if the position of the object center is far from image center.
\textit{Disruptive Coloration (DC):}~\cite{background_matching_color_disruptive} is the attribute to show whether there exists disruptive patterns around the object edges.
\textit{Mimicry (MM):} The camouflaged instance mimics certain species or non-living objects that often occur in their habitat, and we define this attribute as \enquote{Mimicry}.
\textit{Occlusion (OC):} indicates the object is partially occluded by other objects.
\textit{Salient Object (SA):} is defined as the attribute showing easier camouflaged instance.
\textit{Small Object (SO):} We define camouflaged instances that are smaller than
a pre-defined threshold, \eg~0.02 of the image, as having the \enquote{Small Object} attribute.



\begin{figure}[t!]
   \begin{center}
   \begin{tabular}{c@{ } c@{ }}
   {\includegraphics[height=0.32\linewidth]{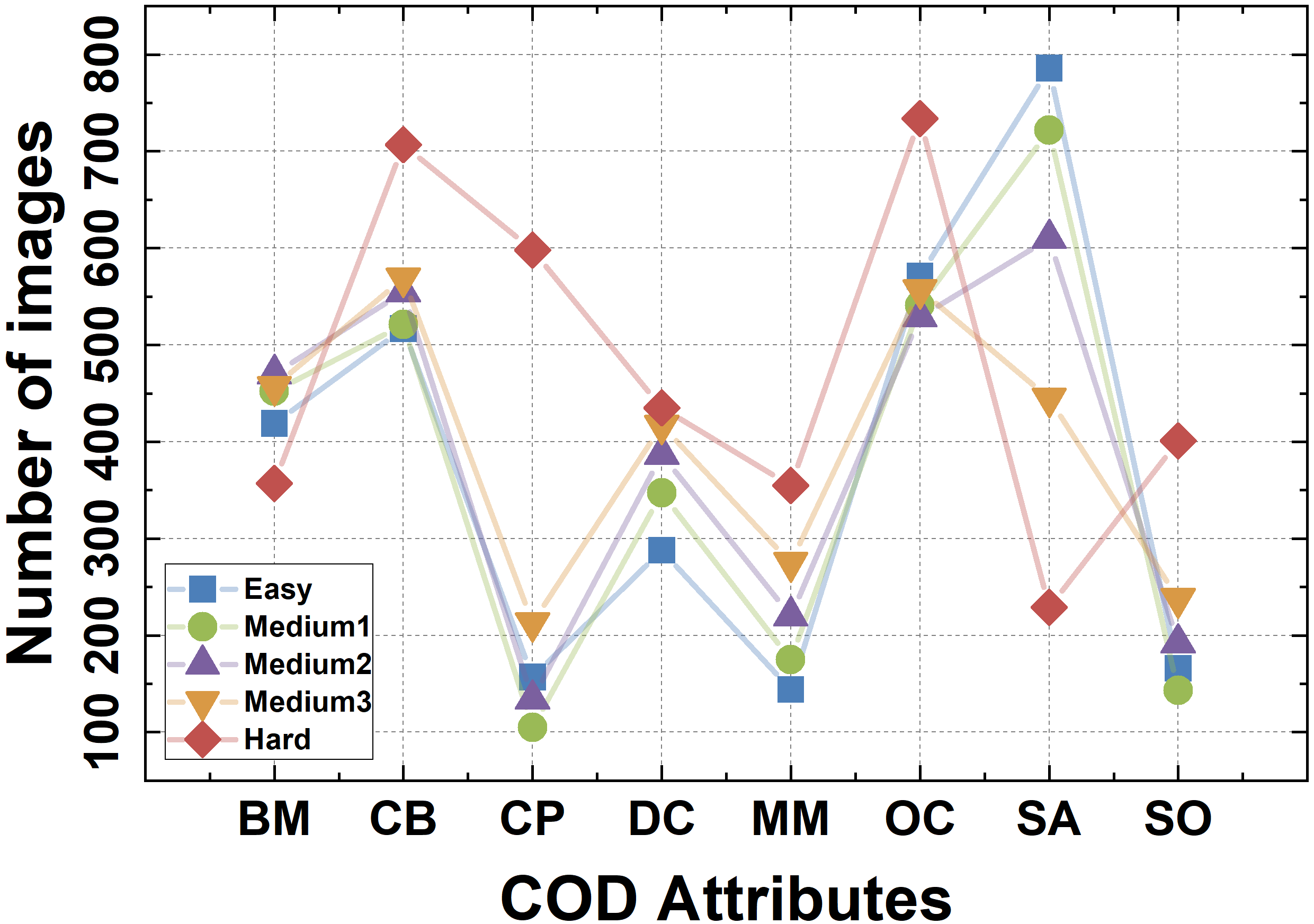}} &
   {\includegraphics[height=0.32\linewidth]{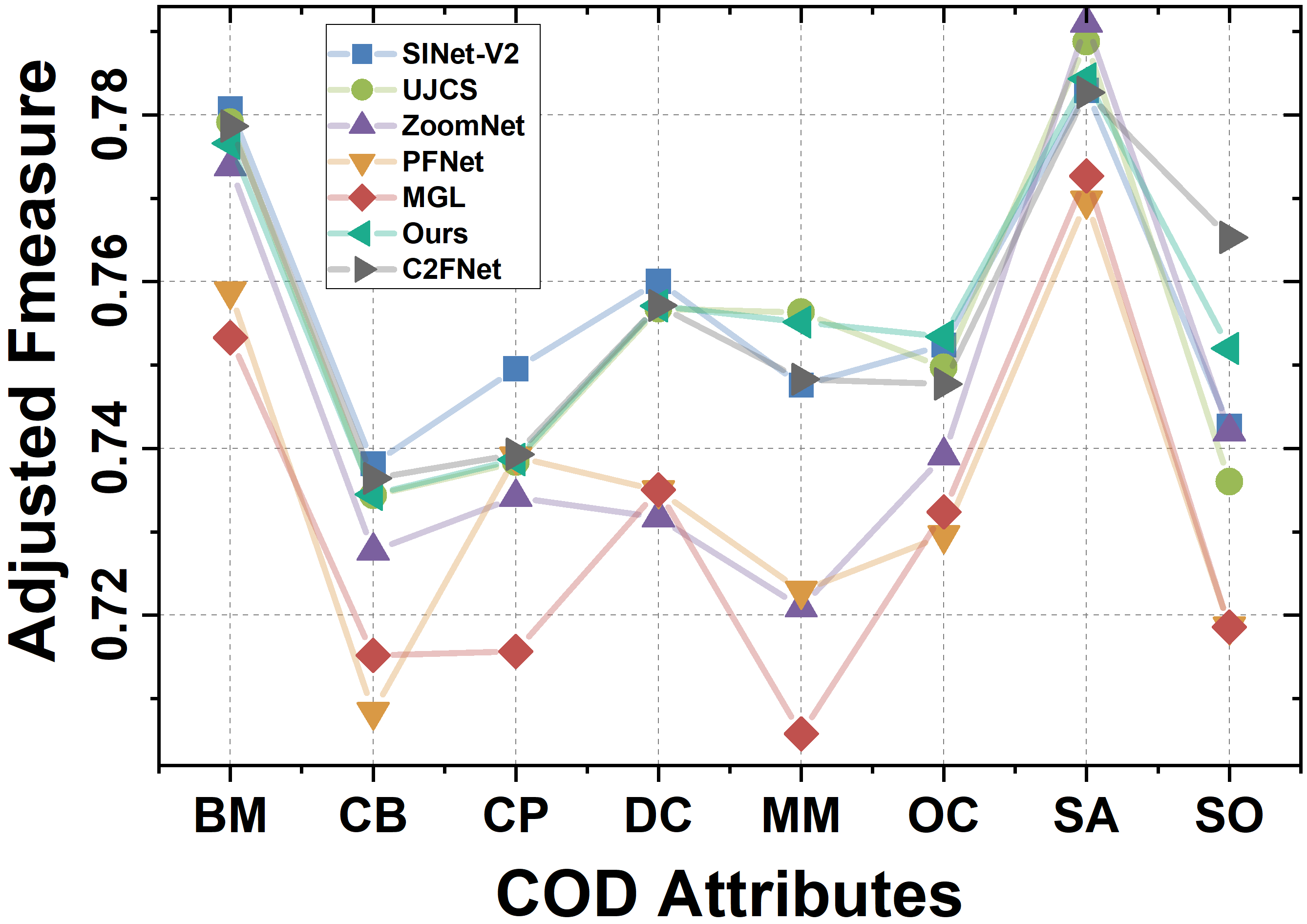}}\\
   (a)&(b)\\
   \end{tabular}
   \end{center}
   \caption{(a) The number of images for each rank level with respect to the fine-grained camouflage attributes. (b) Model performance with respect to camouflage attributes, where \enquote{Ours} is the base COD model in Table \ref{tab:ablation_study_cod}.
   }
\label{fig:camouflage_attributes}
\end{figure}

Based on the above fine-grained attributes, we re-sort the existing COD datasets (including the CAMO training dataset \cite{le2019anabranch}, the COD10K training dataset \cite{fan2020camouflaged} and the four benchmark testing datasets \cite{le2019anabranch, fan2020camouflaged, Chameleon2018, yunqiu_cod21})
to extensively analyse the effectiveness of each COD model for fine-grained task exploration. 
Firstly, we want to analyse the factors leading to the highest level of camouflage ranking (the hard samples). To do so, we summarize the ranking level of each camouflaged image with respect to each fine-grained camouflage attribute, and show the result in Fig.~\ref{fig:camouflage_attributes} (a). We observe that for images with the \textit{Salient Object} (SA) attribute, they are seldom classified as being a
\enquote{Hard} camouflaged instance. On the contrary, for images with the \textit{Complex Background} (CB), \textit{Corner Position} (CP), \textit{Occlusion} (OC) or \textit{Small Object} (SO) attributes, they tend to be \enquote{Hard} samples.
We can then conclude that
the hard camouflaged instances usually are evolved to obtain those attributes
to disguise themselves. Another interesting observation is that for the \textit{Mimicry}
(MM) 
attribute, we have the largest number of \enquote{Hard} samples, and least number of \enquote{Easy} samples, however, the opposite situation happens for the \textit{Salient Object} (SA) attribute, which is consistent with our understanding about camouflage. With this observation, we claim
that the
objects with higher camouflage degree tend to be less salient and vice versa.

With the fine-grained image attributes, we further analyse
performance of the COD methods with respect to each attribute, and show the results in Fig.~\ref{fig:camouflage_attributes} (b). For better visualization, we adjust the mean F-measure to 0.70-0.80 while maintaining the difference between the F-measure of all methods.
For example, the F-measure of C2FNet \cite{sun2021c2fnet} and SINet-V2 \cite{fan2021concealed} in \textit{Small Object} is adjusted from 0.442 and 0.465 to 0.742 and 0.765. We observe that SINet-V2 \cite{fan2021concealed} achieves the best performance in \textit{Background Matching}, \textit{Complex Background} and \textit{Corner Position} and \textit{Disruptive Coloration} attributes. UJCS \cite{li2021uncertainty} performs relatively superior in \textit{Background Matching}, \textit{Mimicry} (best) and \textit{Salient Object}. The performance of C2FNet \cite{sun2021c2fnet} ranks the top three and it excels in \textit{Small Object}. ZoomNet \cite{ZoomNet_CVPR2022} surpasses other algorithms in \textit{Salient Object}, although it has a relatively low performance in other attributes. Our model achieves the best performance in \textit{occlusion}, and ranks second in \textit{Disruptive Coloration}, \textit{Mimicry} and \textit{Small Object}. It has a balanced performance since its performance is in the middle in other attributes. As a result, the camouflaged attributes can help us to make a more comprehensive analysis of the performance of COD methods.

\section{Conclusions}
We have introduced two new tasks for camouflaged object detection, namely camouflaged object discriminative region localization and camouflaged object ranking, along with the relabeled corresponding datasets. The former aims to find the discriminative regions that make the camouflaged object detectable, and the latter tries to explain the level of camouflage, indicating the evolution degree of the camouflaged object. We built our network in a triple-task learning framework to simultaneously \emph{localize, segment, and rank} the camouflaged objects. Experimental results show that our proposed framework can achieve state-of-the-art performance, where the produced discriminative region and rank map provide insights toward understanding the nature of camouflage. We further provide camouflage attributes analysis, which can be insightful for further camouflage model designing. 



\bibliographystyle{IEEEtran}
\bibliography{Reference}
\end{document}